\documentclass[lettersize,journal]{IEEEtran}
\usepackage{amsmath,amsfonts}
\usepackage{algorithmic}
\usepackage{algorithm}
\usepackage{array}
\usepackage{textcomp}
\usepackage{stfloats}
\usepackage{url}
\usepackage{verbatim}
\usepackage{graphicx}
\usepackage{cite}
\usepackage{booktabs}
\usepackage{multirow, hhline}
\usepackage{subcaption}
\usepackage{flushend}
\usepackage{array}
\usepackage{pifont}% http://ctan.org/pkg/pifont
\newcommand{\cmark}{\ding{51}}%
\newcommand*{\genbf}[1]{\ifmmode\mathbf{#1}\else\textbf{#1}\fi}

\begin{document}

\title{Are Neural Architecture Search Benchmarks Well Designed? A Deeper Look Into Operation Importance}

\author{Vasco Lopes \qquad Bruno Degardin \qquad Luís A. Alexandre %\IEEEmembership{Staff,~IEEE,}
        % <-this % stops a space
\thanks{%Manuscript received 7 November, 2022. 
This work was supported by `FCT - Fundação para a Ciência e Tecnologia' through the research grant `2020.04588.BD', and partially supported by NOVA LINCS (UIDB/04516/2020) with the financial support of FCT, through national funds and by CENTRO-01-0247-FEDER-113023 - DeepNeuronic.
}
\thanks{V. Lopes and L. A. Alexandre are with NOVA Lincs, Universidade da Beira Interior (email: vasco.lopes@ubi.pt). B. Degardin is with Universidade da Beira Interior.}% <-this % stops a space

}

\markboth{V. Lopes and B. Degardin and L. A. Alexandre: }{}


\maketitle

\begin{abstract}
Neural Architecture Search (NAS) benchmarks significantly improved the capability of developing and comparing NAS methods while at the same time drastically reduced the computational overhead by providing meta-information about thousands of trained neural networks. However, tabular benchmarks have several drawbacks that can hinder fair comparisons and provide unreliable results. These usually focus on providing a small pool of operations in heavily constrained search spaces -- usually cell-based neural networks with pre-defined outer-skeletons. In this work, we conducted an empirical analysis of the widely used NAS-Bench-101, NAS-Bench-201 and TransNAS-Bench-101 benchmarks in terms of their generability and how different operations influence the performance of the generated architectures. We found that only a subset of the operation pool is required to generate architectures close to the upper-bound of the performance range. Also, the performance distribution is negatively skewed, having a higher density of architectures in the upper-bound range. We consistently found convolution layers to have the highest impact on the architecture's performance, and that specific combination of operations favors top-scoring architectures. 
These findings shed insights on the correct evaluation and comparison of NAS methods using NAS benchmarks, showing that directly searching on NAS-Bench-201, ImageNet16-120 and TransNAS-Bench-101 produces more reliable results than searching only on CIFAR-10. Furthermore, with this work we provide suggestions for future benchmark evaluations and design. The code used to conduct the evaluations is available at \url{https://github.com/VascoLopes/NAS-Benchmark-Evaluation}.
\end{abstract}

\begin{IEEEkeywords}
Neural Architecture Search, AutoML, Convolutional Neural Networks, Benchmarks, Cell-search, Operation Importance
\end{IEEEkeywords}

\section{Introduction}
%\IEEEPARstart{}{} 
Advances in deep learning algorithms obtained remarkable progress in various problems, mainly due to human experts' ingenuity and engineering efforts that exhaustively designed and engineered high performance architectures. Amongst deep learning algorithms, Convolutional Neural Networks (CNNs) are within the most used ones, having been applied with great success to a wide range of tasks with unprecedented results, from image classification \cite{deng2014deep,goodfellow2016deep}, to semantic segmentation \cite{garcia2018survey}, text analysis \cite{DBLP:conf/eacl/SchwenkBCL17}, amongst many others \cite{lecun2015deep, schmidhuber2015deep, khan2020survey}. 

CNN's inherent capability of feature extraction automated the process of feature engineering and provided a mechanism to ease transferability between different problems. Over the years, several proposals, such as skip and residual connections, resulted in improving CNN's capabilities to learn more efficiently while at the same time reducing their inference time and associated problems \cite{DBLP:conf/nips/KrizhevskySH12,DBLP:conf/cvpr/SzegedyLJSRAEVR15,DBLP:conf/cvpr/HeZRS16,DBLP:conf/cvpr/HuangLMW17,DBLP:conf/icml/TanL19,DBLP:conf/iclr/DosovitskiyB0WZ21}. However, such proposals are the result of years of expertise and trial and error, as designing CNNs is an endeavor that heavily relies on human expertise. Design choices intrinsic to the architectures, layer combination, and training require extensive architecture engineering, are heavily dependent on expertize and require heavy computation for a systematic and exhaustive evaluation. Thus, automating this process became logical, creating a growing interest in Neural Architecture Search (NAS) \cite{hutter2019automated}.  
NAS methods have successfully been applied to a panoply of problems, ranging from image classification tasks \cite{wei2021self,DBLP:conf/iclr/LiuSY19,zela2020understanding,xue2021self,DBLP:journals/corr/abs-2203-05508, lopes2020auto}, semantic segmentation \cite{liu2019auto, DBLP:conf/eccv/LiuDHGYX20}, object detection \cite{chen2019detnas,9492168} and image generation \cite{gong2019autogan,gao2020adversarialnas}, by consistently improving the search mechanisms \cite{DBLP:conf/iclr/LiuSY19, DBLP:conf/icml/PhamGZLD18, carlucci2019manas, DBLP:journals/corr/abs-2110-15232} and the performance estimation mechanisms \cite{DBLP:conf/aaai/WhiteNS21, mellor2020neural, epenas, DBLP:conf/iclr/ChenGW21}. NAS methods are usually comprised of three components. The search space defines the possible operations to be sampled and their connections, thus defining the type of architecture the search method can generate. The search method is the approach used to explore the search space and generate architectures. Common NAS methods use gradient-based, evolution strategies or Bayesian optimization to perform the search \cite{DBLP:journals/corr/abs-2301-08727,elsken2019neural,ren2021comprehensive}. Finally, the third component is the performance estimation strategy, which is the method used to evaluate the generated architectures. In a nutshell, the goal of a NAS method is to, based on the search method, efficiently search the space of possible networks to find an optimal architecture for a given problem, with minimal user interaction \cite{elsken2019neural}. 

Evaluating and comparing NAS methods is challenging, as different search spaces and training protocols difficult a fair comparison and hinder a true evaluation of how well a NAS method behaves. Researches have continuously focused on analyzing different NAS baselines to propose proper comparison metrics \cite{lindauer2020best,DBLP:journals/corr/abs-2107-03719,DBLP:conf/iclr/YuSJMS20}. Li and Talwaker extensively studied random-search in NAS and found that it presents a strong baseline for comparison, outperforming several NAS proposals  \cite{li2020random}. Then, Yang \textit{et. al.} conducted an extensive study to evaluate the impact of the training protocols on the results of a generated architecture, showing that well-engineered training protocols usually have a greater impact on the final performance of a NAS method than the search strategy, thus suggesting that NAS methods should provide results using the same protocols to ensure fair comparisons \cite{Yang2020NAS}. In \cite{wan2022on}, the authors analyzed popular cell-based search spaces. They found that existing search spaces contain a high degree of redundancy and generated architectures from distinct NAS methods have similar patterns. Studies to evaluate NAS performance predictors in terms of their correlation to the final architecture's performance, inference and initialization time have also been conducted \cite{DBLP:journals/corr/abs-2008-03064, white2021powerful, ning2021evaluating}. However, evaluating different NAS methods is still a challenge due to the unconstrained search spaces and training protocols. To mitigate the difficulty in designing and evaluating NAS methods, several benchmarks have been proposed to smooth the design and evaluation of NAS methods by providing fixed search spaces and hyper-parameters \cite{mehta2022nasbenchsuite}. Nonetheless, little attention has been devoted to the analysis of the benchmarks themselves to ensure that these provide fair and competitive settings that can and should be used as common ground for NAS methods' comparisons. %But, none of these sought to extensively evaluate the importance of the different operations of a benchmark search space and evaluate if it contains common patterns that jeopardize fair comparisons.

In this work, we set out to evaluate the impact that the operations, their combination, and their occurrences have in the final validation accuracy of an architecture in NAS-Bench-201, as well as evaluating the search space distribution in terms of the architecture's accuracy and their correlation between data sets. With this, we extend prior works that looked into proposing best practices for NAS methods, as well as evaluating search spaces and benchmarks \cite{lindauer2020best,Yang2020NAS,wan2022on,mehta2022nasbenchsuite}.

The main contributions of this paper are:
\begin{itemize}
    \item A detailed analysis of the importance of different operations in the final performance of a generated architecture, as well as their positioning, combinations, and occurrences. We found that convolutional layers and skip connections are the most important in all benchmarks evaluated.
    \item We study the inter-data set ranking of top-scoring architectures, showing that TransNAS-Bench-201 presents more heterogeneity than NAS-Bench-201, and that in all data sets, specific operations are present in key edges of the cell.
    \item We suggest a set of good practices for the design of future benchmarks and comparisons between NAS methods, when using different NAS benchmarks.
\end{itemize}

\section{Neural Architecture Search Benchmarks Overview}
\label{sec:preliminaries}

\begin{table*}[!t]
  \caption{Overview of NAS benchmarks and categorization based on their properties. Acronyms used in the table: Image Classification (IC), Computer Vision (CV), Automatic Speech Recognition (ASR), Natural Language Processing (NLP). \label{table:relatedbenchmarkeval}}
\centering
\resizebox{1\textwidth}{!}{%
\begin{tabular}{@{}lcccccccccl@{}}

\toprule
\multirow{2}{*}{\textbf{Benchmark}} & \multirow{2}{*}{\textbf{Size}} & \textbf{Search} & \multicolumn{2}{c}{\textbf{Type}} & \multirow{2}{*}{\#\textbf{Ops}} & \textbf{Data} & \multirow{2}{*}{\textbf{Task}} & \multirow{2}{*}{\textbf{License}} & \multirow{2}{*}{\textbf{Year}} & \multirow{2}{*}{\textbf{Venue}} \\ \cmidrule{4-5}
 &  & \textbf{Scheme} & \textbf{Tab.} & \textbf{Surr.} &  & \textbf{sets} &  &  &  &  \\
\midrule
NAS-Bench-101 \cite{DBLP:conf/icml/YingKCR0H19} & $423$k & cell & \cmark & & $3$ & $1$ & IC & A2.0 & 2019 & ICML \\
\midrule
NDS \cite{DBLP:conf/iccv/Radosavovic0XLD19} & $139$M & cell & \cmark$\dagger$ & & $5$-$8$ & $2$ & IC & MIT & 2019 & ICCV \\ % only 105k trained and available
\midrule
NAS-Bench-1Shot1 \cite{DBLP:conf/iclr/ZelaSH20} & $364$k & cell & \cmark & & $3$ & $1$ & IC & A2.0 & 2020 & ICLR \\
\midrule
\begin{tabular}[c]{@{}l@{}}NAS-Bench-201 \cite{Dong2020NAS-Bench-201}\\NATS-Bench \cite{dong2021nats}\end{tabular} & \begin{tabular}[c]{@{}l@{}}$15$k\\ $32$k\end{tabular} & \begin{tabular}[c]{@{}c@{}}cell\\ macro\end{tabular} & \cmark & & $5$ & $3$ & IC & MIT & \begin{tabular}[c]{@{}l@{}}2020\\ 2021\end{tabular} & \begin{tabular}[c]{@{}l@{}}ICML\\ TPAMI\end{tabular} \\
\midrule
LatBench \cite{DBLP:conf/nips/DudziakCALKL20} & $15$k & cell & \cmark & & $5$ & $3$ & IC & A2.0 & 2020 & NeurIPS \\ 
\midrule
TransNAS-Bench-101 \cite{DBLP:conf/cvpr/DuanCXCLZL21} & \begin{tabular}[c]{@{}c@{}}$4$k\\ 
$3$k\end{tabular} & \begin{tabular}[c]{@{}c@{}}cell\\ macro\end{tabular} & \cmark & & $4$ & $7$ & CV & MIT & 2021 & CVPR \\
\midrule
NAS-Bench-Macro \cite{su2021prioritized} & $6$k & macro & \cmark & & $3$ & $1$ & IC & A2.0 & 2021 & CVPR \\ %GOOD INTRO ON NAS WITH FORMULAS
\midrule
NAS-Bench-ASR \cite{DBLP:conf/iclr/MehrotraRBDVCAI21} & $8$k & cell & \cmark & & $8$ & $1$ & ASR & A2.0 & 2021 & ICLR \\
\midrule
NAS-Bench-111 \cite{yan2021bench} & $423$k & cell & & \cmark & $3$ & $1$ & IC & A2.0 & 2021 & NeurIPS \\
\midrule
NAS-Bench-311 \cite{yan2021bench} & $10^{18}$ & cell & & \cmark & $8$ & $1$ & IC & A2.0 & 2021 & NeurIPS \\ %(CIFAR-10)
\midrule
NAS-Bench-NLP11 \cite{yan2021bench} & $10^{53}$ & cell & & \cmark & $7$ & $1$ & NLP & A2.0 & 2021 & NeurIPS \\
\midrule
HW-NAS-Bench-201 \cite{DBLP:conf/iclr/LiYFZZYY0HL21} & $15$k & cell & \cmark & & $3$ & $3$ & IC & MIT & 2021 & ICLR \\
\midrule
HW-NAS-Bench-FBNet \cite{DBLP:conf/iclr/LiYFZZYY0HL21} & $10^{21}$ & chain & \cmark & & $9$ & $1$ & IC & MIT & 2021 & ICLR \\
\midrule
BLOX \cite{chau2022blox} & $91$k & macro & \cmark & & $5$ & $1$ & IC & CC & 2022 & NeurIPS \\ %(CIFAR-100)  A-NC 4.0
\midrule
JAHS-Bench-201 \cite{bansal2022jahsbench} & $270$k & cell & & \cmark & $5$ & $3$ & IC & MIT & 2022 & NeurIPS \\
\midrule
NAS-Bench-NLP \cite{DBLP:journals/corr/abs-2006-07116} & $10^{53}$ & cell & \cmark$\dagger$ &  & $7$ & $1$ & NLP & A2.0 & 2022 & \begin{tabular}[c]{@{}l@{}}IEEE\\Access\end{tabular} \\
\midrule
\begin{tabular}[c]{@{}l@{}}NAS-Bench-301 \cite{DBLP:journals/corr/abs-2008-09777}\\(Surr-NAS-Bench-DARTS)\end{tabular} & $10^{18}$ & cell & & \cmark & $8$ & $1$ & IC & A2.0 & 2022 & ICLR \\ %(CIFAR-10) \\ 
\midrule
Surr-NAS-Bench-FBNet \cite{DBLP:journals/corr/abs-2008-09777} & $10^{21}$ & chain & & \cmark & $9$ & $1$ & IC & A2.0 & 2022 & ICLR \\ %(CIFAR-100)
\midrule
NAS-Bench-MR \cite{DBLP:conf/iclr/DingHLYW00L22} & $10^{23}$ & cell & & \cmark & - & $9$ & CV & BSD & 2022 & ICLR \\ %BSD 2-Clause
\bottomrule
\end{tabular}
}
\begin{flushleft}\footnotesize $\dagger$ only a small subset of the entire search space was fully trained and provided as a tabular benchmark.\end{flushleft}
%\multicolumn{10}{p{\dimexpr\textwidth-2\tabcolsep\relax}}{$\dagger$ {\footnotesize Only a small subset of the entire search space was fully trained and provided as a tabular benchmark}}
\vspace{-1.5em}
\end{table*}
% related benchmarks
% NAS-Bench-Zero (openreview only - let's wait for arxiv or publication)
% LCBench

Evaluating and comparing NAS methods can be a laborious task. Different NAS methods often employ tailored training protocols, use search spaces with different operation pools, and report different evaluation metrics, making it difficult to make optimal comparisons. This often leads to sub-optimal comparisons and necessitates high computational costs to ensure fair comparisons in similar settings. To address this issue, several benchmarks have been proposed to facilitate the design and evaluation of NAS methods. These benchmarks allow NAS methods to access information about a unified search space and force fairer comparisons between methods by fixing hyper-parameters and training protocols \cite{lindauer2020best}. There are two types of benchmarks: tabular and surrogate. Tabular benchmarks provide precomputed information on all possible architectures trained in one or more data sets, while surrogate benchmarks provide a model capable of predicting the performance of an architecture, along with precomputed information about some of the architectures. By providing access to information, either through a tabular setting or a surrogate model, benchmarks can reduce the need for computational resources, enabling quick and efficient evaluations.

Table \ref{table:relatedbenchmarkeval} provides a detailed overview of the existing NAS benchmarks. NAS-Bench-101 \cite{DBLP:conf/icml/YingKCR0H19} was the first tabular benchmark proposed and provides information about the validation and test accuracy of the 423,624 cell-based architectures that make up the search space. Each architecture was trained on CIFAR-10 with different initialization random seeds. NAS-Bench-1Shot1 \cite{DBLP:conf/iclr/ZelaSH20} leverages the information in NAS-Bench-101 by defining subsets of the search space with a fixed number of nodes, enabling one-shot NAS methods. The NDS benchmark \cite{DBLP:conf/iccv/Radosavovic0XLD19}, on the other hand, focuses on evaluating different randomly sampled architectures from five search spaces.

Another tabular benchmark, NAS-Bench-201 \cite{Dong2020NAS-Bench-201}, proposes a smaller search space but with added information in different data sets. This benchmark includes 15,625 trained architectures on CIFAR-10, CIFAR-100, and ImageNet16-120. Additionally, LatBench augments NAS-Bench-201 by providing latency information about all the architectures in six different devices. Similarly, HW-NAS-Bench \cite{DBLP:conf/iclr/LiYFZZYY0HL21} proposes two benchmarks: HW-NAS-Bench-201, an extension of NAS-Bench-201, and HW-NAS-Bench-FBNet, which uses the FBNet search space. The focus of these benchmarks is to measure and estimate the latency and energy consumption of all architectures in the initial search spaces in six different devices. The authors of NAS-Bench-201 also proposed NATS-Bench \cite{dong2021nats}, which provides a macro search space with tabular information for 32,768 architectures. More, JAHS-Bench-201 \cite{bansal2022jahsbench} also extends the search space proposed by NAS-Bench-201 to a joint optimization of the architecture design and hyper-parameters. The authors designed a combination of the 15,625 architectures with different hyper-parameters, totaling 270,000 configurations for which 20 performance metrics in 3 data sets were evaluated, thus providing more than 161M surrogate data points.

Benchmarks that provide a macro search space enable the evaluation of NAS methods beyond traditional operation sampling, extending to decisions about node connections, operation parameters, or the architecture skeleton. One such benchmark is NAS-Bench-Macro \cite{su2021prioritized}, which proposes a search space with 6,561 architectures. The goal of this benchmark is to search for 8 layers with a pool of 3 possible blocks. Another macro-based benchmark is Blox \cite{chau2022blox}, which provides tabular information about $91,125$ unique architectures. These architectures are designed by fixing an outer-skeleton and searching for 3 blocks with diverse connectivities.

The original focus of NAS benchmark was on evaluating architectures for image classification (IC) tasks. However, recent benchmarks have expanded to allow the evaluation of NAS methods in terms of generability and transferability to different tasks. One example is TransNAS-Bench-101, which offers both a cell-based and macro-based search space, and evaluates architectures over seven different computer vision (CV) problems, with the aim of providing a framework for evaluating NAS transferability across tasks \cite{DBLP:conf/cvpr/DuanCXCLZL21}. Another example is NAS-Bench-MR, which offers a surrogate benchmark for four different CV tasks, including segmentation \cite{DBLP:journals/corr/abs-2103-13253}. To further expand the scope of NAS applications, NAS-Bench-ASR proposes a cell-based search space with $8,242$ trained architectures for automatic speech recognition \cite{DBLP:conf/iclr/MehrotraRBDVCAI21}, while NAS-Bench-NLP focuses on proposing a benchmark for natural language processing, providing tabular information for a subset of the entire search space \cite{DBLP:journals/corr/abs-2006-07116}.

%The initial focus of NAS benchmarks was to evaluate architectures on image classification (IC) tasks. However, to allow the evaluation of NAS methods in terms of their generability and transferability, recent benchmarks have been proposing the use of different tasks. TransNAS-Bench-101 \cite{DBLP:conf/cvpr/DuanCXCLZL21} proposes both a cell and a macro-based space and provides architecture evaluations over 7 different computer vision (CV) problems. The focus is to provide a setup for the evaluation of NAS transferability between problems. NAS-Bench-MR provides a surrogate benchmark for 4 different CV tasks, including segmentation \cite{DBLP:journals/corr/abs-2103-13253}. NAS-Bench-ASR extends NAS applications to automatic speech recognition by proposing a cell-based search space with 8242 trained architectures \cite{DBLP:conf/iclr/MehrotraRBDVCAI21}. Similarly, NAS-Bench-NLP \cite{DBLP:journals/corr/abs-2006-07116} focuses on proposing a benchmark for natural language processing and provides tabular information of a subset of the entire search space. 

%More recently, surrogate benchmarks have been proposed to allow larger search space evaluations by providing a surrogate model. 

Surrogate benchmarks have been proposed to enable the design and evaluation of larger search spaces by training a set of architectures and then fitting a surrogate model capable of inferring the performance of newly generated architectures. Two such benchmarks are NAS-Bench-301 and Surr-NAS-Bench-FBNet, which fully train a set of architectures and fit surrogate models to estimate the final performance of the remaining architectures \cite{DBLP:journals/corr/abs-2008-09777}. However, the use of surrogate benchmarks often only provides the final surrogate performance estimation, which does not support the development of multi-fidelity NAS methods. To address this issue, the authors of \cite{yan2021bench} propose NAS-Bench-x11, which provides a surrogate method for predicting the full learning curve of an architecture. By using this method, the authors extend three existing benchmarks (NAS-Bench-101, NAS-Bench-301, and NAS-Bench-NLP) to include the full learning curve, enabling the development and evaluation of NAS methods that rely on the training curve.

NAS benchmark suites aim to unify various benchmarks by providing a common interface to query them, allowing for easy transferability and evaluation of NAS methods. The first suite, NAS-Bench-360 \cite{DBLP:journals/corr/abs-2110-05668}, introduced 2D and 1D data sets. NAS-Bench-Suite \cite{mehta2022nasbenchsuite} is a more comprehensive suite, consisting of 28 different tasks. To further improve NAS-Bench-Suite, NAS-Bench-Suite-Zero \cite{DBLP:journals/corr/abs-2210-03230} provides precomputed information about several zero-cost proxy estimators. These suite-benchmarks make it easier for researchers to evaluate and compare NAS methods across different tasks and data sets, ultimately aiding in the development of more effective and generalizable NAS methods.

%For this study, we focused on evaluating the importance of the operations that compose a search space in different benchmarks. The hypothesis is that the design of the benchmarks by itself directly correlates with the final performance of the entire pool of architectures, thus indirectly promoting better results for NAS methods that learn the underlying operation bias. For this, we focused on using tabular benchmarks, as they provide the performance for several architectures after being trained under the same settings. To verify the proposed hypothesis we focused on evaluating the importance of operations on NAS-Bench-101, NAS-Bench-201, and TransNAS-Bench-101 as these benchmarks are the most used tabular NAS benchmarks for evaluating search methods.

In this study, our main objective is to assess the significance of operations in a search space across various benchmarks. We hypothesize that the benchmark design plays a crucial role in determining the performance of the architecture pool, and thus indirectly influences the success of NAS methods that learn the underlying operation preference. To test this hypothesis, we limit our focus to tabular benchmarks that provide architecture performance data after being trained under similar settings. Specifically, we examine the importance of operations in NAS-Bench-101, NAS-Bench-201, and TransNAS-Bench-101 as they are arguably the most commonly used tabular benchmarks for evaluating search methods.

\begin{figure*}[tb]
\centering
\begin{subfigure}[t]{0.4\textwidth}
\centering
    \includegraphics[width=1\linewidth]{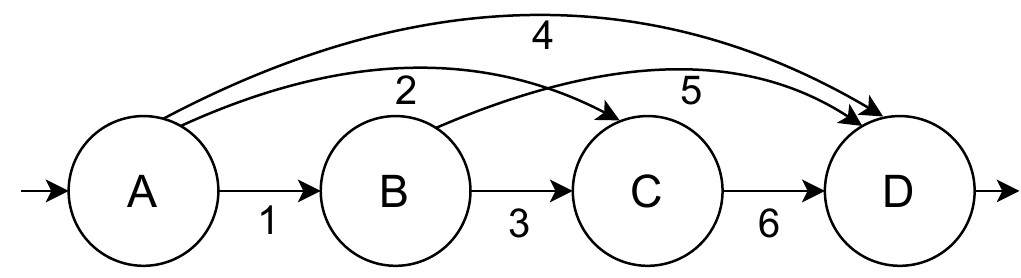}
    \caption{Cell Structure}
    \label{fig:inclu}
\end{subfigure}%
\begin{subfigure}[t]{0.6\textwidth}
\centering
    \includegraphics[width=0.9\linewidth]{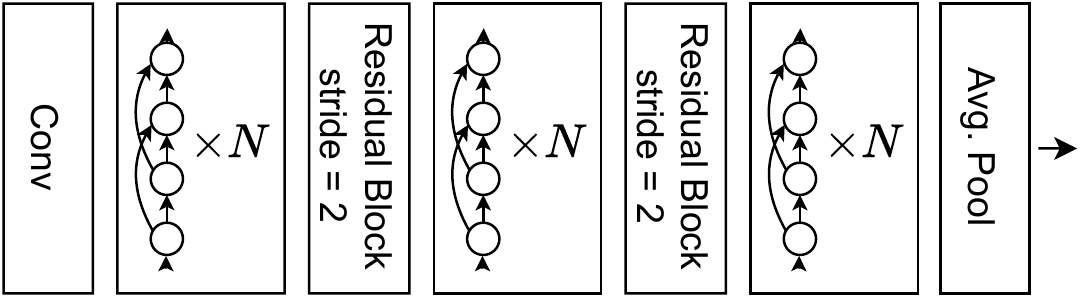}
    \caption{Outer-skeleton Structure}
    \label{fig:deform}
\end{subfigure}
\caption{Representation of the NAS-Bench-201 and TransNAS-Bench-201 cell structure (a), and the pre-defined outer-skeleton (b). A cell is composed of 4 nodes and 6 edges, where edges perform operations from an input node and add them to a posterior node. There are 5 possible operations to be used as edges in NAS-Bench-201 and 4 in TransNAS-Bench-101. A cell is used as the building block of the outer-skeleton.}\label{fig:nb201cellskeleton}
\vspace{-0.7em}
\end{figure*}

\section{Results}
\label{sec:benchevaluation}

\subsection{Search Spaces}
\label{subsec:nb101eval}
\textbf{NAS-Bench-101} is a tabular benchmark used for evaluating the performance of different NAS methods. It consists of a cell-based search space comprising $423,624$ neural networks trained on CIFAR-10 for 108 epochs with 3 different weight initializations. All the architectures in this benchmark share a common outer-skeleton, and the differences between them lie in the cells placed in the outer-skeleton. The cells in NAS-Bench-101 are defined as directed acyclic graphs with a maximum of 7 nodes and 9 edges and are encoded as a $7\times7$ upper-triangular matrix. The operation pool consists of three possible layers: convolutional $1\times1$ ($C_{1\times1}$), convolution $3\times3$ ($C_{3\times3}$), and $3\times3$ max pooling ($MP_{3\times3}$).

%is a tabular cell-based search space consisting of $423,624$ neural networks trained on CIFAR-10 for 108 epochs with 3 different weight's initialisation. All architectures share the a common outer-skeleton where the differences are the cells that are placed in the outer-skeleton. In NAS-Bench-101 cells are defined as a directed acyclic graph with a maximum of 7 nodes and 9 edges, and are encoded as a $7\times7$ upper-triangular matrix. The operation pool in NAS-Bench-101 consists 3 possible layers: convolutional $1\times1$ ($C_{1\times1}$), convolution $3\times3$ ($C_{3\times3}$), and $3\times3$ max pooling ($MP_{3\times3}$). 

\noindent \textbf{NAS-Bench-201} is a tabular cell-based search space with 15625 possible architectures, each composed of a fixed cell-based design with 5 possible operations: None, Skip Connection ($SC$), $C_{1\times1}$, $C_{3\times3}$, and Average Pooling $3\times3$ ($AP_{3\times3}$). The structure of a cell is represented as a directed acyclic graph with 6 edges and 4 nodes, where edges represent operations. All architectures share a common outer-skeleton and are designed by interleaving searched cells with residual down-sampling blocks. The benchmark provides information about the learning curve and final performance of the architectures on CIFAR-10, CIFAR-100, and ImageNet16-120, recorded and provided as a tabular benchmark. Figure \ref{fig:nb201cellskeleton} depicts the representation of the cell structure and the outer-skeleton.
 %NAS-Bench-201 provides information regarding the training and performance of all possible networks in the search space in three different data sets: CIFAR-10, CIFAR-100 and ImageNet16-120.%, thus proposing a controlled setting that allows different NAS methods to be directly compared, as they are forced to use the search space, training procedures, and hyper-parameters. 

\noindent \textbf{TransNAS-Bench-101} benchmark provides the performance of architectures across seven vision tasks, including classification, regression, pixel-level prediction, and self-supervised tasks. By having multiple tasks that can be evaluated using the same input, TransNAS-Bench-101 provides an easy setup to evaluate the generality and transferability of NAS across different tasks. There are two types of search spaces in this benchmark: a cell-based search space containing $4096$ possible cells and a macro skeleton search space based on residual blocks containing $3256$ architectures. For both search spaces, the pool of operations has four possible candidates: zeroize, $SC$, $C_{1\times1}$, and $C_{3\times3}$. TransNAS-Bench-101 provides tabular information about the training and performance of all architectures in the search space using the same training protocols and hyper-parameters within each task. In this study, we focus on the cell search space. The representation of the cell structure is depicted in Figure \ref{fig:nb201cellskeleton}.
%The 7 tasks of this benchmark are: object classification, scene classification, autoenconding, surface normal, semantic segmentation, room layout and jigsaw.

%%%%%%%%%%%
%   dist  %

\begin{figure}[tb]
    \centering
    \resizebox{0.9\columnwidth}{!}{%
        \includegraphics[width=0.5\linewidth]{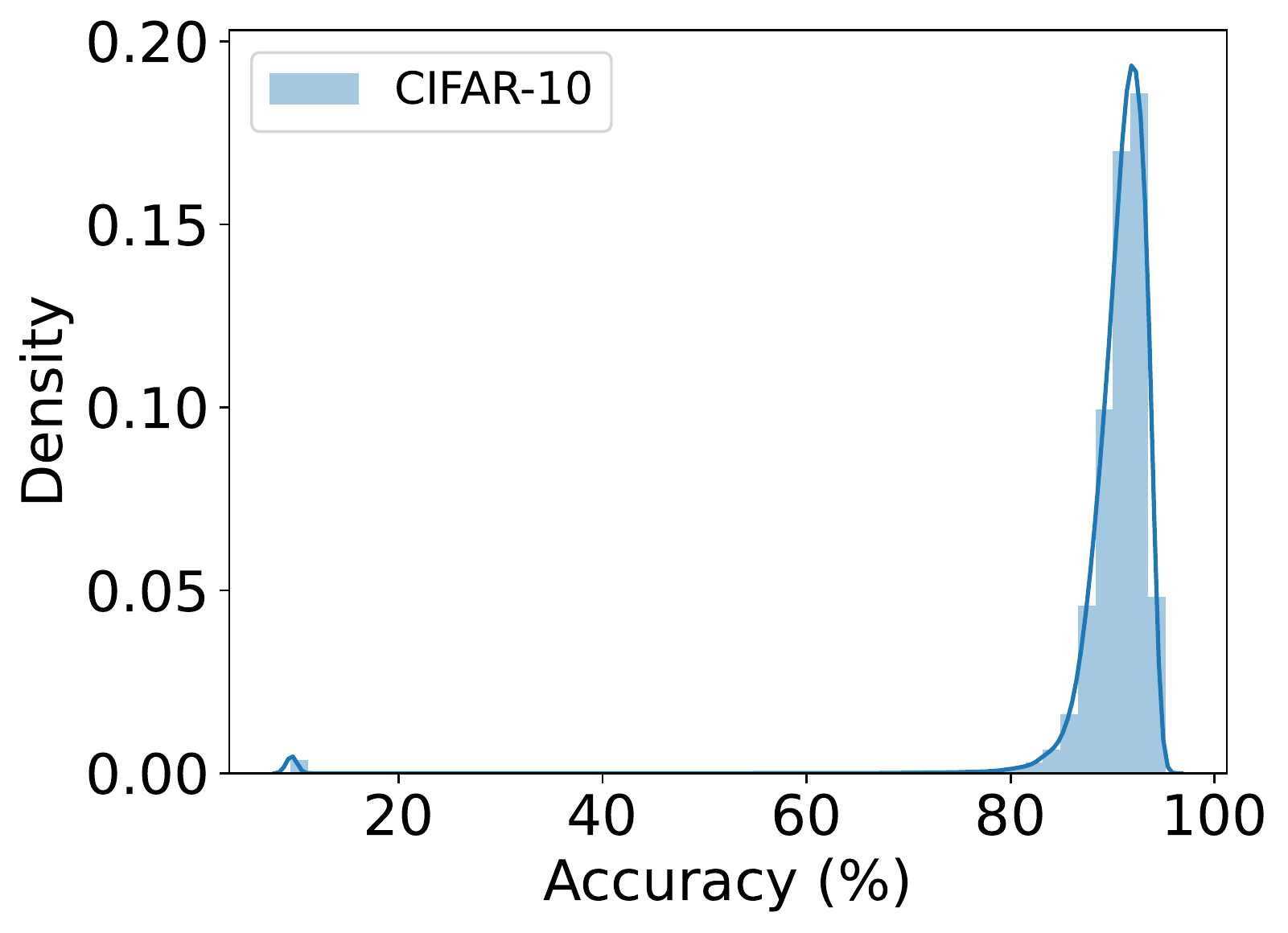}
    }
  \caption{Distribution and density of all architectures, based on their validation accuracy (\%) in NAS-Bench-101. \label{fig:nb101dist}}
\end{figure}

\begin{figure}[tb]
    \centering
    \resizebox{0.9\columnwidth}{!}{%
        \includegraphics[width=0.5\linewidth]{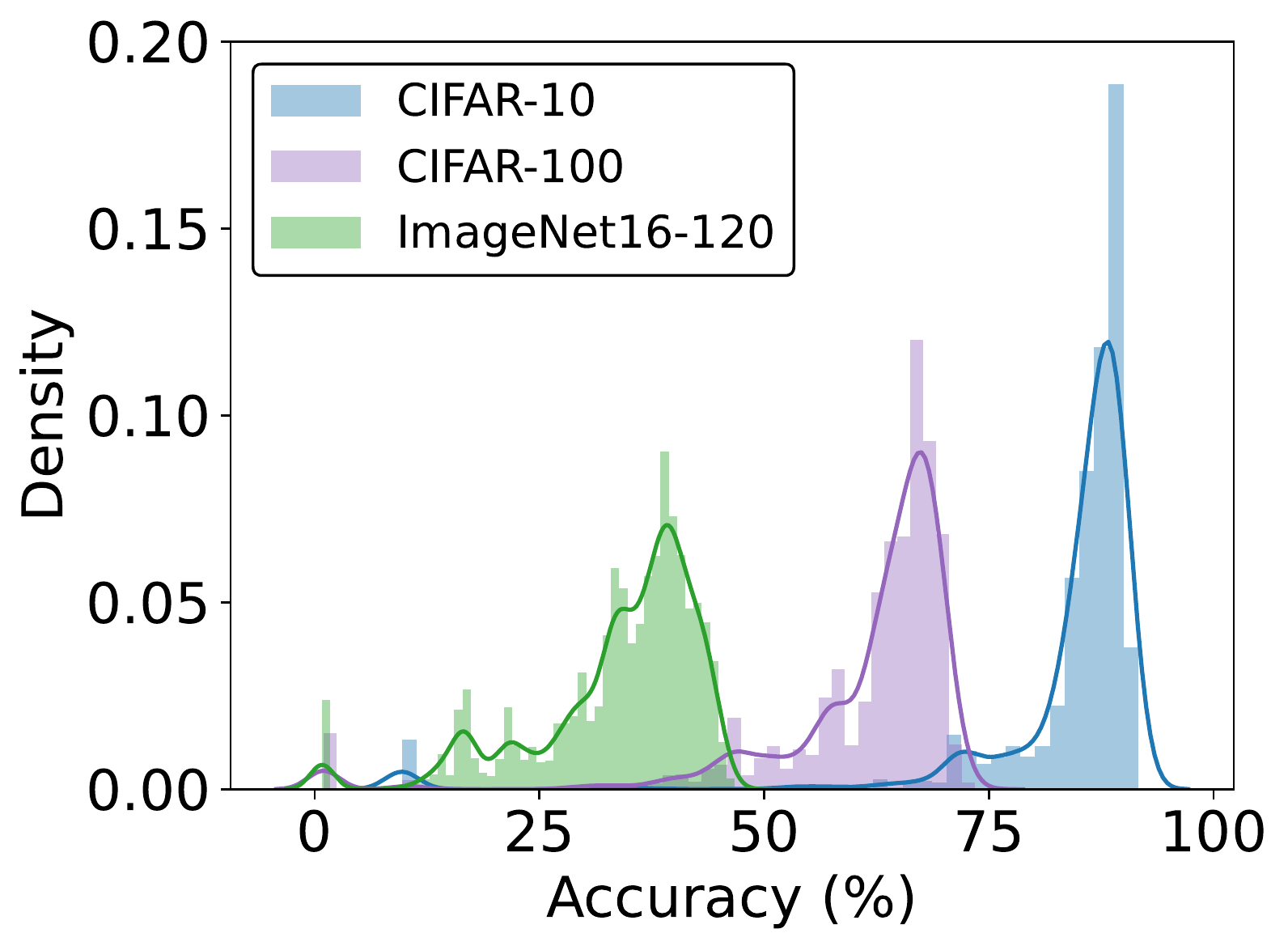}
    }
  \caption{Distribution and density of all architectures, based on their validation accuracy (\%), for all the data sets present in NAS-Bench-201. \label{fig:nb201dist}}
\end{figure}

\begin{figure}[tb]
    \centering
    \resizebox{0.9\columnwidth}{!}{%
        \includegraphics[width=0.5\linewidth]{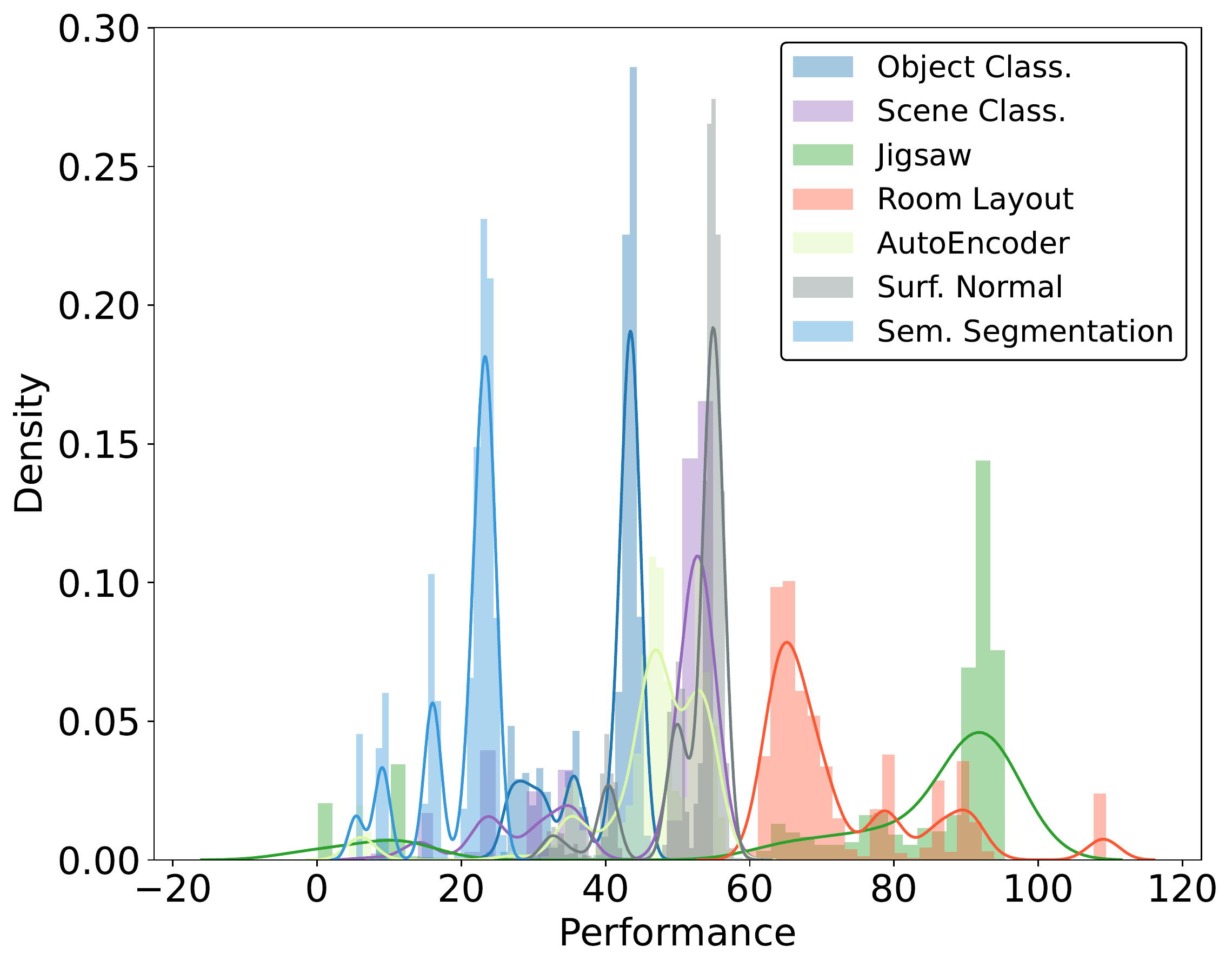}
    }
  \caption{Distribution and density of all architectures, based on their performance on all the data sets present in TransNAS-Bench-101. Metrics were scaled to positive values to match accuracy ranges.\label{fig:tnb201dist}}
\end{figure}

%   dist  %
%%%%%%%%%%%

\subsection{Evaluation}
%NAS-Bench-201 provides a tabular benchmark that allows querying the performance of all architectures in the search space. As such, we initially sought to evaluate the \textbf{distribution} of the architectures based on their validation accuracy.
\subsubsection{Performance Distribution}
By providing tabular information about trained architectures, NAS benchmarks allow direct querying of an architecture performance while searching. Given the information on all architectures, we evaluated the distribution of the architecture's performance in NAS-Bench-101, NAS-Bench-201, and TransNAS-Bench-201. For NAS-Bench-101, Fig. \ref{fig:nb101dist} shows a negatively skewed distribution of architecture performances, with a higher density of architectures closer to the upper bound of validation accuracy. This indicates that there are many optimal architectures with small differences, making it challenging to compare different NAS methods. Similarly, NAS-Bench-201 also shows a negatively skewed distribution, as seen in Fig. \ref{fig:nb201dist}. However, the skewed distributions are more evident on CIFAR data sets, suggesting that ImageNet16-120 is a more competitive data set with more likelihood of showing more differences between different NAS methods. Regarding TransNAS-Bench-201, as shown in Figure \ref{fig:tnb201dist}, there is a higher density of architectures closer to the optimal performance in all tasks, with semantic segmentation and room layout being the tasks with the sparsest distribution. This observation suggests that it is important for NAS methods to indicate the results of Random Search (RS) when comparing on a specific benchmark to demonstrate how well an RS method can sample the search space. Additionally, when comparing gains in performance, it is important to compare with the most optimal architecture in each task to ensure that small increments in performance are significant if closer to the upper-bound.

%Fig. \ref{fig:nb101dist} depicts the distribution and density of all architectures in NAS-Bench-101 in terms of their final validation accuracy. The results show a negatively skewed distribution, with a higher density of architectures closer to the upper bound of the validation, suggesting that the results of the possible optimal generated architectures have small differences, therefore creating a challenging evaluation and comparison of different NAS methods. Similarly, NAS-Bench-201 also presents a negatively skewed distribution. This can be seen in Fig. \ref{fig:nb201dist}, where the skewed distributions are more evident on CIFAR data sets, where the number of architectures with high accuracies is substantially higher, thus suggesting that ImageNet16-120 is a more competitive data set with more likelihood of showing more differences between different NAS methods. Finally, TransNAS-Bench-201 (Figure \ref{fig:tnb201dist}) shows a higher density closer to the optimal performance in all data sets, being semantic segmentation, room layout, and semantic segmentation the tasks with a more sparse distribution. This behavior further promotes the idea that NAS methods should indicate the results of Random Search (RS) when comparing on a specific benchmark to showcase how well a RS method can model the search space. Furthermore, gains in performance should be compared with the most optimal architecture in each task to ensure that increments in performance are evidenced.

%%%%%%%%%%%
% spider  %

\begin{figure}[tb]
    \centering
    \resizebox{0.8\columnwidth}{!}{%
        \includegraphics[width=0.5\linewidth]{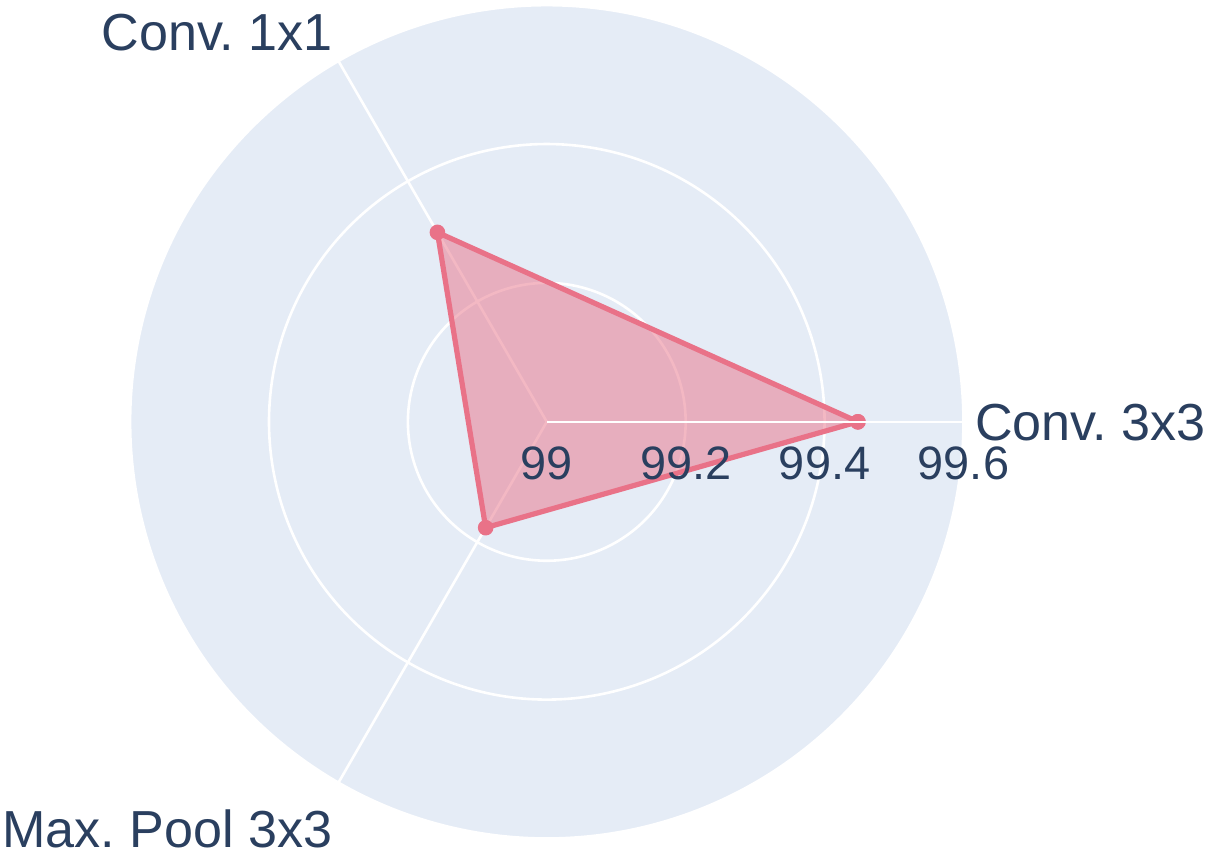}
    }
  \caption{Mean accuracy (\%) of all the architectures that have at least one operation of a given type in NAS-Bench-101.\label{fig:nb101spider}}
\end{figure}

\begin{figure*}[tb]
    \centering
    \resizebox{0.85\textwidth}{!}{%
    \begin{subfigure}[t]{0.3\textwidth}
    \centering
        \includegraphics[width=1\linewidth]{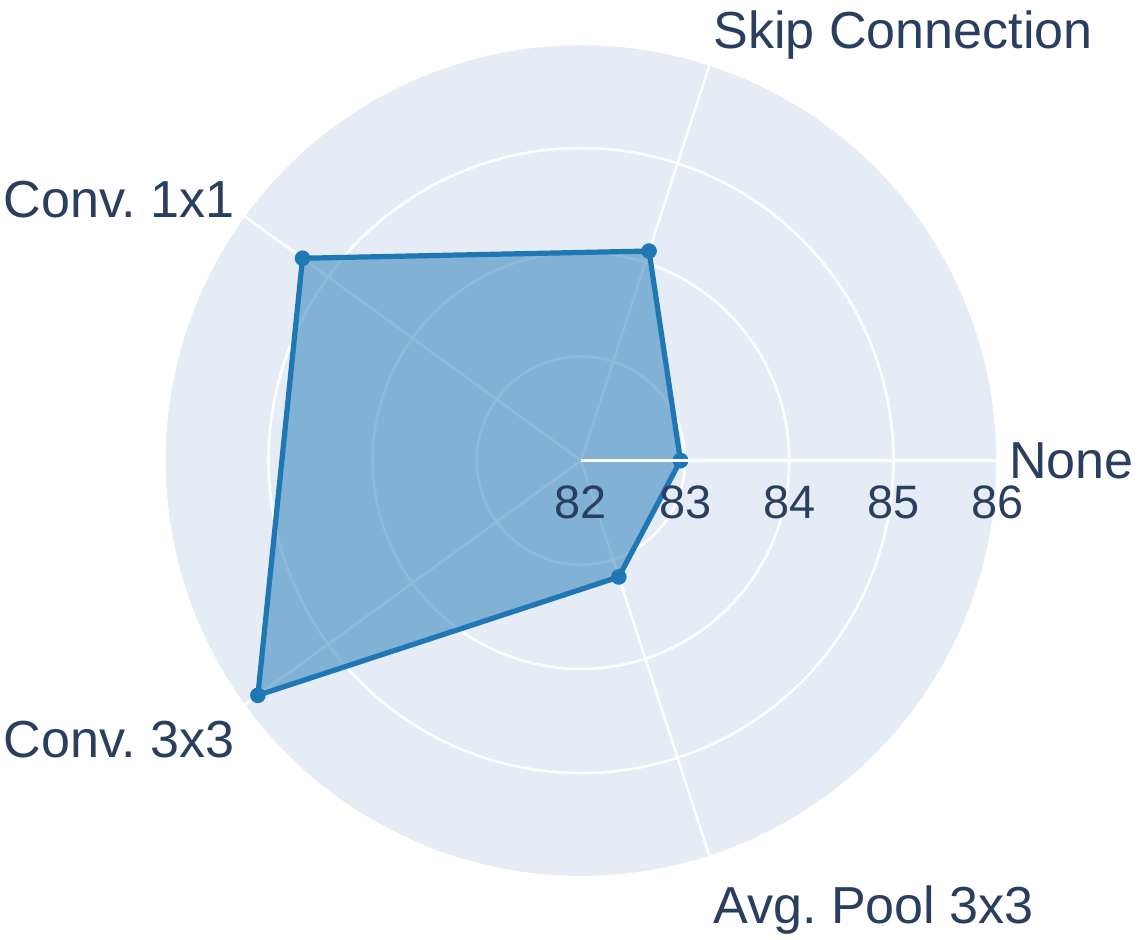}
        \caption{CIFAR-10}
        \label{fig:spider-c10}
    \end{subfigure}%
    \begin{subfigure}[t]{0.3\textwidth}
    \centering
        \includegraphics[width=1\linewidth]{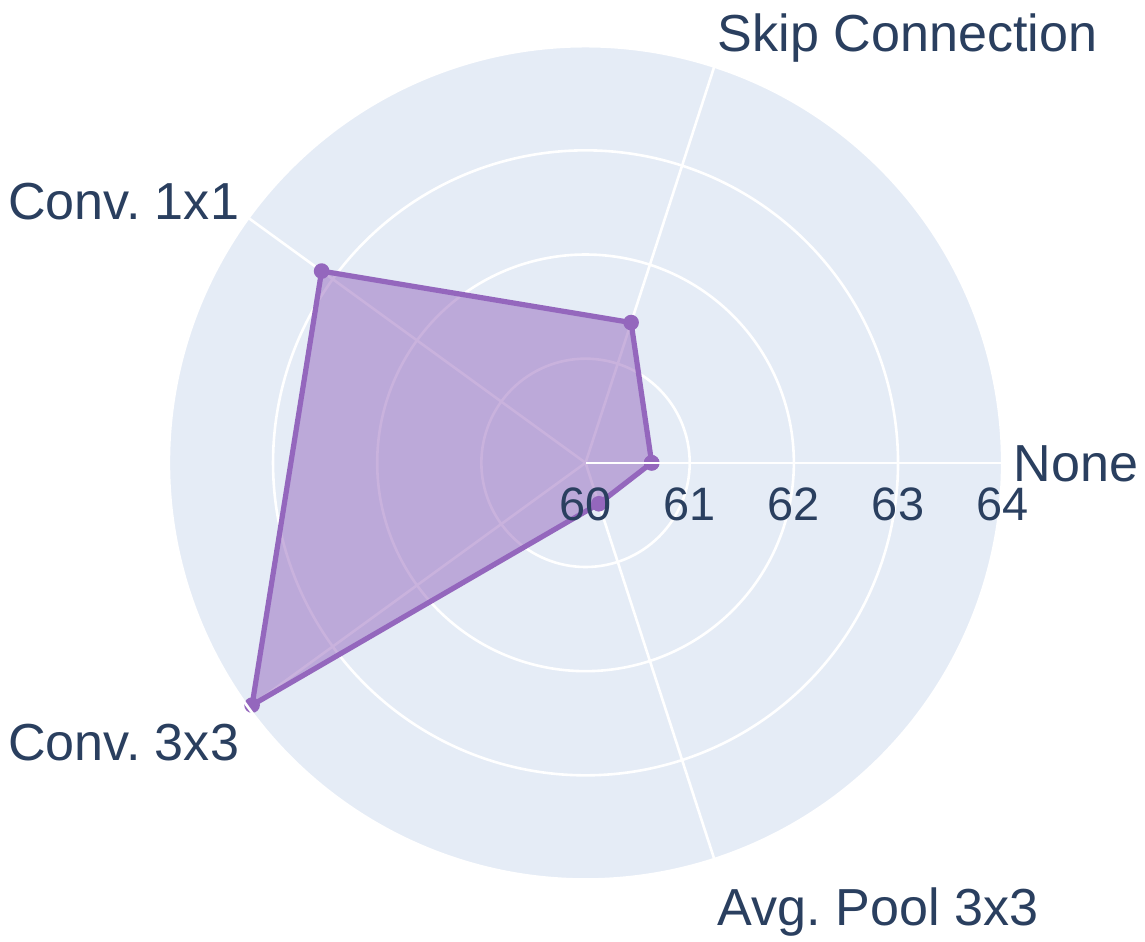}
        \caption{CIFAR-100}
        \label{fig:spider-c100}
    \end{subfigure}
        \begin{subfigure}[t]{0.3\textwidth}
    \centering
        \includegraphics[width=1\linewidth]{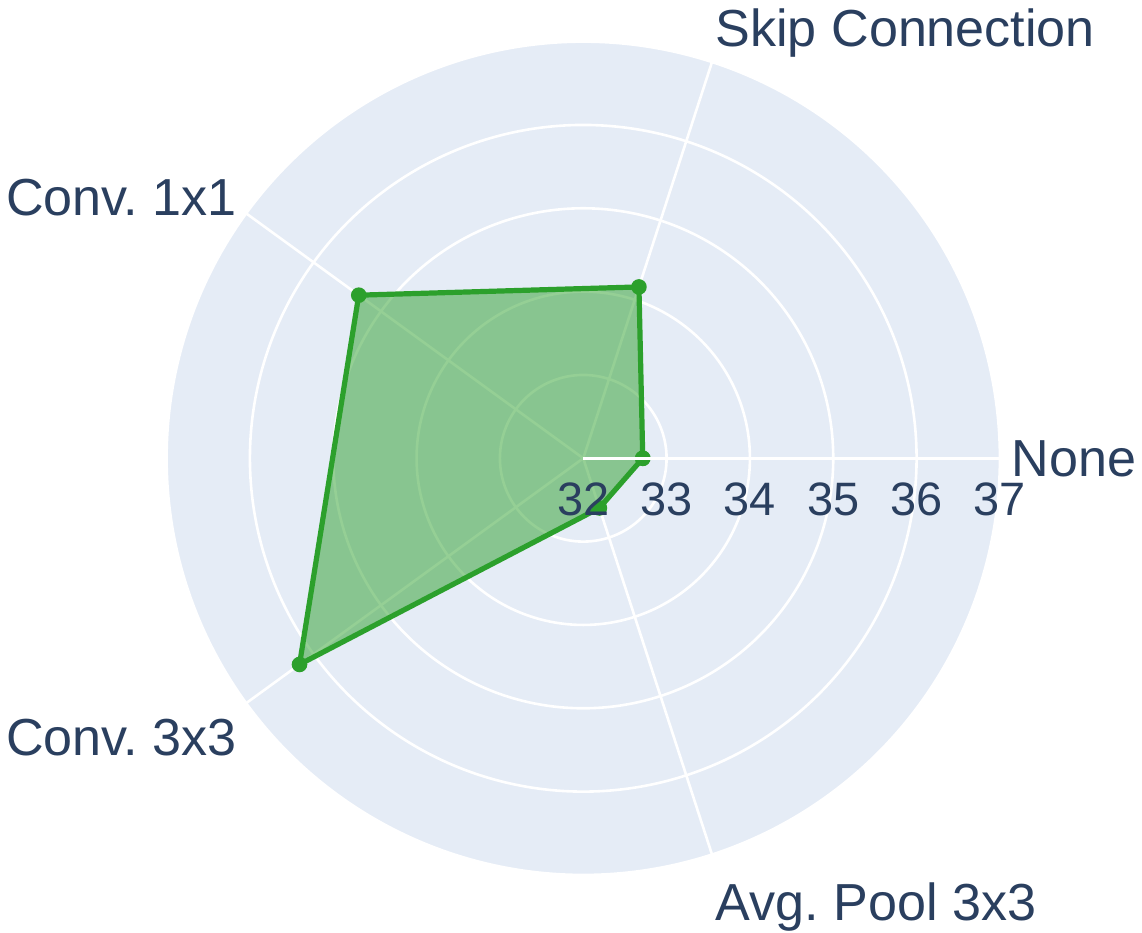}
        \caption{ImageNet16-120}
        \label{fig:spider-imgnet16120}
    \end{subfigure}
    }
  \caption{Mean accuracy (\%) of all the architectures that have at least one operation of a given type in NAS-Bench-201. Architectures with at least 1 convolutional layer show higher mean validation accuracies (\%) in all data sets. \label{fig:nb201spider}}
\end{figure*}

\begin{figure*}[tb]
    \centering
    \begin{subfigure}[t]{0.24\textwidth}
    \centering
        \includegraphics[width=1\linewidth]{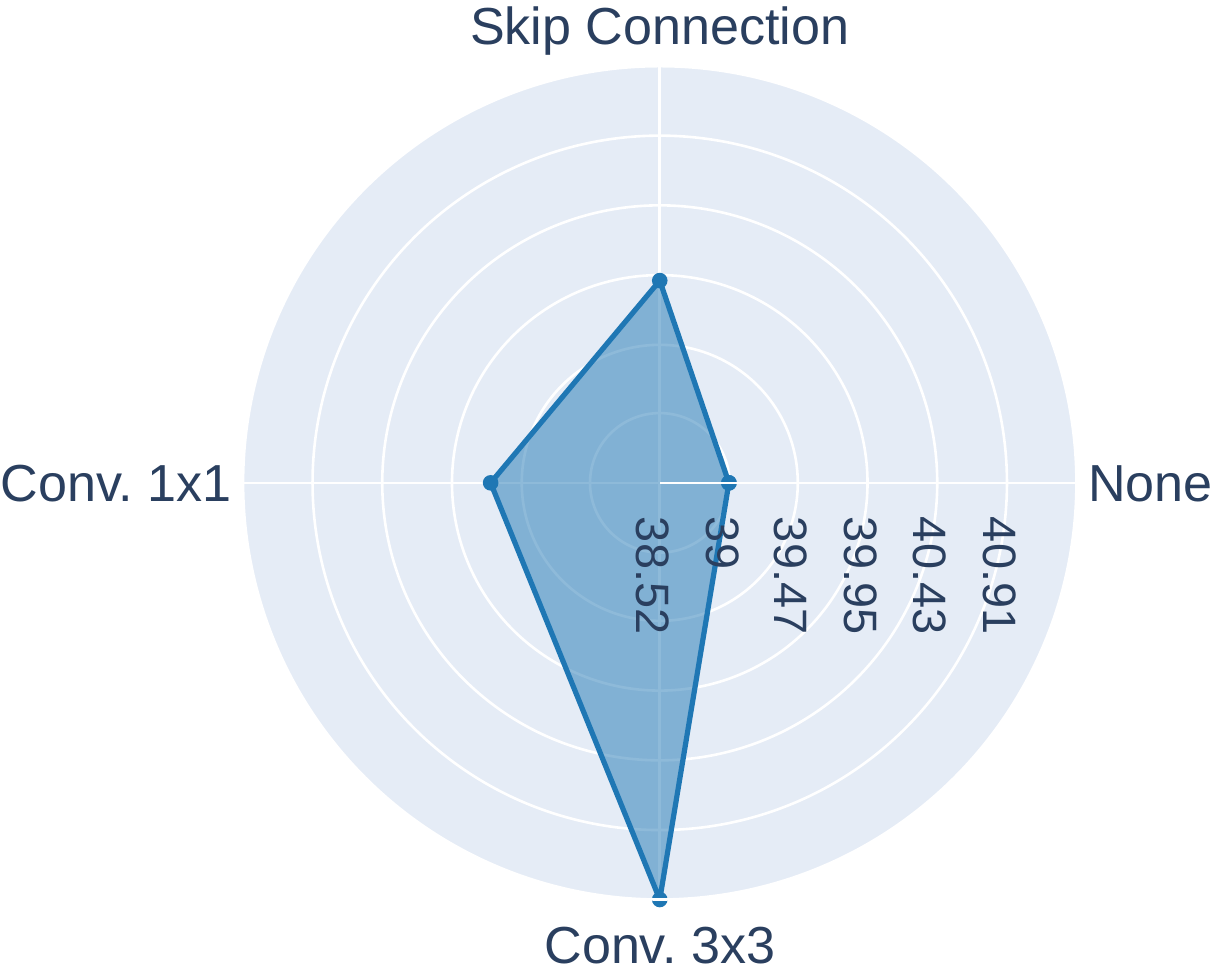}
        \caption{Cls. Object}
        \label{fig:spider-object}
    \end{subfigure}%
    \begin{subfigure}[t]{0.24\textwidth}
    \centering
        \includegraphics[width=1\linewidth]{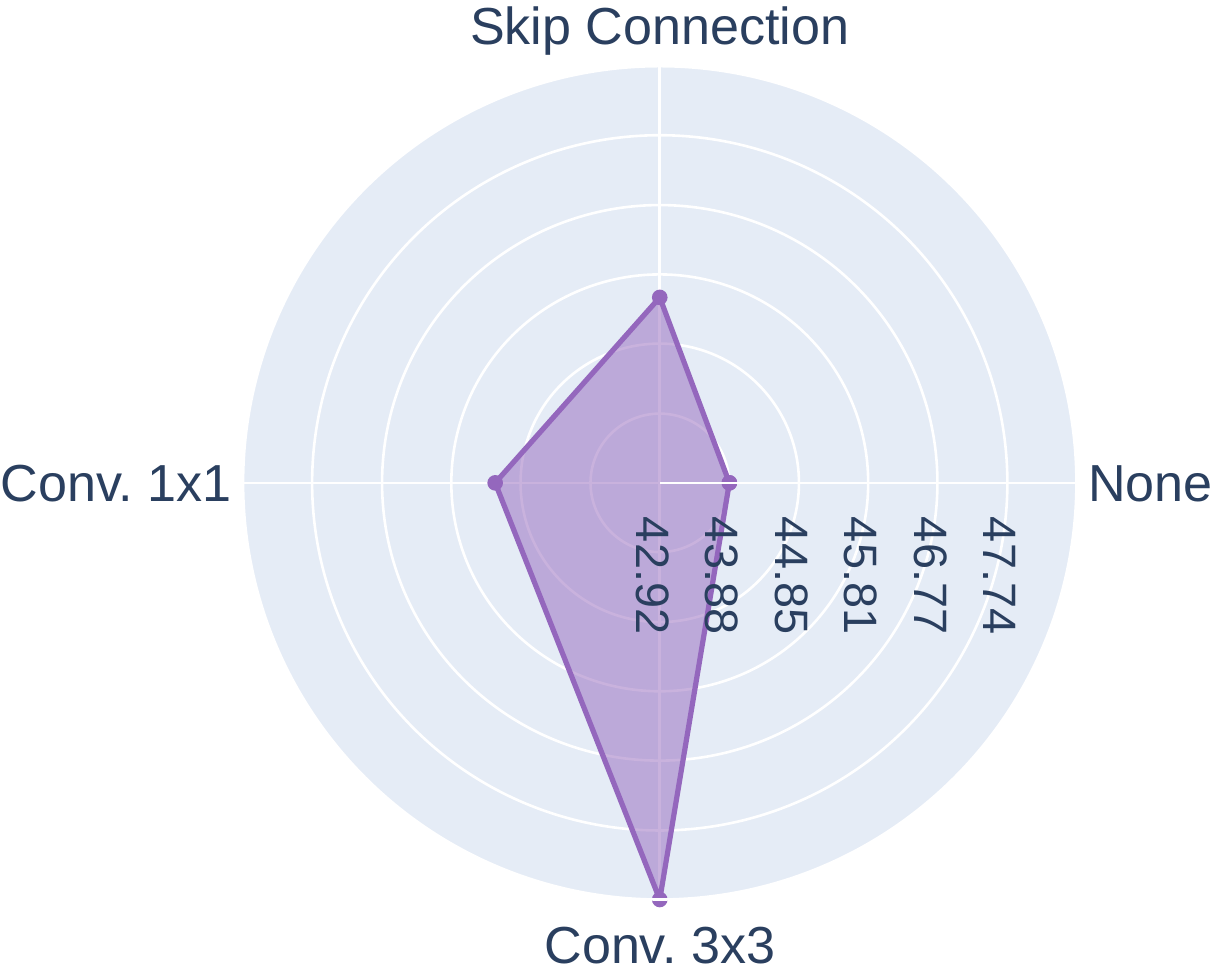}
        \caption{Cls. Scene}
        \label{fig:spider-scene}
    \end{subfigure}
        \begin{subfigure}[t]{0.24\textwidth}
    \centering
        \includegraphics[width=1\linewidth]{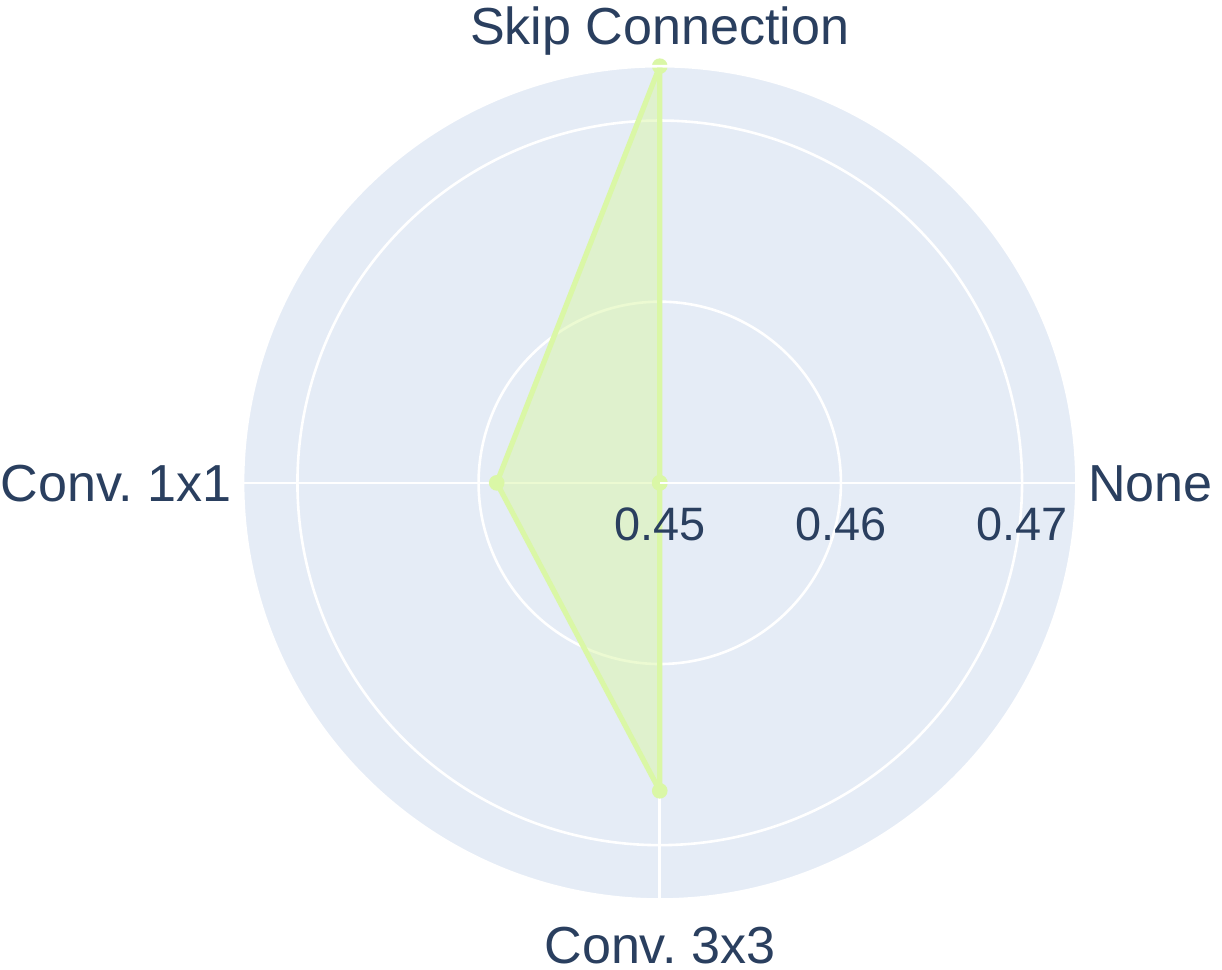}
        \caption{Autoencoding}
        \label{fig:spider-autoenc}
    \end{subfigure}
    \begin{subfigure}[t]{0.24\textwidth}
    \centering
        \includegraphics[width=1\linewidth]{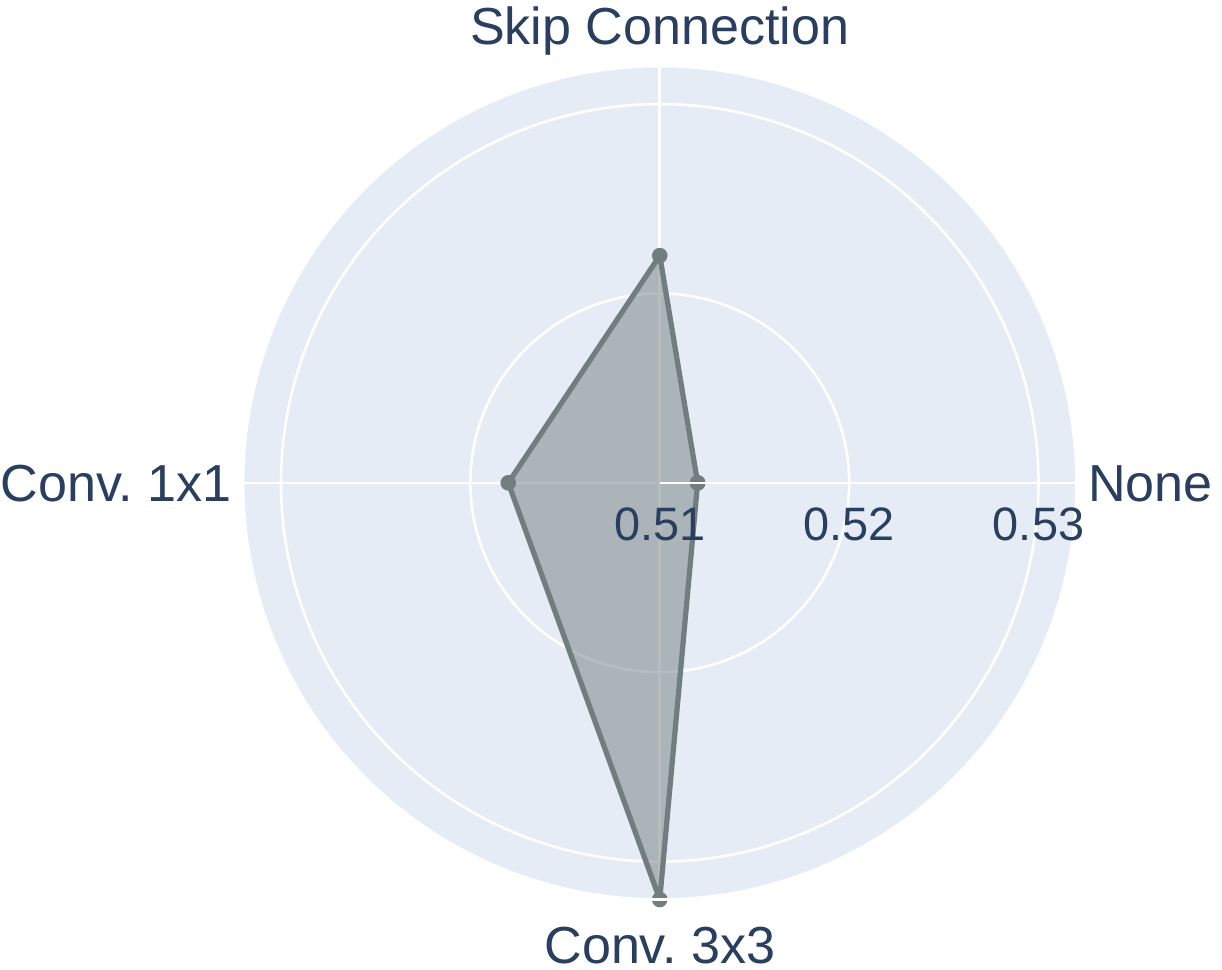}
        \caption{Surf. Normal}
        \label{fig:spider-normal}
    \end{subfigure}%
    \\[2pt]
    \begin{subfigure}[t]{0.24\textwidth}
    \centering
        \includegraphics[width=1\linewidth]{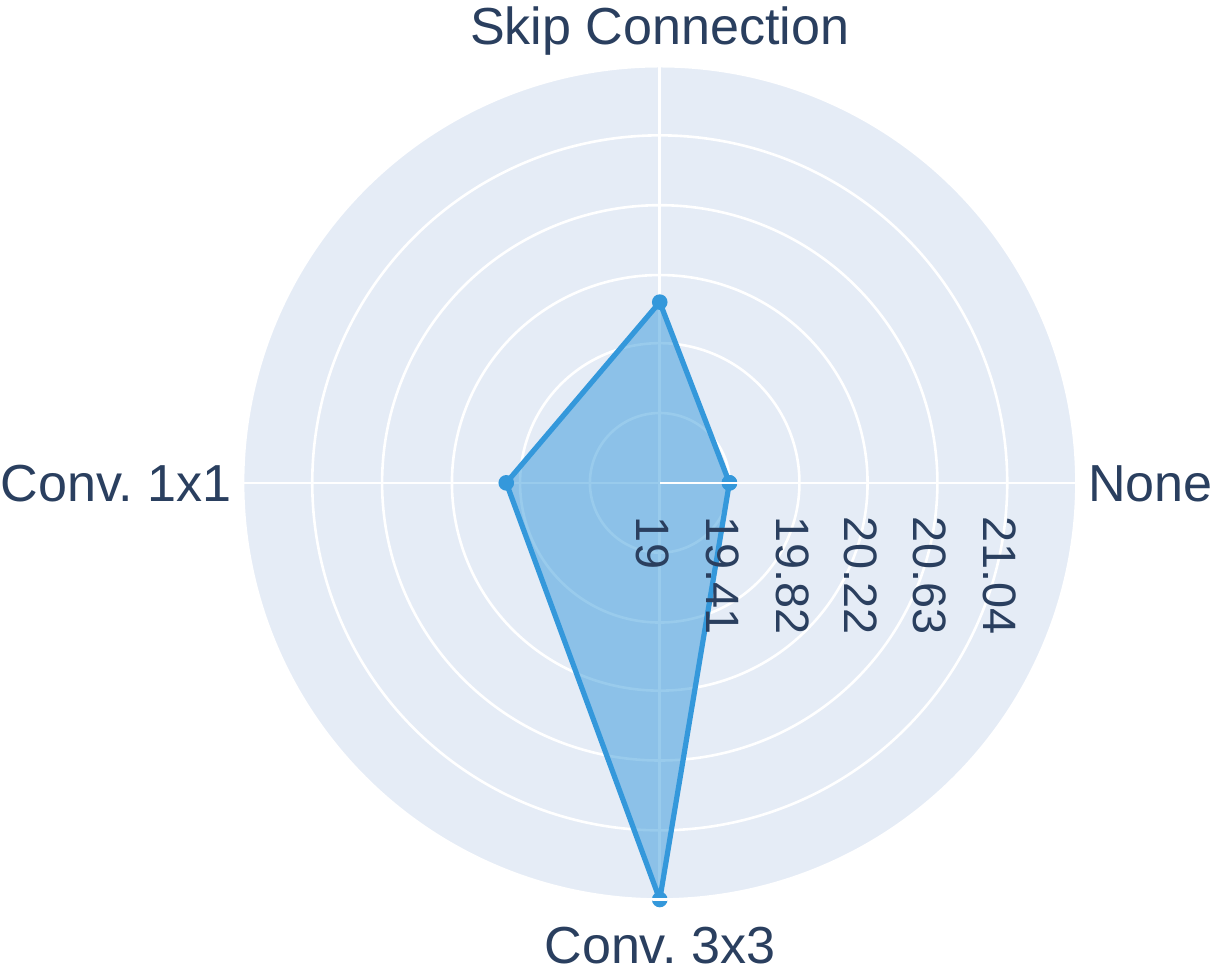}
        \caption{Sem. Segment.}
        \label{fig:spider-semseg}
    \end{subfigure}
    \begin{subfigure}[t]{0.24\textwidth}
    \centering
        \includegraphics[width=1\linewidth]{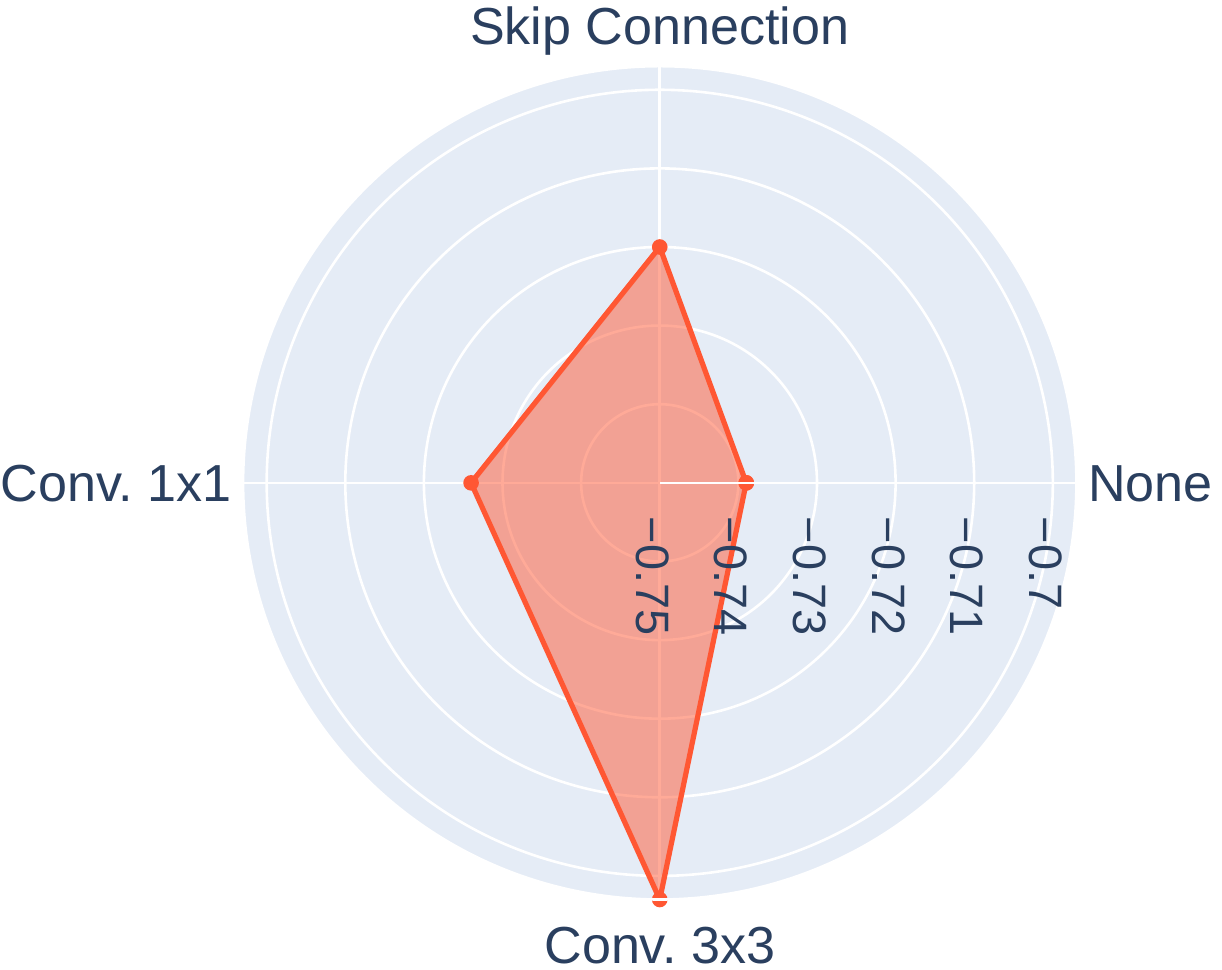}
        \caption{Room Layout}
        \label{fig:spider-layout}
    \end{subfigure}
    \begin{subfigure}[t]{0.24\textwidth}
        \centering
        \includegraphics[width=1\linewidth]{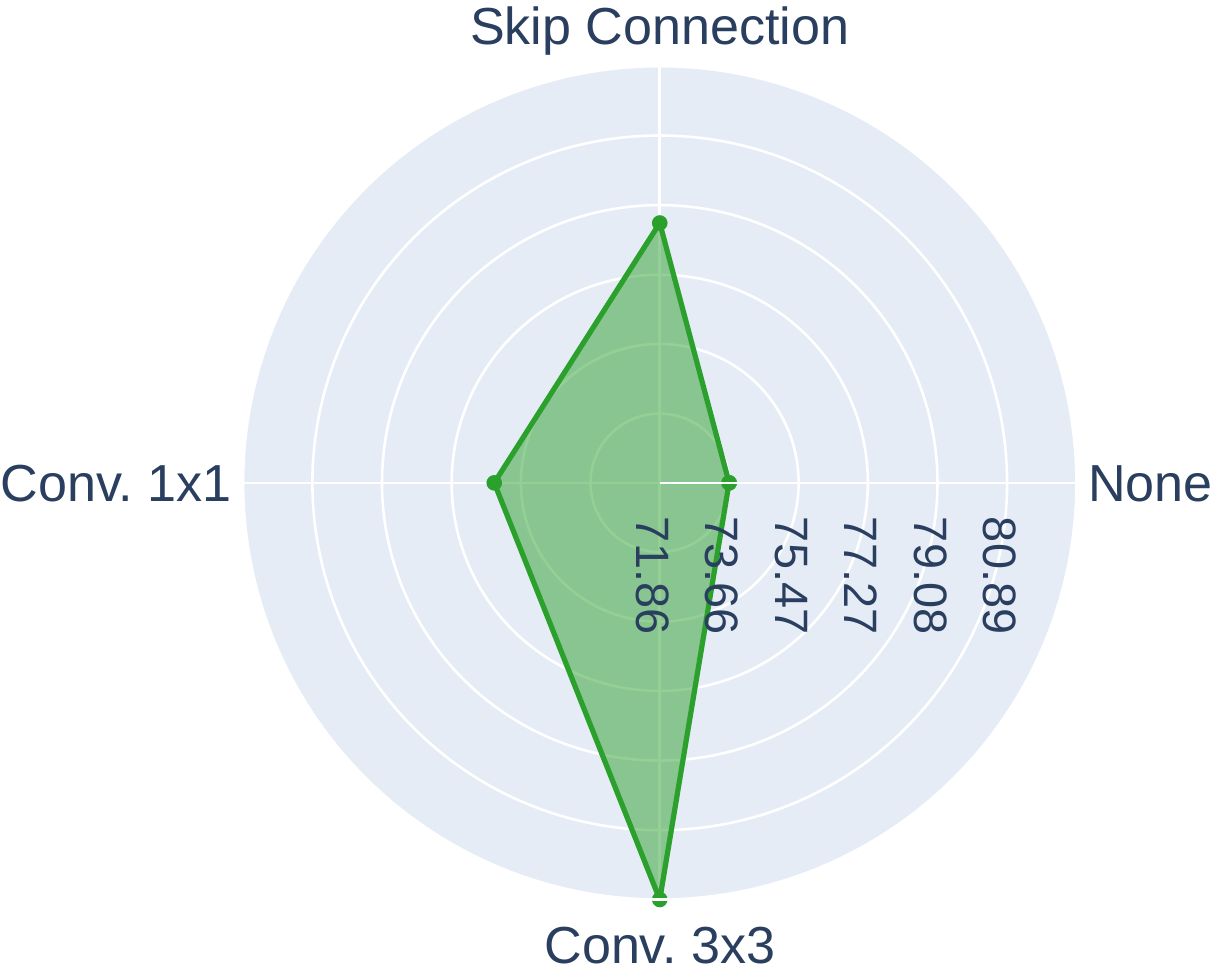}
        \caption{Jigsaw}
        \label{fig:spider-jigsaw}
    \end{subfigure}
  \caption{Mean performance of all the architectures that have at least one operation of a given type in TransNAS-Bench-101. Architectures with at least 1 convolutional layer show higher mean performance in all data sets except in Autoencoding, where $SC$ has the best mean performance.\label{fig:spidertransnas101}}
  \vspace{-1em}
\end{figure*}

% spider  %
%%%%%%%%%%%

\subsubsection{Operation Presence}
Based on the architecture's distribution, we now focus on evaluating the impact of each operation on the final architecture accuracy. First, we look at the mean accuracy (\%) of all architectures that contain at least 1 operation of a given type. Practically, this means that we evaluated all architectures that contain 1 or more $C_{3\times3}$ independently of the rest, and so on and so forth for all operations. What we found, is that on all benchmarks and all data sets, convolutional layers yield the best mean performances, with the exception of the TransNAS-Bench-101 Autoencoding task, where skip connections have the best mean results. Radar graphs are depicted in Figs. \ref{fig:nb101spider}, \ref{fig:nb201spider} and \ref{fig:spidertransnas101} for NAS-Bench-101, NAS-Bench-201 and TransNas-Bench-101, respectively. We justify the importance of convolutional layers due to the fact that cells were carefully designed and have fixed operations that perform pooling and residual connections, which benefit from receiving rich feature maps that can be obtained by performing convolutional operations. On the opposite spectrum, architectures with the operation $None$ have the worst results, which indicates that all edges and nodes on a small cell are important and benefit from having an operation that computes features. 

\subsubsection{Operation Occurrence}
\begin{figure*}[tb]
\centering
\includegraphics[width=0.95\textwidth]{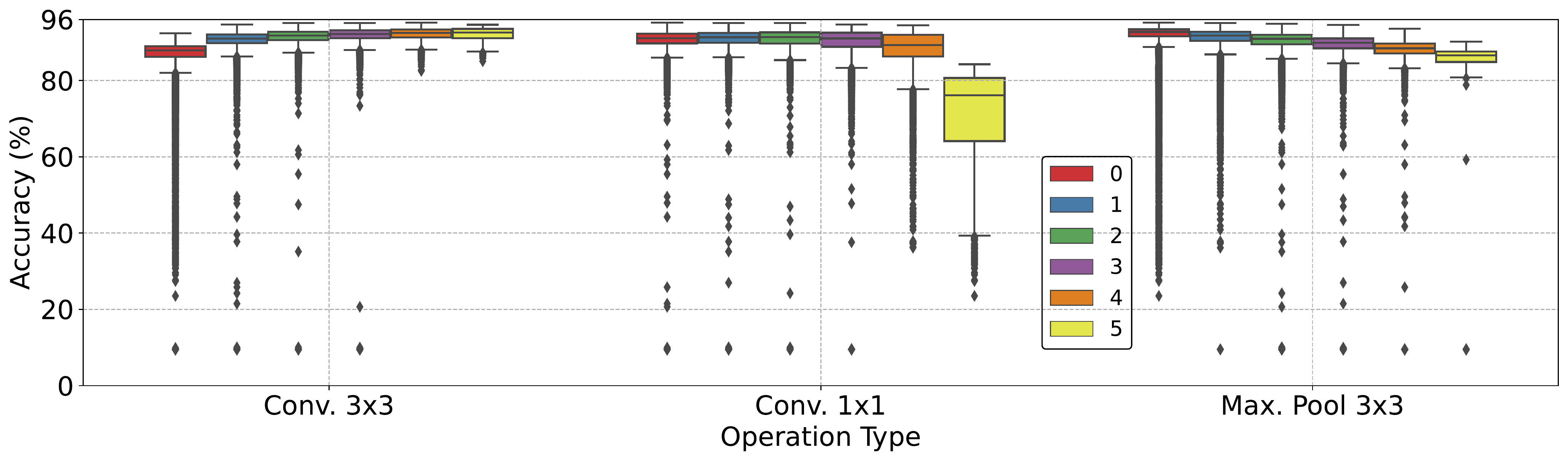}
\caption{Boxplot analysis of the mean accuracy (\%) of all architectures based on the number of occurrences of the different operations on a cell. nas-bench-101.\label{fig:nb101boxplots}}
\end{figure*}

\begin{figure*}[tb]
\centering
\includegraphics[width=0.9\textwidth]{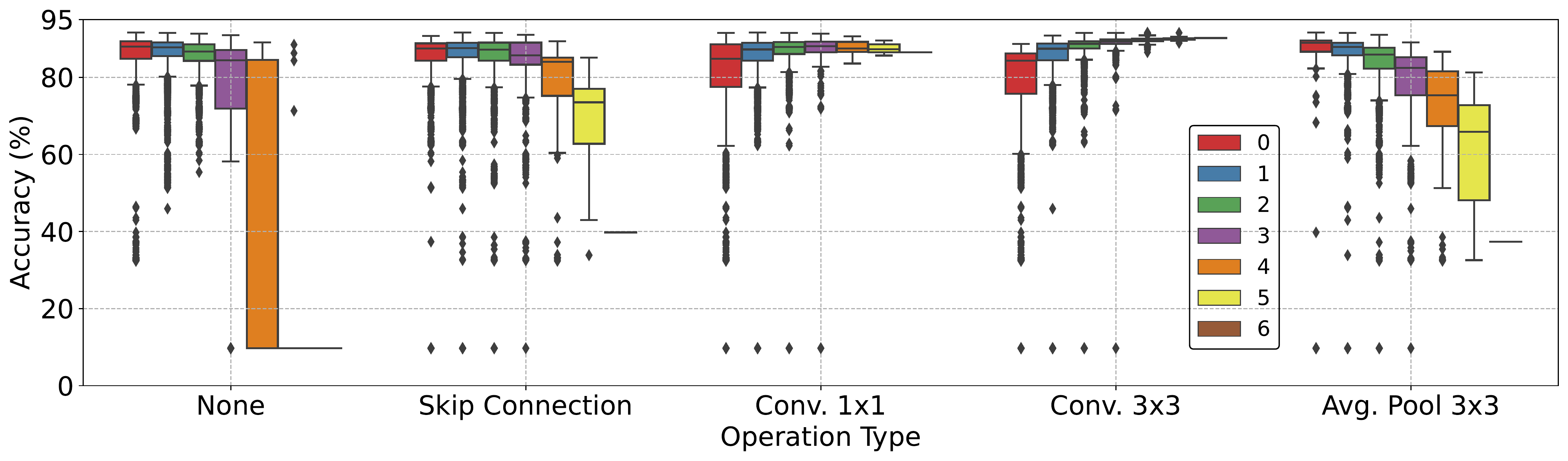}
\includegraphics[width=0.9\textwidth]{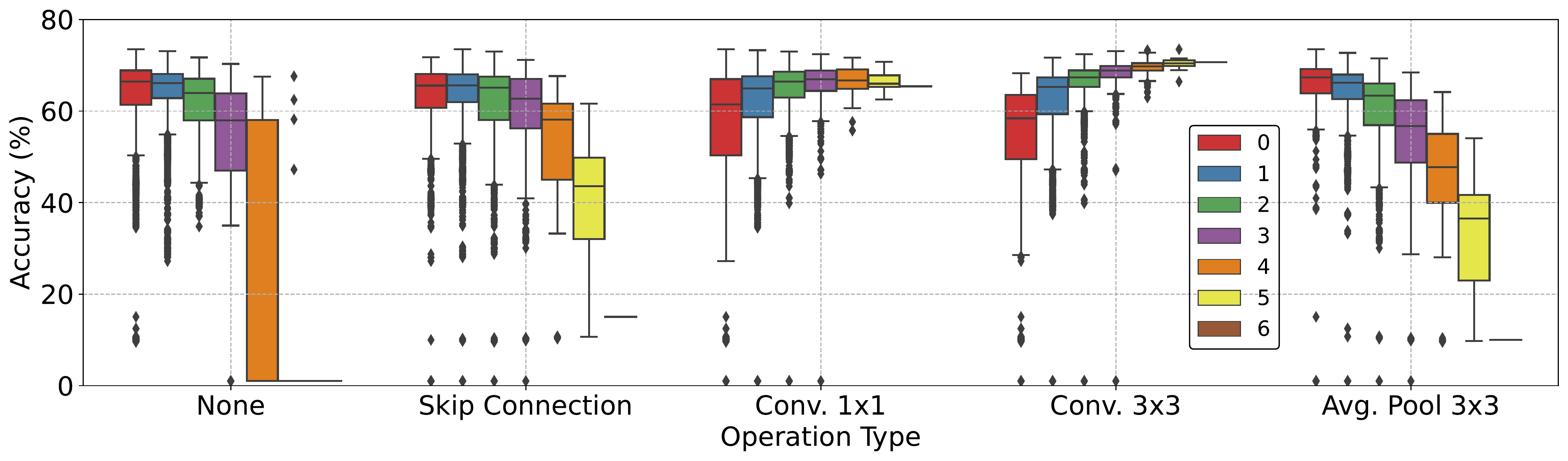}
\includegraphics[width=0.9\textwidth]{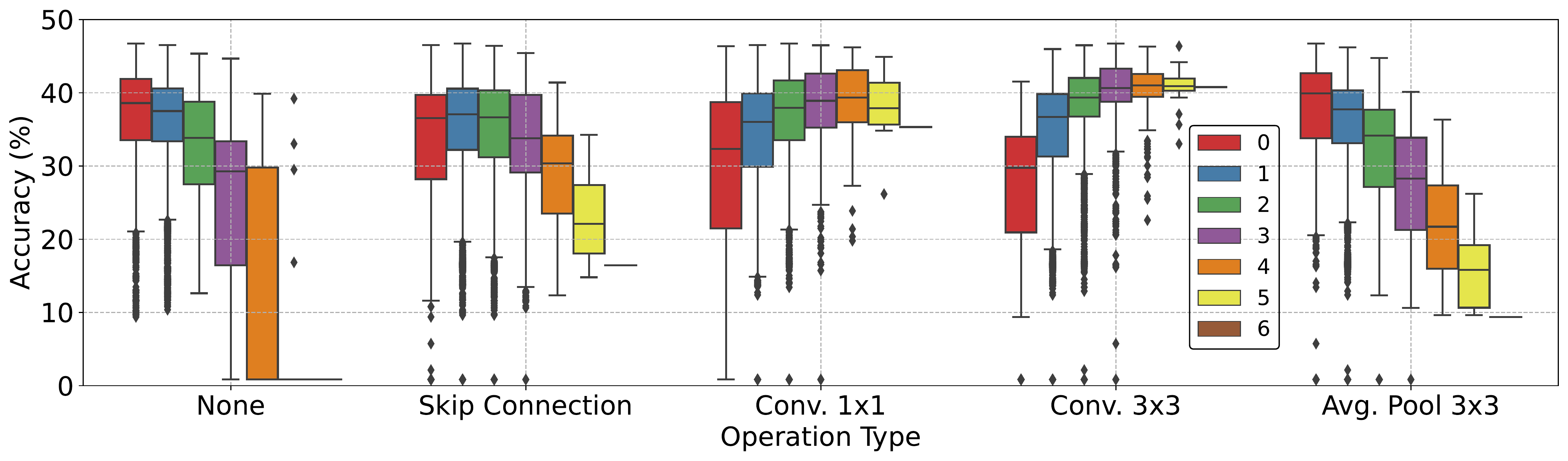}
\caption{Boxplot analysis of the mean accuracy (\%) of all architectures based on the number of occurrences of the different operations on a cell. The top figure presents results on CIFAR10, the middle one on C100 and the bottom one on ImageNet16-120.\label{fig:nb201boxplots}}
\vspace{-1em}
\end{figure*}

Based on the mean performance of architectures that rely on the presence of different operations, we now evaluated how architectures behave based on the presence of different occurrences of a given operation. Practically, in NAS-Bench-101 we evaluated architectures based on a cell having 0 to 5 occurrences of a given operation, and from 0 to 6 on NAS-Bench-201 and TransNas-Bench-101, as it is the maximum number of operations that each cell accommodates. Boxplot graphs shown in Figs. \ref{fig:nb101boxplots}, \ref{fig:nb201boxplots} and \ref{fig:tn101boxplots} depict the minimum, first quartile, median, third quartile, maximum, and outlier values for all possible occurrences of all possible operations for NAS-Bench-101, NAS-Bench-201, and TransNAS-Bench-101 respectively. The results corroborate the hypothesis that convolutional layers yield better cells, as increasing the number of occurrences of such operations consistently results in increases in mean performance. In all benchmarks, mean performances show that $C_{3\times3}$ is the optimal layer, followed by $C_{1\times1}$, and when present, $SC$. More, increasing occurrences of $None$ or pooling operations have a negative impact on the architecture's performance. This negative impact on the performance further corroborates the hypothesis that the structure of the cells is dependent on operations that are capable of creating rich features and that the fixed outer-skeleton and pre-defined operations perform enough data normalization for the data sets analyzed. These results indicate that NAS methods that exhaustively search only for convolutional layers will yield satisfactory results. Therefore, researchers should promote the indication of the performance of NAS methods over time, and if possible, representations of the designed cells, therefore promoting discussion if the method is indeed capable of searching or quickly falls into looking only for convolutional layers.

%%%%%%%%%%%
% boxplot %

\begin{figure*}[!tb]
\centering
\includegraphics[width=0.85\textwidth]{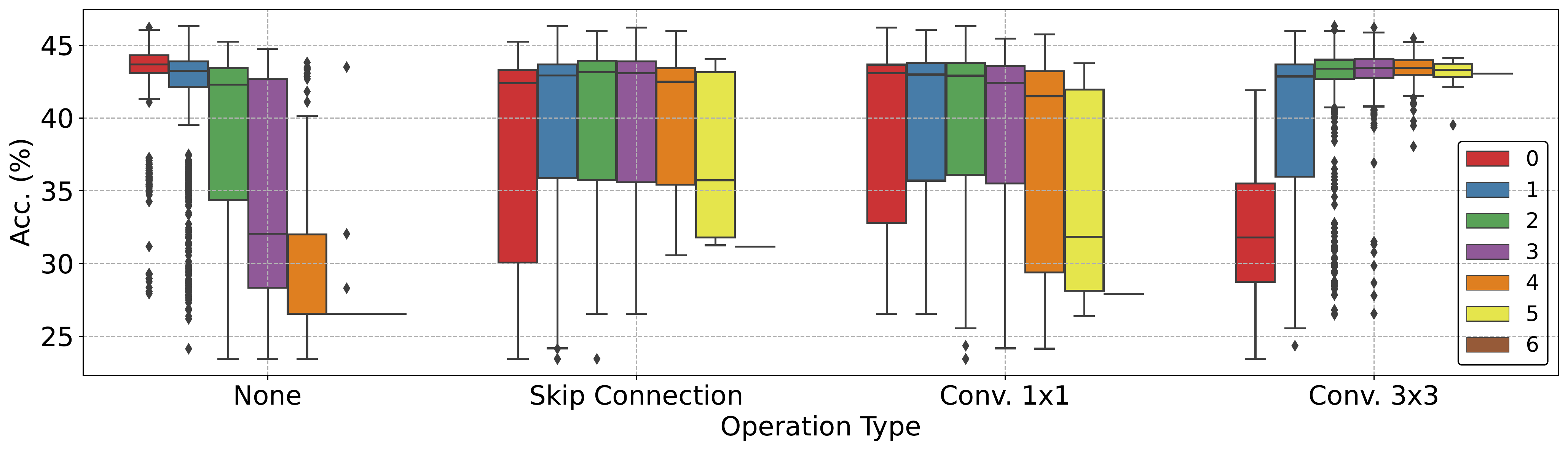}
\includegraphics[width=0.85\textwidth]{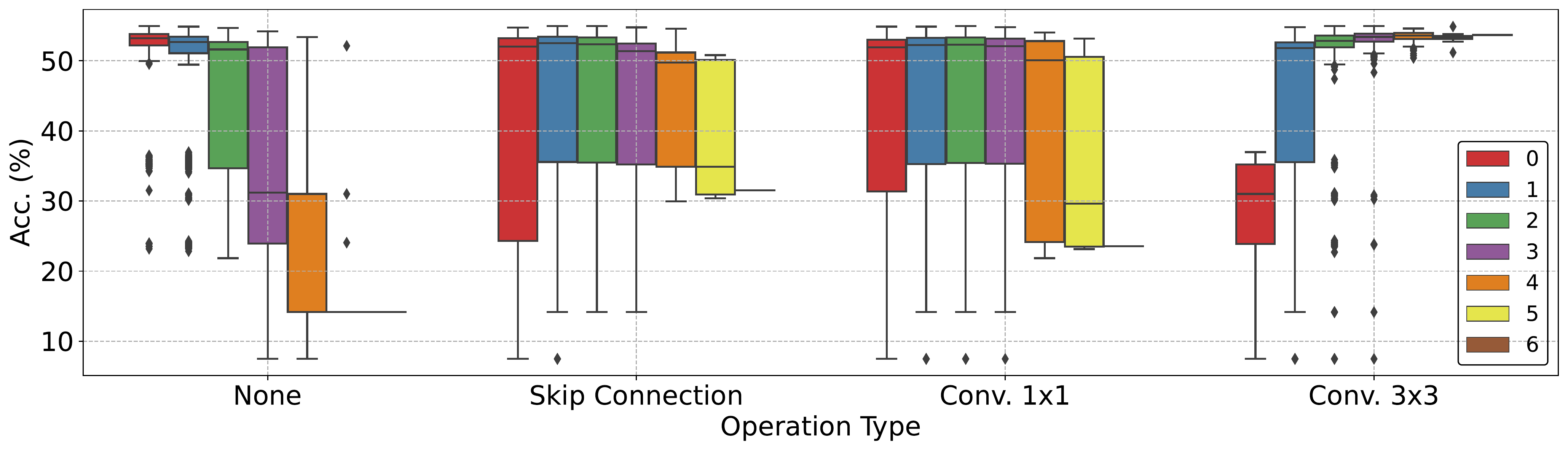}
\includegraphics[width=0.85\textwidth]{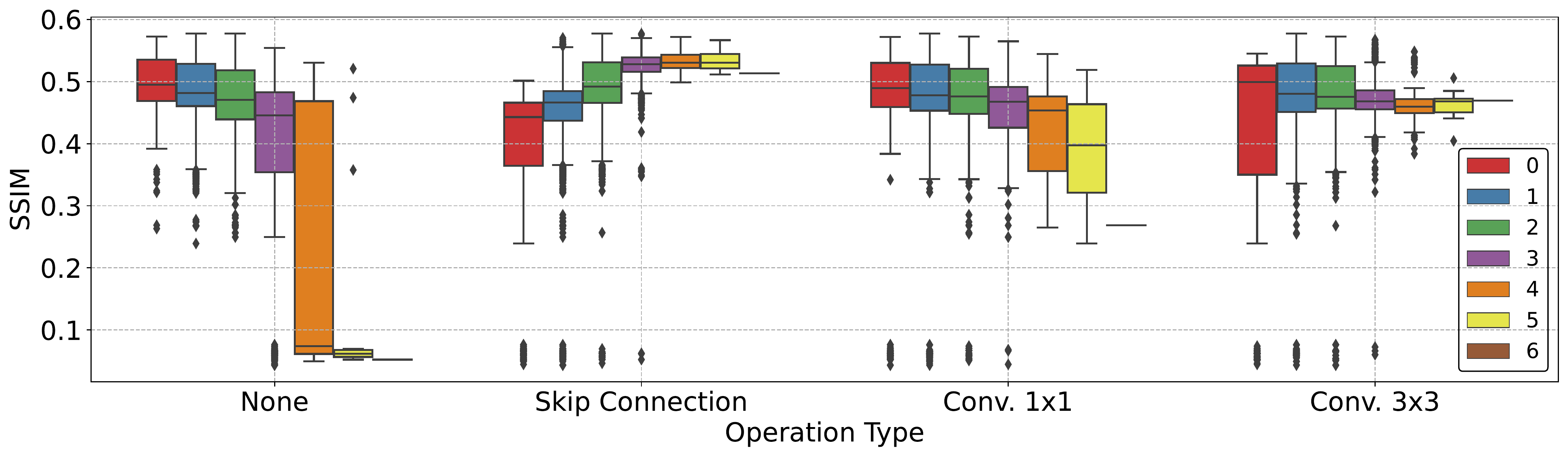}
\includegraphics[width=0.85\textwidth]{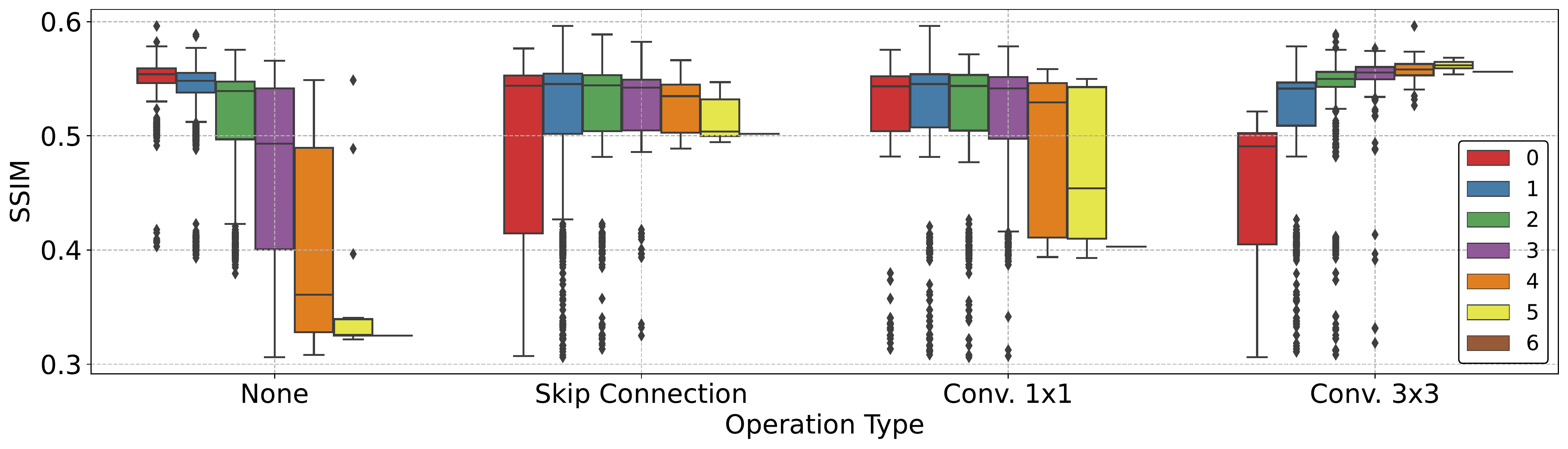}
\includegraphics[width=0.85\textwidth]{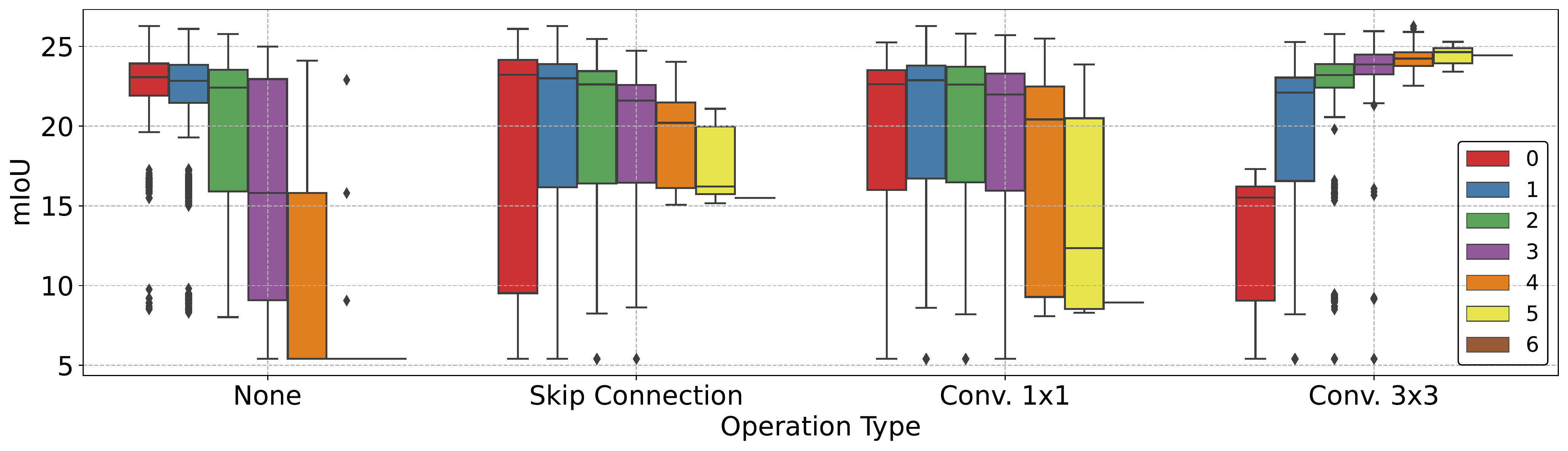}
\end{figure*}
\begin{figure*}[tb]
\ContinuedFloat
\centering
\includegraphics[width=0.85\textwidth]{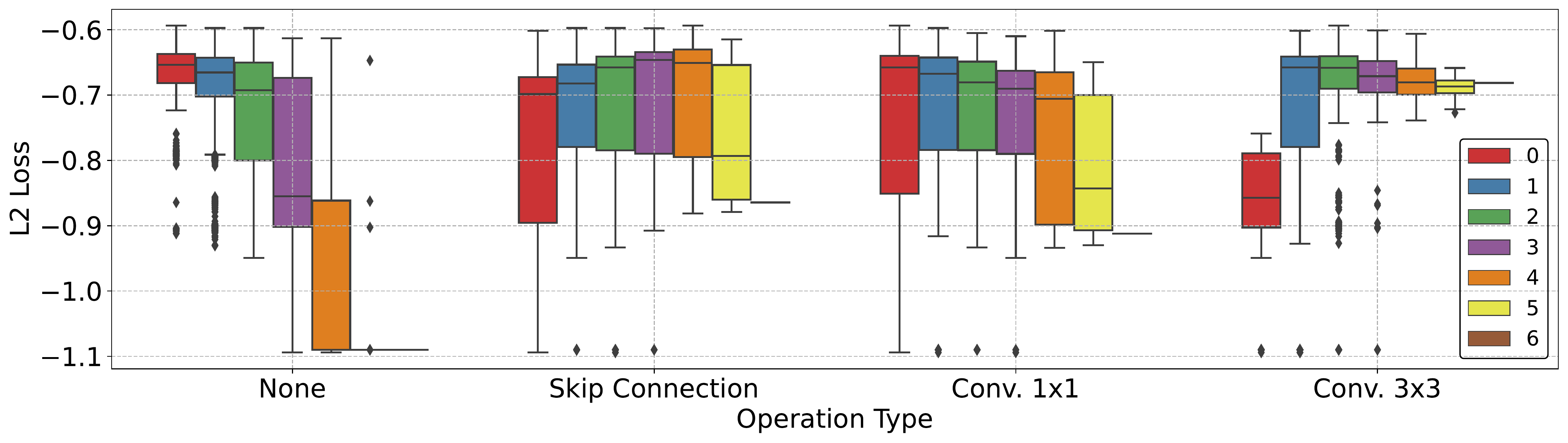}
\includegraphics[width=0.85\textwidth]{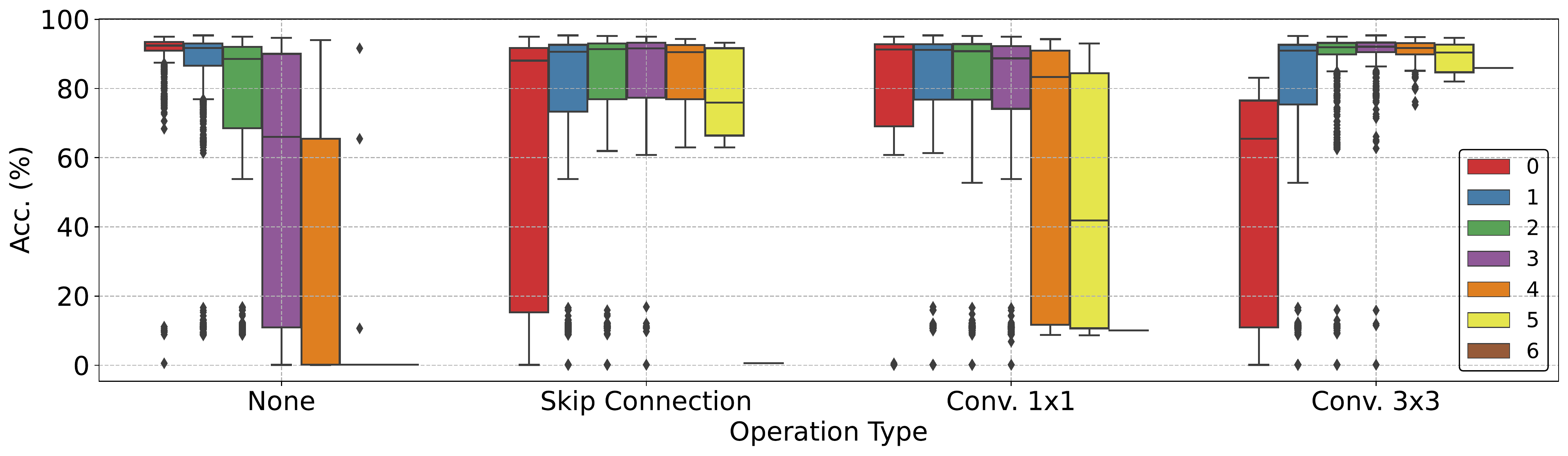}
\caption{Boxplot analysis of the mean performance of all architectures based on the number of occurrences of the different operations on a cell in TransNAS-Bench-101. Tasks from top to bottom: object classification, scene classification, autoenconding, surface normal, semantic segmentation, room layout and jigsaw.\label{fig:tn101boxplots}}
\vspace{-0.5em}
\end{figure*}

% boxplot %
%%%%%%%%%%%

\subsubsection{Operation Positioning}
Given that convolutional layers, especially the ones with larger kernel sizes, lead on average to better results, we now focus on evaluating the performance of the architectures based on having an operation on specific edge positions. For this, we show the mean performance and standard deviation of all architectures that contain an operation on a given edge position $i$, for $i=1,...,I$, where $I$ is the maximum number of edges in a given benchmark. The rationale behind this evaluation is to further corroborate that the pre-defined structure of the cells and the outer-skeletons have a significant impact on the decisions that a NAS method performs while searching. NAS-Bench-101 allows operations in 5 out of the 7 positions, where the remaining 2 are reserved for input and output operations. The results in Table \ref{tab:nb101op_position} show that architectures with $C_{3\times3}$ in any position yield the highest mean validation accuracies, followed by $C_{1\times1}$ with an exception for the second position, where $MP_{3\times3}$ has a higher mean validation accuracy, which is justified by the fact that the second position is the first to receive the input values from previous cells, thus operations that reduce the complexity of the input feature maps can have a direct impact in the final performance. Results for NAS-Bench-201, shown in Table \ref{tab:nb201op_position}, further promote that convolutional layers are preferable, yielding the best results in all possible positions for all data sets. Results also show that $SC$ layers show improved importance in positions that connect earlier nodes to later nodes in the cell, which benefit from residual connections. Also, $None$ operations tend to be better in the middle of the cell, and $AP_{3\times3}$ stays consistent throughout. These findings suggest that convolutions are preferable throughout the cell structure, but other operations, such as $SC$, can have importance if placed strategically in specific positions of the cell. Similarly, in all data sets of TransNAS-Bench-101 with exception of Autoencoding, $C_{3\times3}$ yields the best results, whereas in Autoencoding, $SC$ is the best operation to use in all positions (Table \ref{tab:tbnas101_op_position}). $SC$ and $C_{1\times1}$ behave similarly in TransNAS-Bench-101, achieving better results earlier in the cell, whereas $C_{3\times3}$ consistently performs throughout all possible positions in the cell. $None$ operation has the worst mean performances in all data sets, further promoting the idea that all cell operations have an impact on the final architecture's performance. Results in all data sets show that convolutional layers have the most impact on an architecture's performance and that in some specific cases, the presence of a skip connection interleaved with convolutional layers can be fruitful. 

\subsubsection{Operation Combination}
We further extend the evaluation of the operation's importance by looking at two consecutive combinations of operations, as subsequent operations in the cell's structure. This analysis evaluates not only the importance of a single operation, but how they behave if combined. For this, we looked at all possible combinations of two operations and calculated the mean performance and standard deviation of all architectures that contain such combinations. Results for NAS-Bench-101, shown in Table \ref{tab:nb101opcombs}, further promote that $C_{3\times3}$ operations yield the best results, even when combined with any other operation. Furthermore, $C_{1\times1}$ follows the pattern, by achieving the next higher mean performances when combined with any convolutional operation, followed by $MP_{3\times3}$. In NAS-Bench-201, we see the same behaviour. Results in Table \ref{tab:nb201opcombs} show that combining convolutional layers results, on average, in the best performant architectures. From the table it is also possible to see that combining operations with $None$ and $AP_{3\times3}$ leads to worse performant architectures, and that a $SC$ combined with convolutional layers, achieves results close to the best combination, further suggesting that in some specific cases, $SC$ layers associated with convolutional layers result in good cell designs. Finally, the evaluation of the operation combinations on TransNAS-Bench-101 is shown in Table \ref{tab:tnas101opcombs}, where in all tasks, the combination of two $C_{3\times3}$ achieves the highest mean performances, with exception for Autoencoding, where combining two $SC$ has the best results. From the table, it is possible to see that for all tasks, with exception for Autoencoding, combining an operation with $None$ always yields the worst results, whereas combining any operation with $C_{3\times3}$ consistently achieves better results when compared with combining with other operation. Furthermore, the results show that in TransNAS-Bench-201, $C_{1\times1}$ and $SC$ operations have similar results when combined with other operations, further suggesting that operations that create rich features are preferred, but when not present, operations that help reducing the complexity of the feature maps are good trade-offs, as they achieve good results by having a significant impact on the flatness of the loss landscape, as shown in previous works \cite{DBLP:conf/nips/Li0TSG18}. Results across all benchmarks corroborate the hypothesis that the operation pool and the structure of the cells (search space) have a high influence on the quality of the final architectures, which is shown by the consistent results that convolutional and skip connection layers have in all three benchmarks.

%%%%%%%%%%%
% op_pos  %

\begin{table}[tb]
  \caption{Mean validation accuracy (\%) and standard deviation of all architectures that contain an operation on a specific edge in NAS-Bench-101.}
  \label{tab:nb101op_position}
  \centering
  \resizebox{\columnwidth}{!}{%
  \begin{tabular}{@{}llllll@{}} \toprule
  \multirow{2}{*}{\textbf{Operation}}                & \multicolumn{5}{c}{\textbf{Position}} \\ \cline{2-6}
       & \multicolumn{1}{c}{\textbf{2}} & \multicolumn{1}{c}{\textbf{3}} & \multicolumn{1}{c}{\textbf{4}} & \multicolumn{1}{c}{\textbf{5}} & \multicolumn{1}{c}{\textbf{6}}   \\ \midrule
       \multicolumn{6}{c}{\textbf{CIFAR-10}} \\ \midrule
    $C_{1\times1}$    & 89.64 $\pm$ 7.67        & 89.99 $\pm$ 6.36        & 90.13 $\pm$ 5.71       & 90.30 $\pm$ 5.00        & 90.58 $\pm$ 4.43             \\
    $C_{3\times3}$    & \textbf{90.92} $\pm$ 7.92        & \textbf{91.19} $\pm$ 5.63       & \textbf{91.27} $\pm$ 4.82       & \textbf{91.45} $\pm$ 3.89        & \textbf{91.70} $\pm$ 3.13        \\ 
    $MP_{3\times3}$   & 90.18 $\pm$ 4.96       & 89.56 $\pm$ 8.62       & 89.34 $\pm$ 9.48       & 89.01 $\pm$ 10.22        & 88.22 $\pm$ 10.66          \\
    % first position is input, last is output
    \bottomrule
      \end{tabular}
    }
\end{table}

\begin{table}[tb]
  \caption{Mean validation accuracy (\%) and standard deviation of all architectures that contain an operation on a specific edge in NAS-Bench-201.}
  \label{tab:nb201op_position}
  \centering
  \resizebox{\columnwidth}{!}{%
  \begin{tabular}{@{}lllllll@{}} \toprule
  \multirow{2}{*}{\textbf{Operation}}                & \multicolumn{6}{c}{\textbf{Position}} \\ \cline{2-7}
       & \multicolumn{1}{c}{\textbf{1}} & \multicolumn{1}{c}{\textbf{2}} & \multicolumn{1}{c}{\textbf{3}} & \multicolumn{1}{c}{\textbf{4}} & \multicolumn{1}{c}{\textbf{5}} & \multicolumn{1}{c}{\textbf{6}} \\ \midrule
       \multicolumn{7}{c}{\textbf{CIFAR-10}} \\ \midrule
    $None$            & 79.02 $\pm$ 20.17       & 81.02 $\pm$ 17.68       & 83.40 $\pm$ 13.34       & 75.04 $\pm$ 24.02       & 81.18 $\pm$ 17.68       & 78.99 $\pm$ 20.17       \\
    $C_{1\times1}$    & 86.62 $\pm$ 7.93        & 85.59 $\pm$ 9.93        & 84.51 $\pm$ 11.87       & 87.33 $\pm$ 2.41        & 85.37 $\pm$ 10.12       & 86.34 $\pm$ 8.24        \\
    $C_{3\times3}$    & \textbf{87.61} $\pm$ {7.98}    & \textbf{86.31} $\pm$ 10.00       & \textbf{84.93} $\pm$ 11.93       & \textbf{88.31} $\pm$ 1.99      & \textbf{85.99} $\pm$ 10.09       & \textbf{87.15} $\pm$ 8.13  \\ 
    $AP_{3\times3}$   & 81.32 $\pm$ 11.08       & 81.91 $\pm$ 12.15       & 82.58 $\pm$ 13.38       & 81.46 $\pm$ 8.74        & 82.32 $\pm$ 12.09       & 82.11 $\pm$ 11.08       \\
    $SC$              & 83.81 $\pm$ 10.59       & 83.54 $\pm$ 11.85       & 82.95 $\pm$ 13.26       & 85.33 $\pm$ 7.38        & 83.50 $\pm$ 11.85       & 83.77 $\pm$ 10.62       \\
    \midrule  
    \multicolumn{7}{c}{\textbf{CIFAR-100}} \\ \midrule
    $None$           & 56.83 $\pm$ 17.16       & 58.90 $\pm$ 15.43       & 61.05 $\pm$ 12.23       & 54.47 $\pm$ 20.13       & 59.03 $\pm$ 15.41       & 56.80 $\pm$ 17.13       \\
    $C_{1\times1}$   & 64.95 $\pm$ 7.53        & 63.51 $\pm$ 9.47        & 62.37 $\pm$ 11.11       & 65.27 $\pm$ 4.30        & 63.34 $\pm$ 9.74        & 64.56 $\pm$ 8.00        \\
    $C_{3\times3}$   & \textbf{66.35} $\pm$ 7.43        & \textbf{64.58} $\pm$ 9.48        & \textbf{63.03} $\pm$ 11.11       & \textbf{66.87} $\pm$ 3.54        & \textbf{64.35} $\pm$ 9.57        & \textbf{65.85} $\pm$ 7.67       \\ 
    $AP_{3\times3}$  & 57.81 $\pm$ 11.64       & 58.82 $\pm$ 12.23       & 59.76 $\pm$ 12.97       & 57.96 $\pm$ 10.38       & 59.17 $\pm$ 12.16       & 58.61 $\pm$ 11.63       \\
    $SC$             & 60.47 $\pm$ 11.13       & 60.60 $\pm$ 11.91       & 60.20 $\pm$ 12.84       & 61.84 $\pm$ 9.40        & 60.52 $\pm$ 11.89       & 60.58 $\pm$ 11.24       \\
    \midrule   
    \multicolumn{7}{c}{\textbf{ImageNet16-120}} \\ \midrule
    $None$            & 29.77 $\pm$ 11.43       & 31.47 $\pm$ 10.86       & 33.22 $\pm$ 9.43        & 27.42 $\pm$ 12.38       & 31.74 $\pm$ 10.81       & 29.75 $\pm$ 11.38       \\
    $C_{1\times1}$    & 37.32 $\pm$ 6.65        & 35.75 $\pm$ 7.70        & 34.59 $\pm$ 8.87        & 37.83 $\pm$ 4.79        & 35.46 $\pm$ 8.18        & 36.54 $\pm$ 7.43        \\
    $C_{3\times3}$    & \textbf{38.29} $\pm$ 6.68        & \textbf{36.77} $\pm$ 7.75        & \textbf{35.38} $\pm$ 8.98        & \textbf{39.19} $\pm$ 2.98        & \textbf{36.36} $\pm$ 8.13        & \textbf{37.76} $\pm$ 7.31        \\ 
    $AP_{3\times3}$   & 30.10 $\pm$ 8.97        & 31.26 $\pm$ 9.33        & 32.62 $\pm$ 9.37        & 29.78 $\pm$ 8.93        & 31.82 $\pm$ 9.05        & 31.17 $\pm$ 8.57        \\
    $SC$              & 33.77 $\pm$ 8.17        & 33.70 $\pm$ 8.84        & 33.15 $\pm$ 9.25        & 34.74 $\pm$ 7.69        & 33.59 $\pm$ 8.82        & 33.75 $\pm$ 8.30        \\
    \bottomrule
    \end{tabular}
    }
    \vspace{-2em}
\end{table}

\begin{table}[tb]
  \caption{Mean performance and standard deviation of all architectures that contain an operation on a specific edge in TransNAS-Bench-101.}
  \label{tab:tbnas101_op_position}
  \centering
  \resizebox{0.95\columnwidth}{!}{%
  \begin{tabular}{@{}llllllll@{}} \toprule
  \multirow{2}{*}{\textbf{Operation}}                & \multicolumn{6}{c}{\textbf{Position}} \\ \cline{2-7}
       & \multicolumn{1}{c}{\textbf{1}} & \multicolumn{1}{c}{\textbf{2}} & \multicolumn{1}{c}{\textbf{3}} & \multicolumn{1}{c}{\textbf{4}} & \multicolumn{1}{c}{\textbf{5}} & \multicolumn{1}{c}{\textbf{6}} \\ 
    \midrule  \multicolumn{7}{c}{\textbf{Cls. Object} (Acc. (\%) $\uparrow$)} \\ \midrule
    $None$            & 36.45 $\pm$ 6.89       & 37.95 $\pm$ 6.67       & 39.07 $\pm$ 6.24       & 37.10 $\pm$ 6.88       & 38.05 $\pm$ 6.66       & 36.50 $\pm$ 6.91       \\
    $SC$              & 40.36 $\pm$ 4.87       & 39.95 $\pm$ 5.33       & 39.57 $\pm$ 5.93       & 39.87 $\pm$ 5.36       & 39.67 $\pm$ 5.90       & 40.10 $\pm$ 5.60       \\
    $C_{1\times1}$    & 39.47 $\pm$ 5.99       & 39.17 $\pm$ 6.15       & 39.66 $\pm$ 5.90       & 38.36 $\pm$ 6.38       & 39.66 $\pm$ 5.85       & 40.06 $\pm$ 5.56       \\
    $C_{3\times3}$    & \textbf{42.57} $\pm$ 3.73       & \textbf{41.77} $\pm$ 4.68       & \textbf{40.54} $\pm$ 5.51       & \textbf{43.52} $\pm$ 0.64       & \textbf{41.46} $\pm$ 4.60       & \textbf{42.19} $\pm$ 3.69       \\ 
    \midrule  \multicolumn{7}{c}{\textbf{Cls. Scene} (Acc. (\%) $\uparrow$)} \\ \midrule
    $None$            & 38.52 $\pm$ 14.57      & 41.67 $\pm$ 14.12       & 43.84 $\pm$ 12.68       & 40.30 $\pm$ 15.49     & 41.97 $\pm$ 13.81       & 38.69 $\pm$ 14.36       \\
    $SC$              & 46.27 $\pm$ 9.93       & 45.43 $\pm$ 10.86       & 44.92 $\pm$ 12.08       & 44.89 $\pm$ 9.79      & 44.74 $\pm$ 11.75       & 45.47 $\pm$ 11.16       \\
    $C_{1\times1}$    & 44.86 $\pm$ 11.85      & 44.33 $\pm$ 12.28       & 45.05 $\pm$ 12.16       & 42.79 $\pm$ 12.83     & 45.06 $\pm$ 11.90       & 45.97 $\pm$ 11.23       \\
    $C_{3\times3}$    & \textbf{51.26} $\pm$ 7.54       & \textbf{49.49} $\pm$ 9.54        & \textbf{47.11} $\pm$ 11.43       & \textbf{52.93} $\pm$ 0.84      & \textbf{49.14} $\pm$ 9.71        & \textbf{50.78} $\pm$ 7.73       \\ 
    \midrule  \multicolumn{7}{c}{\textbf{Autoencoding} (SSIM $\uparrow$)} \\ \midrule
    %$None$            & 0.420 $\pm$ 0.14       & 0.433 $\pm$ 0.128       & 0.455 $\pm$ 0.105       & 0.381 $\pm$ 0.153       & 0.434 $\pm$ 0.127       & 0.420 $\pm$ 0.140       \\
    $None$            & 0.42 $\pm$ 0.14       & 0.43 $\pm$ 0.13       & 0.46 $\pm$ 0.11       & 0.38 $\pm$ 0.15       & 0.43 $\pm$ 0.13      & 0.42 $\pm$ 0.14       \\
    %$SC$              & 0.489 $\pm$ 0.073      & 0.482 $\pm$ 0.085       & 0.462 $\pm$ 0.099       & 0.533 $\pm$ 0.014       & 0.477 $\pm$ 0.090       & 0.484 $\pm$ 0.081       \\
    $SC$              & \textbf{0.49} $\pm$ 0.07      & \textbf{0.48} $\pm$ 0.09       & 0.46 $\pm$ 0.10       & \textbf{0.53} $\pm$ 0.01       & \textbf{0.48} $\pm$ 0.09       & \textbf{0.48} $\pm$ 0.08       \\
    %$C_{1\times1}$    & 0.455 $\pm$ 0.086      & 0.453 $\pm$ 0.093       & 0.459 $\pm$ 0.100       & 0.443 $\pm$ 0.064       & 0.460 $\pm$ 0.090       & 0.464 $\pm$ 0.081       \\
    $C_{1\times1}$    & 0.46 $\pm$ 0.09      & 0.45 $\pm$ 0.09       & 0.46 $\pm$ 0.10       & 0.44 $\pm$ 0.06       & 0.46 $\pm$ 0.09       & 0.46 $\pm$ 0.08       \\
    %$C_{3\times3}$    & 0.472 $\pm$ 0.072      & 0.468 $\pm$ 0.084       & 0.460 $\pm$ 0.098       & 0.479 $\pm$ 0.027       & 0.465 $\pm$ 0.084       & 0.468 $\pm$ 0.074       \\ 
    $C_{3\times3}$    & 0.47 $\pm$ 0.07      & 0.47 $\pm$ 0.08       & 0.46 $\pm$ 0.10       & 0.48 $\pm$ 0.03       & 0.47 $\pm$ 0.08       & 0.47 $\pm$ 0.07       \\ 
    \midrule  \multicolumn{7}{c}{\textbf{Surf. Normal} (SSIM $\uparrow$)} \\ \midrule
    %$None$            & 0.486 $\pm$ 0.074      & 0.501 $\pm$ 0.070       & 0.512 $\pm$ 0.062       & 0.488 $\pm$ 0.082       & 0.502 $\pm$ 0.071       & 0.486 $\pm$ 0.075       \\
    %$SC$              & 0.528 $\pm$ 0.041      & 0.524 $\pm$ 0.048       & 0.517 $\pm$ 0.058       & 0.530 $\pm$ 0.027       & 0.518 $\pm$ 0.055       & 0.523 $\pm$ 0.050       \\
    %$C_{1\times1}$    & 0.515 $\pm$ 0.058      & 0.514 $\pm$ 0.060       & 0.519 $\pm$ 0.058       & 0.505 $\pm$ 0.062       & 0.519 $\pm$ 0.056       & 0.523 $\pm$ 0.051       \\
    %$C_{3\times3}$    & 0.545 $\pm$ 0.039      & 0.536 $\pm$ 0.047       & 0.526 $\pm$ 0.055       & 0.552 $\pm$ 0.009       & 0.535 $\pm$ 0.047       & 0.542 $\pm$ 0.038       \\ 
    $None$            & 0.49 $\pm$ 0.07      & 0.50 $\pm$ 0.07       & 0.51 $\pm$ 0.06       & 0.49 $\pm$ 0.08       & 0.50 $\pm$ 0.07       & 0.49 $\pm$ 0.08       \\
    $SC$              & 0.53 $\pm$ 0.04      & 0.52 $\pm$ 0.05       & 0.52 $\pm$ 0.06       & 0.53 $\pm$ 0.03       & 0.52 $\pm$ 0.06       & 0.52 $\pm$ 0.05       \\
    $C_{1\times1}$    & 0.52 $\pm$ 0.06      & 0.51 $\pm$ 0.06       & 0.52 $\pm$ 0.06       & 0.51 $\pm$ 0.06       & 0.52 $\pm$ 0.06       & 0.52 $\pm$ 0.05       \\
    $C_{3\times3}$    & \textbf{0.55} $\pm$ 0.04      & \textbf{0.54} $\pm$ 0.05       & \textbf{0.53} $\pm$ 0.06       & \textbf{0.55} $\pm$ 0.01       & \textbf{0.54} $\pm$ 0.05       & \textbf{0.54} $\pm$ 0.04       \\ 
    \midrule  \multicolumn{7}{c}{\textbf{Sem. Segment.} (mIoU $\uparrow$)} \\ \midrule
    $None$            & 17.10 $\pm$ 6.68       & 18.48 $\pm$ 6.44        & 19.36 $\pm$ 5.78        & 17.81 $\pm$ 7.25       & 18.55 $\pm$ 6.38       & 17.12 $\pm$ 6.69       \\
    $SC$              & 20.32 $\pm$ 4.08       & 20.00 $\pm$ 4.64        & 19.81 $\pm$ 5.48        & 20.00 $\pm$ 3.31       & 19.67 $\pm$ 5.24       & 20.01 $\pm$ 4.91       \\
    $C_{1\times1}$    & 19.45 $\pm$ 5.52       & 19.36 $\pm$ 5.73        & 19.88 $\pm$ 5.53        & 18.44 $\pm$ 6.16       & 19.90 $\pm$ 5.44       & 20.24 $\pm$ 5.14       \\
    $C_{3\times3}$    & \textbf{22.92} $\pm$ 3.57       & \textbf{21.95} $\pm$ 4.42        & \textbf{20.75} $\pm$ 5.18        & \textbf{23.54} $\pm$ 0.88       & \textbf{21.67} $\pm$ 4.38       & \textbf{22.42} $\pm$ 3.52       \\ 
    \midrule  \multicolumn{7}{c}{\textbf{Room Layout} (L2 Loss $\downarrow$)} \\ \midrule
    %$None$            & 0.788 $\pm$ 0.147      & 0.763 $\pm$ 0.137       & 0.737 $\pm$ 0.123       & 0.789 $\pm$ 0.156       & 0.759 $\pm$ 0.138       & 0.786 $\pm$ 0.148       \\
    %$SC$              & 0.705 $\pm$ 0.095      & 0.716 $\pm$ 0.105       & 0.726 $\pm$ 0.116       & 0.715 $\pm$ 0.095       & 0.722 $\pm$ 0.113       & 0.712 $\pm$ 0.106       \\
    %$C_{1\times1}$    & 0.730 $\pm$ 0.106      & 0.734 $\pm$ 0.112       & 0.729 $\pm$ 0.114       & 0.745 $\pm$ 0.104       & 0.726 $\pm$ 0.107       & 0.719 $\pm$ 0.098       \\
    %$C_{3\times3}$    & 0.684 $\pm$ 0.077      & 0.693 $\pm$ 0.095       & 0.714 $\pm$ 0.110       & 0.657 $\pm$ 0.024       & 0.699 $\pm$ 0.093       & 0.689 $\pm$ 0.076       \\ 
    $None$            & 0.79 $\pm$ 0.15      & 0.76 $\pm$ 0.14       & 0.74 $\pm$ 0.12       & 0.79 $\pm$ 0.16       & 0.76 $\pm$ 0.14       & 0.79 $\pm$ 0.15       \\
    $SC$              & 0.71 $\pm$ 0.10      & 0.72 $\pm$ 0.11       & 0.73 $\pm$ 0.12       & 0.72 $\pm$ 0.10       & 0.72 $\pm$ 0.11       & 0.71 $\pm$ 0.11       \\
    $C_{1\times1}$    & 0.73 $\pm$ 0.11      & 0.73 $\pm$ 0.11       & 0.73 $\pm$ 0.11       & 0.75 $\pm$ 0.10       & 0.73 $\pm$ 0.11       & 0.72 $\pm$ 0.10       \\
    $C_{3\times3}$    & \textbf{0.68} $\pm$ 0.08      & \textbf{0.69} $\pm$ 0.10       & \textbf{0.71} $\pm$ 0.11       & \textbf{0.66} $\pm$ 0.02       & \textbf{\textbf{0.70}} $\pm$ 0.09       & \textbf{0.69} $\pm$ 0.08       \\ 
    \midrule  \multicolumn{7}{c}{\textbf{Jigsaw} (Acc. (\%) $\uparrow$)} \\ \midrule
    $None$            & 62.08 $\pm$ 35.94       & 68.13 $\pm$ 33.58       & 74.14 $\pm$ 30.10       & 62.40 $\pm$ 36.16     & 69.08 $\pm$ 33.38       & 62.27 $\pm$ 35.98       \\
    $SC$              & 83.16 $\pm$ 18.51       & 80.60 $\pm$ 22.42       & 76.46 $\pm$ 28.29       & 84.10 $\pm$ 11.54     & 77.31 $\pm$ 27.62       & 79.41 $\pm$ 25.60       \\
    $C_{1\times1}$    & 73.84 $\pm$ 29.95       & 73.30 $\pm$ 30.57       & 76.42 $\pm$ 28.18       & 67.65 $\pm$ 34.49     & 77.18 $\pm$ 27.43       & 79.12 $\pm$ 25.49       \\
    $C_{3\times3}$    & \textbf{87.21} $\pm$ 17.53       & \textbf{84.26} $\pm$ 22.14       & \textbf{79.27} $\pm$ 26.43       & \textbf{92.14} $\pm$ 1.77      & \textbf{82.71} $\pm$ 22.09       & \textbf{85.50} $\pm$ 17.72       \\ 
    \bottomrule
    \end{tabular}
    }
    \vspace{-1em}
\end{table}

% op_pos  %
%%%%%%%%%%%

%%%%%%%%%%%
% opcomb  %

\begin{table}[tb]
  \caption{Mean validation accuracy (\%) and standard deviation of all architectures, evaluated based on the presence of the combination of 2 operations in NAS-Bench-101.}
  \label{tab:nb101opcombs}
  \centering
  \resizebox{\columnwidth}{!}{%
    \begin{tabular}{@{}llllll@{}}
    \toprule
    \textbf{Operation}        & $Input$             & $C_{1\times1}$        & $C_{3\times3}$          & $MP_{3\times3}$     & $Output$ \\ \midrule
       %\multicolumn{6}{c}{\textbf{CIFAR-10}} \\ \midrule
        $Input$          & --                       & 89.65 $\pm$ 7.60      & 90.92 $\pm$ 7.90        & 90.19 $\pm$ 4.83   & 83.16 $\pm$ 0.00  \\
        $C_{1\times1}$   & --                       & 89.24 $\pm$ 6.59      & 91.46 $\pm$ 3.77        & 89.29 $\pm$ 7.58   & 90.64 $\pm$ 4.40  \\
        $C_{3\times3}$   & --                       & 91.48 $\pm$ 3.98      & 91.66 $\pm$ 4.10        & 90.32 $\pm$ 8.36   & \textbf{91.83} $\pm$ 3.00  \\
        $MP_{3\times3}$  & --                       & 90.00 $\pm$ 5.19      & 91.06 $\pm$ 5.47        & 87.67 $\pm$ 12.22  & 88.36 $\pm$ 10.46 \\
        $Output$         & --                       & --                    & --                      & --                 & -- \\ \midrule
    \end{tabular}
    }
    %\vspace{-0.5em}
\end{table}

\begin{table}[tb]
  \caption{Mean validation accuracy (\%) and standard deviation of all architectures, evaluated based on the presence of the combination of 2 operations in NAS-Bench-201.}
  \label{tab:nb201opcombs}
  \centering
  \resizebox{\columnwidth}{!}{%
    \begin{tabular}{@{}llllll@{}}
    \toprule
    \textbf{Operation}        & $None$             & $C_{1\times1}$        & $C_{3\times3}$           & $AP_{3\times3}$     & $SC$ \\ \midrule
       \multicolumn{6}{c}{\textbf{CIFAR-10}} \\ \midrule
        $None$           & 70.78 $\pm$ 28.43 & 83.63 $\pm$ 14.53 & 84.65 $\pm$ 14.65 & 79.38 $\pm$ 15.28 & 82.10 $\pm$ 15.05 \\
        $C_{1\times1}$   & 83.73 $\pm$ 14.52 & 87.00 $\pm$ 6.60  & {87.80} $\pm$ {6.59}  & 84.77 $\pm$ 7.58  & 86.12 $\pm$ 7.05  \\
        $C_{3\times3}$   & 84.80 $\pm$ 14.66 & {87.84} $\pm$ {6.59}  & \textbf{88.10} $\pm$ {6.56}  & 85.55 $\pm$ 7.65  & 86.87 $\pm$ 7.11  \\
        $AP_{3\times3}$  & 79.10 $\pm$ 15.22 & 84.57 $\pm$ 7.69  & 85.36 $\pm$ 7.67  & 79.31 $\pm$ 12.09 & 81.25 $\pm$ 11.96 \\
        $SC$             & 82.11 $\pm$ 15.01 & 86.10 $\pm$ 7.14  & 86.78 $\pm$ 7.19  & 81.38 $\pm$ 11.93 & 82.76 $\pm$ 11.50 \\ \midrule
        \multicolumn{6}{c}{\textbf{CIFAR-100}} \\ \midrule
        $None$            & 49.70 $\pm$ 23.23 & 61.65 $\pm$ 12.68 & 63.24 $\pm$ 12.75 & 56.44 $\pm$ 13.81 & 59.26 $\pm$ 13.51 \\
        $C_{1\times1}$    & 61.81 $\pm$ 12.63 & 65.42 $\pm$ 6.45  & {66.62} $\pm$ {6.33}  & 62.04 $\pm$ 8.37  & 63.55 $\pm$ 7.75  \\
        $C_{3\times3}$    & 63.42 $\pm$ 12.77 & {66.65} $\pm$ {6.35}  & \textbf{67.12} $\pm$ {6.25}  & 63.25 $\pm$ 8.29  & 64.73 $\pm$ 7.66  \\
        $AP_{3\times3}$   & 56.17 $\pm$ 13.75 & 61.83 $\pm$ 8.49  & 63.03 $\pm$ 8.35  & 55.20 $\pm$ 12.89 & 57.30 $\pm$ 12.87 \\
        $SC$              & 59.15 $\pm$ 13.42 & 63.51 $\pm$ 7.84  & 64.66 $\pm$ 7.72  & 57.41 $\pm$ 12.82 & 58.90 $\pm$ 12.50 \\ \midrule
        \multicolumn{6}{c}{\textbf{ImageNet16-120}} \\ \midrule
        $None$           & 24.91 $\pm$ 13.43 & 33.68 $\pm$ 9.34 & 34.94 $\pm$ 9.60  & 28.17 $\pm$ 10.10 & 31.91 $\pm$ 9.86 \\
        $C_{1\times1}$   & 33.89 $\pm$ 9.23  & 37.67 $\pm$ 6.10 & \textbf{39.11} $\pm$ {5.85}  & 34.09 $\pm$ 7.46  & 35.90 $\pm$ 6.78  \\
        $C_{3\times3}$   & 35.38 $\pm$ 9.44  & {39.02} $\pm$ {5.93} & {38.91} $\pm$ {5.78}  & 35.25 $\pm$ 7.36  & 37.44 $\pm$ 6.66  \\
        $AP_{3\times3}$  & 27.74 $\pm$ 10.10 & 33.82 $\pm$ 7.69 & 35.04 $\pm$ 7.51  & 28.10 $\pm$ 8.96  & 30.89 $\pm$ 9.02  \\
        $SC$             & 31.69 $\pm$ 9.70  & 35.98 $\pm$ 6.95 & 37.46 $\pm$ 6.71  & 31.04 $\pm$ 8.88  & 32.80 $\pm$ 8.44 \\ \bottomrule
    \end{tabular}
    }
    %\vspace{-0.5em}
    \vspace{-1em}
\end{table}

\begin{table}[tb]
  \caption{Mean performance metric and standard deviation of all architectures, evaluated based on the presence of the combination of 2 operations in TransNAS-Bench-101.}
  \label{tab:tnas101opcombs}
  \centering
  \resizebox{1\columnwidth}{!}{%
    \begin{tabular}{@{}lcccc@{}}
    \toprule
    \textbf{Operation}        & $None$        & $SC$     & $C_{1\times1}$        & $C_{3\times3}$           \\ \midrule
       \multicolumn{5}{c}{\textbf{Cls. Object} (Acc. (\%) $\uparrow$)} \\ \midrule
       %\multicolumn{6}{c}{\textbf{CIFAR-10}} \\ \midrule
        $None$      & 34.43 $\pm$ 7.07      & 38.11 $\pm$ 6.23        & 37.42 $\pm$ 6.69   & 40.94 $\pm$ 5.11  \\
        $SC$        & 38.05 $\pm$ 6.12      & 39.69 $\pm$ 5.29        & 39.60 $\pm$ 5.51   & 42.20 $\pm$ 4.03 \\
        $C_{1\times1}$ & 37.29 $\pm$ 6.71   & 39.34 $\pm$ 5.80        & 38.72 $\pm$ 6.21   & 41.71 $\pm$ 4.51  \\
        $C_{3\times3}$ & 41.18 $\pm$ 5.09   & 42.20 $\pm$ 4.20        & 41.79 $\pm$ 4.49   & \textbf{42.74} $\pm$ 3.05  \\
        \midrule
       \multicolumn{5}{c}{\textbf{Cls. Scene} (Acc. (\%) $\uparrow$)} \\ \midrule
       $None$      &  34.30 $\pm$ 15.69       &  41.74 $\pm$ 12.63         &  40.86 $\pm$ 13.75    &  48.15 $\pm$ 11.15   \\
        $SC$       &  41.79 $\pm$ 12.44       &  44.58 $\pm$ 10.25         &  44.60 $\pm$ 10.98    &  50.03 $\pm$ 7.90  \\
        $C_{1\times1}$ &  40.56 $\pm$ 13.72       &  44.12 $\pm$ 11.43         &  43.40 $\pm$ 12.44    &  49.59 $\pm$ 9.11  \\
        $C_{3\times3}$ &  48.52 $\pm$ 10.89       &  49.93 $\pm$ 8.19         &  49.70 $\pm$ 9.15    &  \textbf{51.79} $\pm$ 6.29  \\
       \midrule \multicolumn{5}{c}{\textbf{Autoencoding} (SSIM $\uparrow$)} \\ \midrule
        $None$      &  0.36 $\pm$ 0.17       &  0.47 $\pm$ 0.11         &  0.43 $\pm$ 0.11    &  0.45 $\pm$ 0.10   \\
        $SC$       &  0.47 $\pm$ 0.12       &  \textbf{0.51} $\pm$ 0.06         &  0.49 $\pm$ 0.07    &  0.50 $\pm$ 0.06  \\
        $C_{1\times1}$ &  0.42 $\pm$ 0.11       &  0.48 $\pm$ 0.08         &  0.45 $\pm$ 0.08    &  0.46 $\pm$ 0.07  \\
        $C_{3\times3}$ &  0.45 $\pm$ 0.10       &  0.50 $\pm$ 0.07         &  0.46 $\pm$ 0.07    &  0.47 $\pm$ 0.06  \\
       \midrule \multicolumn{5}{c}{\textbf{Surf. Normal} (SSIM $\uparrow$)} \\ \midrule
        $None$      &  0.46 $\pm$ 0.09       &  0.51 $\pm$ 0.06         &  0.50 $\pm$ 0.07    &  0.53 $\pm$ 0.06   \\
        $SC$       &  0.51 $\pm$ 0.06       &  0.53 $\pm$ 0.04         &  0.52 $\pm$ 0.05    &  0.54 $\pm$ 0.03  \\
        $C_{1\times1}$ &  0.49 $\pm$ 0.07       &  0.53 $\pm$ 0.05         &  0.51 $\pm$ 0.06    &  0.54 $\pm$ 0.05  \\
        $C_{3\times3}$ &  0.53 $\pm$ 0.06       &  0.54 $\pm$ 0.04         &  0.54 $\pm$ 0.04    &  \textbf{0.55} $\pm$ 0.03  \\
       \midrule \multicolumn{5}{c}{\textbf{Sem. Segment.} (mIoU $\uparrow$)} \\ \midrule
        $None$      &  15.18 $\pm$ 7.36       &  18.55 $\pm$ 5.55         &  17.91 $\pm$ 6.48    &  21.41 $\pm$ 5.09   \\
        $SC$       &  18.63 $\pm$ 5.39       &  19.71 $\pm$ 4.03         &  19.58 $\pm$ 4.79    &  21.93 $\pm$ 3.38  \\
        $C_{1\times1}$ &  17.66 $\pm$ 6.50       &  19.36 $\pm$ 5.06         &  18.81 $\pm$ 5.87    &  21.80 $\pm$ 4.33  \\
        $C_{3\times3}$ &  21.60 $\pm$ 5.09       &  21.99 $\pm$ 3.49         &  21.95 $\pm$ 4.36    &  \textbf{23.12} $\pm$ 2.95 \\
       \midrule \multicolumn{5}{c}{\textbf{Room Layout} (L2 Loss $\downarrow$)} \\ \midrule
        $None$      &  0.84 $\pm$ 0.17       &  0.75 $\pm$ 0.13         &  0.77 $\pm$ 0.13    &  0.71 $\pm$ 0.11   \\
        $SC$       &  0.75 $\pm$ 0.13       &  0.72 $\pm$ 0.10         &  0.72 $\pm$ 0.10    &  0.68 $\pm$ 0.08  \\
        $C_{1\times1}$ &  0.77 $\pm$ 0.13       &  0.73 $\pm$ 0.10         &  0.74 $\pm$ 0.11    &  0.69 $\pm$ 0.08  \\
        $C_{3\times3}$ &  0.71 $\pm$ 0.11       &  0.68 $\pm$ 0.08         &  0.69 $\pm$ 0.08    &  \textbf{0.68} $\pm$ 0.06  \\
       \midrule \multicolumn{5}{c}{\textbf{Jigsaw} (Acc. (\%) $\uparrow$)} \\ \midrule
        $None$      &  50.57 $\pm$ 38.75       &  72.82 $\pm$ 29.02         &  65.32 $\pm$ 34.60    &  79.96 $\pm$ 25.76   \\
        $SC$       &  73.50 $\pm$ 27.72       &  81.67 $\pm$ 19.14         &  78.76 $\pm$ 24.37    &  87.38 $\pm$ 15.53  \\
        $C_{1\times1}$ &  63.77 $\pm$ 35.21       &  76.62 $\pm$ 26.85         &  70.90 $\pm$ 31.92    &  83.43 $\pm$ 22.59  \\
        $C_{3\times3}$ &  80.98 $\pm$ 25.54       &  87.20 $\pm$ 16.32         &  83.96 $\pm$ 22.50    &  \textbf{88.34} $\pm$ 14.39  \\
        \bottomrule
    \end{tabular}
    }
    \vspace{-1em}
\end{table}

% opcomb  %
%%%%%%%%%%%

\subsubsection{Cell's Ranking and Inter-data set Correlation}
\begin{figure*}[tb]
\begin{subfigure}[b]{0.49\textwidth}
  \centering
  \includegraphics[width=0.85\textwidth]{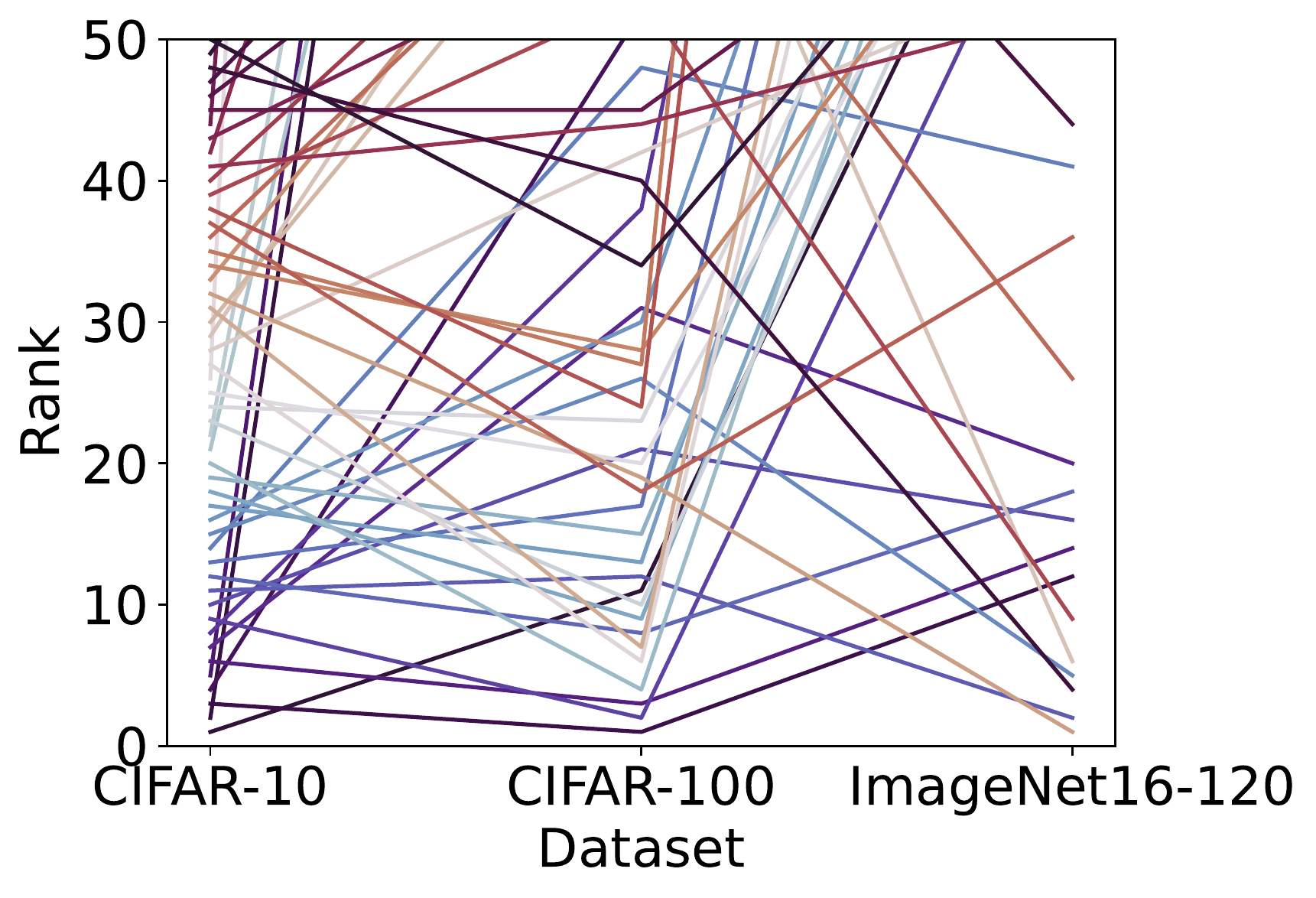}
  \caption{Top-50 Architectures inter-rank.}
  \label{fig:sfig2}
\end{subfigure}
\begin{subfigure}[b]{.49\textwidth}
  \centering
  \includegraphics[width=0.85\textwidth]{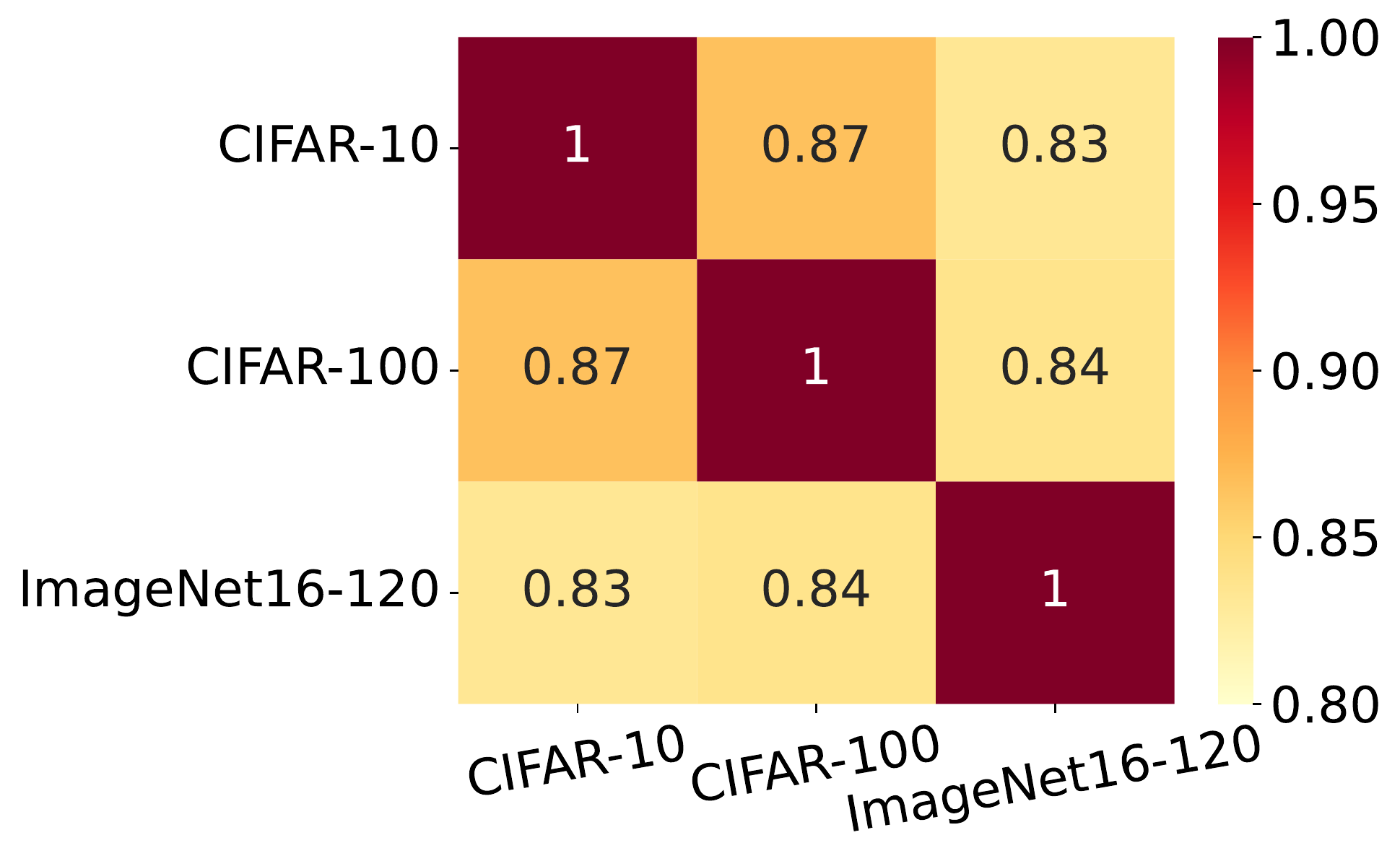}
  \caption{Data set Kendall's tau correlation.}
  \label{fig:sfig4}
\end{subfigure}
  \caption{Evaluation of NAS-Bench-201 architectures in terms of their inter-data set ranking and the correlation between data sets. On (a) we show the ranking of the top-50 architectures from CIFAR-10 when transferred and evaluated to CIFAR-100 and ImageNet16-120; and (b) the Kendall's Tau correlation between data sets.\label{fig:nb201rank}} %Ranking of the top-50 architectures from CIFAR-10 when transferred and evaluated to other NAS-Bench-201 data sets. CIFAR-10 and CIFAR-100 top architectures show similarities in terms of ranking, whereas ImageNet16-120 shows more heterogeneity.
\end{figure*}

\begin{figure*}[tb]
\begin{subfigure}[b]{0.49\textwidth}
  \centering
  \includegraphics[width=0.85\textwidth]{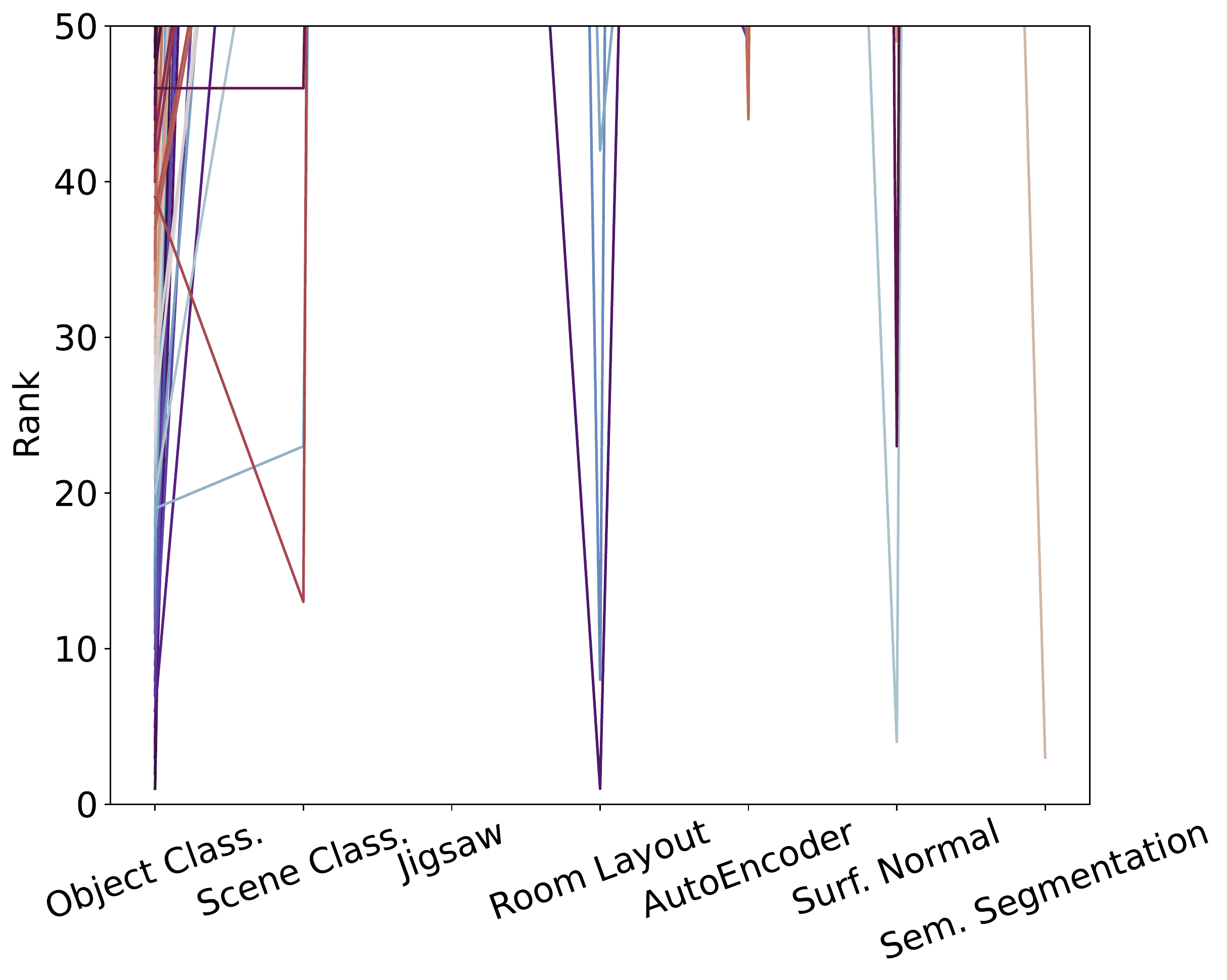}
  \caption{Top-50 Architectures inter-rank.}
  \label{fig:sfig2}
\end{subfigure}
\begin{subfigure}[b]{.49\textwidth}
  \centering
  \includegraphics[width=0.85\textwidth]{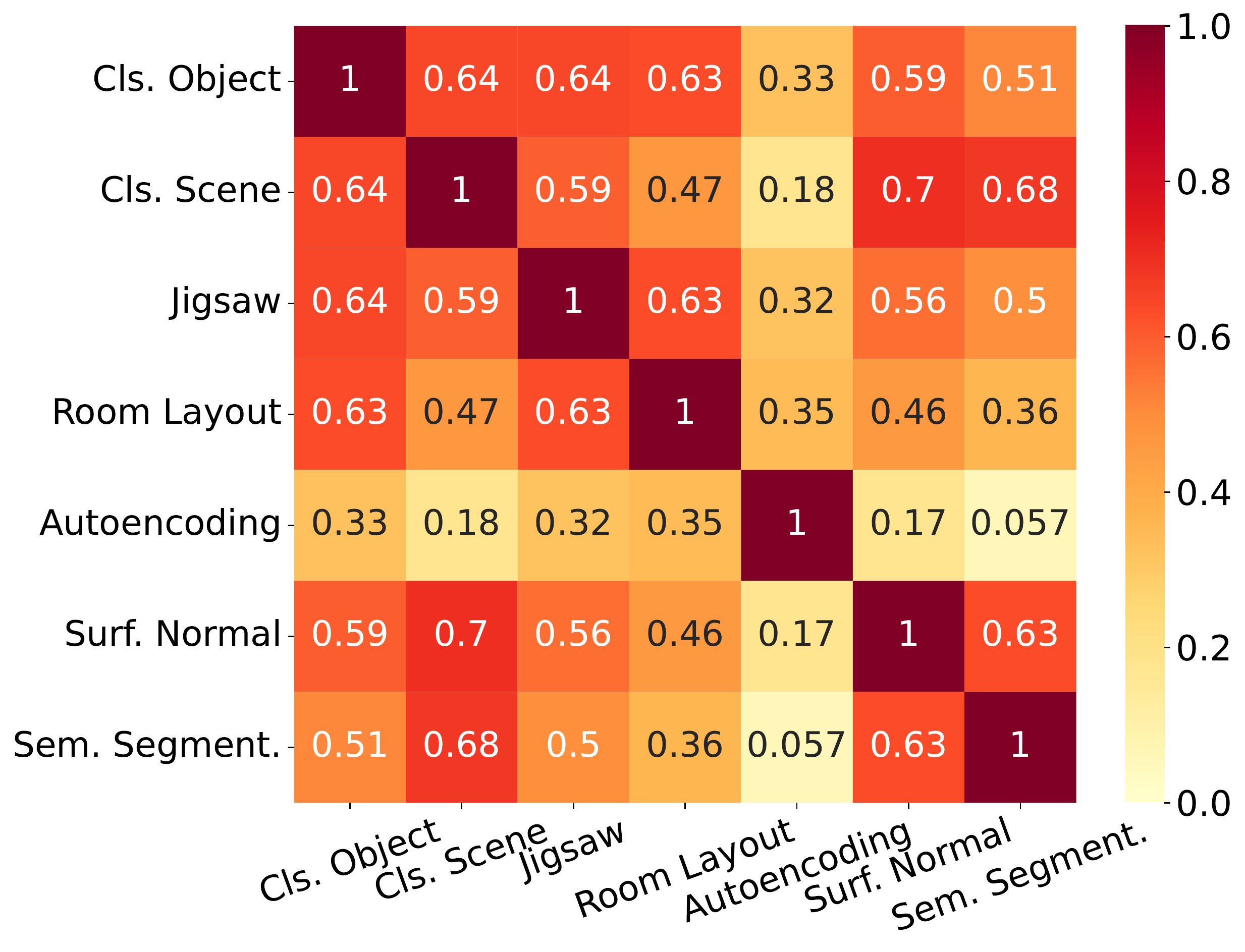}
  \caption{Task Kendall's tau correlation.}
  \label{fig:sfig4}
\end{subfigure}
\caption{Evaluation of TransNAS-Bench-101 architectures in terms of their inter-task ranking and task correlation. On (a) we show the ranking of the top-50 architectures from Object Classification when transferred and evaluated to other tasks; and (b) the Kendall's Tau correlation between tasks.}%
\label{fig:transnas101-corr}
\vspace{-1em}
\end{figure*}

Lastly, we evaluate cells based on their ranking by sorting them using the performance on the validation set. The intuition behind looking at the ranking of the top cells is that it allows a direct analysis of the top cells based on their operation diversity, and on benchmarks with more than 1 data set or task, allows an evaluation of how the different data sets correlate between each other. First, we look at the top 10 performant architectures in NAS-Bench-101. The cells are shown in Table \ref{tab:nb101_top10_archs}, where on the left the operations in each position are shown, and on the right the associated rank and validation accuracy is depicted. From the table, we clearly see that convolutional layers are the preferred operation on all positions, with the optimal cell entirely designed with $C_{3\times3}$ operations. Overall, convolutional layers represent 84\% of the operations present in the top-10 cells in NAS-Bench-101 (not considering the fixed input and output operations), thus suggesting that operations other than convolutional ones have a small impact on the final performance of the cell. More, this indicates that a NAS method that searches solely for convolutional layers will in fact achieve high results without requiring exploring the search space for other operations. By looking at NAS-Bench-201 in Table \ref{tab:nb201_top10_archs}, we see the same behaviour, where convolutional layers represent 77\% of the top cells considering all three data sets, having a representation of 75\% in CIFAR-10, 80\% in CIFAR-100 and 77\% in ImageNet16-120. This further indicates that NAS-Bench-201 is heavily dependant on convolutional layers in top-scoring architectures. More, in all three data sets, it is possible to see patterns of combination between the possible two convolutional layers ($C_{1\times1}$ and $C_{3\times3}$), as well as skip connections on the 4th position of the cell structure. These skip connections allow residual connections from the input node to the output node, which, combined with different position permutations of the convolutional layers, yield the best results. We note that most of the best architectures for all three data sets contain convolutional layers as the first and fifth operations and an $SC$ as the fourth. Hence, one could envision a NAS procedure where these three operations would be fixed and only the remaining ones would be the target of the search, thus achieving excellent results across the three data sets, considering only a subspace related to half of the original number of operations. Furthermore, the cell's ranking between data sets suggests that top-scoring architectures on one data set will have good results on other data sets if transferred. This is more explicit between CIFAR-10 and CIFAR-100, as these two data sets have similarities at the level of the problem itself. To further study this, we evaluated the inter-data set ranking and cross-correlation between data sets on Figure \ref{fig:nb201rank}. On the left (a), we present the top-50 architectures on CIFAR-10 and their rankings on CIFAR-100 and ImageNet16-120, and on the right (b), the correlation between data sets in terms of the performance of all architectures. It is clear that there is a high positive correlation between all data sets, in particular, between the CIFAR data sets. These results suggest that NAS-Bench-201 should not be used alone when evaluating the generability of a NAS method in different data sets. More, when using NAS-Bench-201, it is paramount to show results on directly searching in each data set, as ImageNet16-120 has the lowest correlation with other data sets, it allows a better understanding of how good a NAS method is. As for the position of the operations in the top cells of TransNAS-Bench-101, we see in Table \ref{tab:tnb101_top10_archs} that convolutional layers have a smaller representation across all tasks, with 58\% of the possible operations being convolutional layers, 31\% $SC$ and 11\% $None$. Looking at specific tasks, semantic segmentation has the highest percentage of convolutional layers, with 80\% and 66\% respectively, while autoencoding, room layout and object classification have the lowest, with 42\%, 50\% and 51\% respectively. More, in those that the percentage of convolutional layers is lower, $SC$ represents a larger part, with approximately 42\% in all three. As TransNAS-Bench-101 uses the same cell structure as NAS-Bench-201, we see the same pattern in terms of operation position, where $SC$ is frequently present in the 4th position to create residual connections between the input and output nodes, and convolutional layers are the preferred operation in the first positions. However, the evaluated tasks allow for a more diverse use of operations, requiring NAS methods to be more general and capable of learning which patterns are important for each task, which is further justified by the inter-task ranking, where we see that a good architecture in a specific task will not guarantee a good result when transferred to other tasks. This is further studied in Fig. \ref{fig:transnas101-corr}, where on the left, the top 50 architectures on object classification do appear as top architectures in other tasks. More, on the right figure, the Kendall Tau correlation suggests that the tasks are not as correlated as NAS-Bench-201, and Autoencoding is the task with the lowest correlation with any other. By having a lower correlation between tasks, TransNAS-Bench-101 becomes an interesting benchmark to evaluate the generability of NAS methods, as transferring architectures searched on a single task to others does directly translate to good results.

\begin{table}[tb]
  \caption{Top 10 cells on NAS-Bench-101 based on the validation accuracy (\%).}
  \label{tab:nb101_top10_archs}
    \centering
  \resizebox{\columnwidth}{!}{%
    \begin{tabular}{@{}lllllllll@{}} \toprule
    \multicolumn{7}{c}{\textbf{Operations}} & \multicolumn{2}{c}{\textbf{CIFAR-10}} \\ \cmidrule(lr){1-7} \cmidrule(lr){8-9} 
    \multicolumn{1}{c}{\textbf{1}} & \multicolumn{1}{c}{\textbf{2}} & \multicolumn{1}{c}{\textbf{3}} & \multicolumn{1}{c}{\textbf{4}} & \multicolumn{1}{c}{\textbf{5}} & \multicolumn{1}{c}{\textbf{6}} & \multicolumn{1}{c}{\textbf{7}} & \multicolumn{1}{c}{\textbf{Rank}} & \multicolumn{1}{c}{\textbf{Val. Acc (\%)}}\\ \midrule
    $Input$ & $C_{3\times3}$ & $C_{3\times3}$ & $C_{3\times3}$ & $C_{3\times3}$ & $Output$ & - & \textbf{1} & 95.18\% \\
    $Input$ & $C_{1\times1}$ & $MP_{3\times3}$& $C_{3\times3}$ & $C_{3\times3}$ & $C_{1\times1}$ & $Output$ & \textbf{2} & 95.11\% \\
    $Input$ & $C_{1\times1}$ & $C_{3\times3}$ & $C_{3\times3}$ & $C_{1\times1}$ & $Output$ & - & \textbf{3} & 95.11\% \\
    $Input$ & $C_{3\times3}$ & $C_{3\times3}$ & $C_{1\times1}$ & $C_{3\times3}$ & $Output$ & - & \textbf{4} & 95.07\% \\
    $Input$ & $MP_{3\times3}$ & $C_{3\times3}$ & $C_{1\times1}$ & $C_{3\times3}$ & $C_{1\times1}$ & $Output$ & \textbf{5} & 95.06\% \\
    $Input$ & $C_{3\times3}$ & $MP_{3\times3}$ & $C_{3\times3}$ & $C_{1\times1}$ & $C_{3\times3}$ & $Output$ & \textbf{6} & 95.04\% \\
    $Input$ & $C_{3\times3}$ & $MP_{3\times3}$ & $C_{3\times3}$ & $C_{1\times1}$ & $Output$ & - & \textbf{7} & 95.03\% \\
    $Input$ & $C_{3\times3}$ & $MP_{3\times3}$ & $C_{3\times3}$ & $C_{3\times3}$ & $Output$ & - & \textbf{8} & 94.99\% \\
    $Input$ & $MP_{3\times3}$ & $C_{3\times3}$ & $C_{1\times1}$ & $C_{3\times3}$ & $Output$ & - & \textbf{9} & 94.95\% \\
    $Input$ & $MP_{3\times3}$ & $C_{3\times3}$ & $C_{1\times1}$ & $C_{3\times3}$ & $Output$ & - & \textbf{10} & 94.94 \%\\ \bottomrule
    \end{tabular}
    }
    \vspace{-0.75em}
\end{table}

\begin{table}[tb]
  \caption{Top 10 cells for each one of the data sets on NAS-Bench-201. Validation accuracy (\%) is given only for the top-10 cells in each data set, while ranking is also presented for the architectures outside the 10 best, allowing inter-data set comparison.}
  \label{tab:nb201_top10_archs}
    \centering
  \resizebox{\columnwidth}{!}{%
    \begin{tabular}{@{}llllllllllll@{}} \toprule
    \multicolumn{6}{c}{\textbf{Operations}} & \multicolumn{6}{c}{\textbf{Rank / Val. Acc.} ($\%)$} \\ \cmidrule(lr){1-6} \cmidrule(lr){7-12} 
    \multicolumn{1}{c}{\textbf{1}} & \multicolumn{1}{c}{\textbf{2}} & \multicolumn{1}{c}{\textbf{3}} & \multicolumn{1}{c}{\textbf{4}} & \multicolumn{1}{c}{\textbf{5}} & \multicolumn{1}{c}{\textbf{6}} & \multicolumn{2}{c}{\textbf{C10}} & \multicolumn{2}{c}{\textbf{C100}} & \multicolumn{2}{c}{\textbf{IN16-120}} \\ \midrule
    $C_{3\times3}$ & $C_{3\times3}$ & $C_{3\times3}$ & $SC$ & $C_{3\times3}$ & $C_{1\times1}$  & \textbf{1} & 91.61\% &   11 &  &   74 &  \\
    $AP_{3\times3}$ & $C_{3\times3}$ & $None$ & $SC$ & $C_{1\times1}$ & $C_{3\times3}$         & \textbf{2} & 91.57\% &   202  &  &   1083 &  \\
    $C_{3\times3}$ & $C_{3\times3}$ & $C_{3\times3}$ & $SC$ & $C_{3\times3}$ & $C_{3\times3}$  & \textbf{3} & 91.55\% & \textbf{1} & 73.49\% &   12 &  \\
    $C_{3\times3}$ & $None$ & $C_{3\times3}$ & $SC$ & $C_{1\times1}$ & $C_{1\times1}$          & \textbf{4} & 91.55\% &   52 &  &   87 &  \\
    $C_{3\times3}$ & $C_{1\times1}$ & $SC$ & $SC$ & $C_{3\times3}$ & $C_{1\times1}$            & \textbf{5} & 91.54\% &   219 &  &  112 &  \\
    $C_{3\times3}$ & $C_{1\times1}$ & $C_{3\times3}$ & $SC$ & $C_{3\times3}$ & $C_{3\times3}$  & \textbf{6} & 91.53\% & \textbf{3} & 73.13\% &  14 &  \\
    $C_{3\times3}$ & $C_{3\times3}$ & $C_{1\times1}$ & $SC$ & $C_{3\times3}$ & $C_{1\times1}$  & \textbf{7} & 91.52\% &  31 &  &  20 &  \\
    $C_{3\times3}$ & $C_{3\times3}$ & $None$ & $SC$ & $C_{3\times3}$ & $C_{1\times1}$          & \textbf{8} & 91.51\% &  38 &  &  188 &  \\
    $C_{3\times3}$ & $C_{3\times3}$ & $C_{3\times3}$ & $SC$ & $C_{1\times1}$ & $C_{3\times3}$  & \textbf{9} & 91.50\% & \textbf{2} & 73.31\% &  66 &  \\
    $C_{3\times3}$ & $C_{1\times1}$ & $C_{3\times3}$ & $SC$ & $C_{1\times1}$ & $C_{3\times3}$  & \textbf{10} & 91.48\% &  21 &  &  16 &  \\
    $C_{3\times3}$ & $C_{3\times3}$ & $None$ & $SC$ & $C_{1\times1}$ & $C_{3\times3}$          &  20 &  & \textbf{4} & 73.09\% &  94 &  \\
    $C_{3\times3}$ & $C_{3\times3}$ & $SC$ & $SC$ & $C_{3\times3}$ & $C_{1\times1}$            &  128 &  & \textbf{5} & 73.02\% &  253 &  \\
    $C_{1\times1}$ & $C_{3\times3}$ & $C_{1\times1}$ & $SC$ & $C_{3\times3}$ & $C_{3\times3}$  &  27 &  & \textbf{6} & 72.98\% &  134 &  \\
    $C_{3\times3}$ & $C_{3\times3}$ & $C_{1\times1}$ & $SC$ & $C_{1\times1}$ & $C_{3\times3}$  &  31 &  & \textbf{7} & 72.96\% &  142 &  \\
    $C_{3\times3}$ & $C_{3\times3}$ & $C_{3\times3}$ & $SC$ & $C_{1\times1}$ & $C_{1\times1}$  &  12 &  & \textbf{8} & 72.95\% &  18 &  \\
    $C_{1\times1}$ & $C_{3\times3}$ & $C_{3\times3}$ & $SC$ & $C_{1\times1}$ & $C_{3\times3}$  &  18 &  & \textbf{9} & 72.86\% &  85 &  \\
    $C_{1\times1}$ & $C_{3\times3}$ & $C_{3\times3}$ & $SC$ & $C_{3\times3}$ & $C_{3\times3}$  &  23 &  & \textbf{10} & 72.77\% &  77 &  \\
    $C_{3\times3}$ & $C_{1\times1}$ & $C_{1\times1}$ & $SC$ & $C_{3\times3}$ & $C_{3\times3}$  &  32 &  &  19 &  & \textbf{1} & 46.73\% \\
    $C_{3\times3}$ & $C_{1\times1}$ & $C_{3\times3}$ & $SC$ & $C_{3\times3}$ & $C_{1\times1}$  &  11 &  &  12 &  & \textbf{2} & 46.56\% \\
    $C_{3\times3}$ & $C_{1\times1}$ & $C_{3\times3}$ & $C_{1\times1}$ & $None$ & $C_{3\times3}$&  414 &  &  270 &  & \textbf{3} & 46.52\% \\
    $C_{3\times3}$ & $None$ & $C_{3\times3}$ & $SC$ & $C_{3\times3}$ & $C_{1\times1}$          &  48 & &  40 &  & \textbf{4} & 46.50\% \\
    $C_{1\times1}$ & $C_{3\times3}$ & $C_{3\times3}$ & $SC$ & $C_{3\times3}$ & $C_{1\times1}$  &  15 &  &  26 &  & \textbf{5} & 46.50\% \\
    $C_{3\times3}$ & $C_{1\times1}$ & $C_{3\times3}$ & $SC$ & $C_{1\times1}$ & $C_{1\times1}$  &  29 &  &  75 &  & \textbf{6} & 46.49\% \\
    $C_{3\times3}$ & $C_{1\times1}$ & $C_{3\times3}$ & $C_{1\times1}$ & $SC$ & $C_{1\times1}$  &  753 &  &  728  &  & \textbf{7} & 46.47\% \\
    $C_{1\times1}$ & $C_{3\times3}$ & $C_{3\times3}$ & $SC$ & $C_{1\times1}$ & $C_{1\times1}$  &  280 &  &  55 &  & \textbf{8} & 46.45\% \\
    $C_{3\times3}$ & $SC$ & $C_{3\times3}$ & $SC$ & $C_{1\times1}$ & $C_{3\times3}$            &  39 &  &  53 &  & \textbf{9} & 46.40\% \\
    $C_{1\times1}$ & $C_{1\times1}$ & $C_{3\times3}$ & $SC$ & $C_{3\times3}$ & $C_{3\times3}$  &  83 &  &  22 &  & \textbf{10} & 46.38\% \\ \bottomrule
    \end{tabular}
    }
    \vspace{-1.5em}
\end{table}

\begin{table*}[!tb]
  \caption{Top 10 cells for each task on NAS-Bench-201. Performance is given only for the top-10 cells in each task while ranking is also presented for architectures outside the top 10, promoting inter-task comparison.}
  \label{tab:tnb101_top10_archs}
    \centering
  \resizebox{0.95\textwidth}{!}{%
\begin{tabular}{@{}llllllllllllllllllll@{}}
\hline
\multicolumn{6}{c}{Operations} & \multicolumn{14}{c}{Rank / Performance} \\ \cmidrule(lr){1-6} \cmidrule(lr){7-20}
1 & \multicolumn{1}{c}{2} & \multicolumn{1}{c}{3} & \multicolumn{1}{c}{4} & \multicolumn{1}{c}{5} & \multicolumn{1}{c}{6} & \multicolumn{2}{c}{\begin{tabular}[c]{@{}c@{}}Cls.\\ Object\end{tabular}} & \multicolumn{2}{c}{\begin{tabular}[c]{@{}c@{}}Cls.\\ Scene\end{tabular}} & \multicolumn{2}{c}{Autoencoding} & \multicolumn{2}{c}{\begin{tabular}[c]{@{}c@{}}Surf. \\ Normal\end{tabular}} & \multicolumn{2}{c}{\begin{tabular}[c]{@{}c@{}}Sem. \\ Segment.\end{tabular}} & \multicolumn{2}{c}{\begin{tabular}[c]{@{}c@{}}Room\\ Layout\end{tabular}} & \multicolumn{2}{c}{Jigsaw} \\ \bottomrule
$C_{3\times3}$  & $None$ & $C_{3\times3}$  & $SC$ & $C_{1\times1}$ & $C_{1\times1}$ & \textbf{1} & 46.32 & 534 &  &  791 &  &  172 &  &  582 &  &  2072 &  &  126 &  \\
$C_{3\times3}$  & $SC$ & $C_{3\times3}$  & $SC$ & $SC$ & $C_{3\times3}$  & \textbf{2} &  46.23 &  469 &  &  1124 &  &  495 &  &  1645 &  &  1057 &  &  1460 &  \\
$C_{3\times3}$  & $SC$ & $SC$ & $C_{1\times1}$ & $C_{3\times3}$  & $SC$ & \textbf{3} &  46.06 &  920 &  &  116 &  &  310 &  &  1191 &  &  736 &  &  111 &  \\
$C_{3\times3}$  & $SC$ & $C_{1\times1}$ & $C_{1\times1}$ & $None$ & $SC$ & \textbf{4} &  45.98 &  369 &  &  1367 &  &  1331 &  &  2153 &  &  316 &  &  477 &  \\
$C_{3\times3}$  & $SC$ & $SC$ & $SC$ & $C_{3\times3}$  & $SC$ & \textbf{5} &  45.98 &  292 &  &  390 &  &  701 &  &  2259 &  & \textbf{1} & 0.59 &  146 &  \\
$SC$ & $C_{1\times1}$ & $C_{3\times3}$  & $C_{3\times3}$  & $SC$ & $C_{3\times3}$  & \textbf{6} &  45.88 &  115 &  &  768 &  &  332 &  &  519 &  &  752 &  &  1567 &  \\
$C_{3\times3}$  & $SC$ & $C_{3\times3}$  & $SC$ & $None$ & $SC$ & \textbf{7} &  45.86 &  543 &  &  49 &  &  234 &  &  2504 &  &  74 &  &  150 &  \\
$C_{3\times3}$  & $C_{3\times3}$  & $C_{1\times1}$ & $SC$ & $C_{3\times3}$  & $SC$ & \textbf{8} &  45.86 &  331 &  &  85 &  &  437 &  &  553 &  &  748 &  &  534 &  \\
$SC$ & $C_{3\times3}$  & $None$ & $C_{1\times1}$ & $SC$ & $C_{3\times3}$  & \textbf{9} &  45.86 &  178 &  &  344 &  &  192 &  &  1442 &  &  1595 &  &  240 &  \\
$C_{3\times3}$  & $C_{3\times3}$  & $SC$ & $SC$ & $C_{3\times3}$  & $SC$ & \textbf{10} &  45.81 &  297 &  &  1206 &  &  902 &  &  1536 &  &  115 &  &  778 &  \\
$C_{3\times3}$  & $C_{3\times3}$  & $C_{1\times1}$ & $SC$ & $C_{3\times3}$  & $C_{1\times1}$ &  61 &  & \textbf{1} & 54.94 &  218 &  &  52 &  &  800 &  &  1647 &  &  423 &  \\
$C_{3\times3}$  & $SC$ & $C_{3\times3}$  & $SC$ & $C_{1\times1}$ & $C_{1\times1}$ &  148 &  & \textbf{2} & 54.93 &  958 &  &  1728 &  &  2269 &  &  13 &  &  974 &  \\
$C_{3\times3}$  & $C_{3\times3}$  & $C_{3\times3}$  & $SC$ & $None$ & $C_{1\times1}$ &  213 &  & \textbf{3} & 54.87 &  565 &  &  307 &  &  1871 &  &  1836 &  &  344 &  \\
$C_{3\times3}$  & $C_{3\times3}$  & $C_{3\times3}$  & $SC$ & $C_{3\times3}$  & $C_{3\times3}$  &  572 &  & \textbf{4} & 54.87 &  1411 &  &  266 &  &  1274 &  &  2191 &  &  32 &  \\
$C_{3\times3}$  & $None$ & $SC$ & $C_{1\times1}$ & $C_{3\times3}$  & $SC$ &  1129 &  & \textbf{5} & 54.86 &  3019 &  &  187 &  &  117 &  &  2214 &  &  790 &  \\
$C_{3\times3}$  & $C_{1\times1}$ & $C_{3\times3}$  & $None$ & $SC$ & $SC$ &  1132 &  & \textbf{6} & 54.86 &  3020 &  &  188 &  &  118 &  &  2215 &  &  791 &  \\
$C_{3\times3}$  & $C_{1\times1}$ & $C_{3\times3}$  & $C_{1\times1}$ & $SC$ & $SC$ &  1490 &  & \textbf{7} & 54.85 &  2724 &  &  547 &  &  796 &  &  1772 &  &  153 &  \\
$C_{3\times3}$  & $SC$ & $C_{1\times1}$ & $C_{1\times1}$ & $SC$ & $C_{1\times1}$ &  93 &  & \textbf{8} & 54.80 &  2296 &  &  1834 &  &  229 &  &  1298 &  &  197 &  \\
$C_{3\times3}$  & $C_{3\times3}$  & $None$ & $C_{1\times1}$ & $SC$ & $C_{3\times3}$  &  928 &  & \textbf{9} & 54.79 &  1906 &  &  339 &  &  23 &  &  1058 &  &  1323 &  \\
$C_{3\times3}$  & $SC$ & $C_{1\times1}$ & $C_{3\times3}$  & $C_{3\times3}$  & $SC$ &  381 &  & \textbf{10} & 54.77 &  127 &  &  361 &  &  1287 &  &  310 &  &  123 &  \\
$SC$ & $C_{1\times1}$ & $None$ & $C_{3\times3}$  & $SC$ & $None$ &  529 &  &  2545 &  & \textbf{1} & 0.58 &  1355 &  &  2367 &  &  995 &  &  1839 &  \\
$SC$ & $C_{1\times1}$ & $SC$ & $C_{3\times3}$  & $SC$ & $None$ &  530 &  &  2546 &  & \textbf{2} & 0.58 &  1356 &  &  2368 &  &  996 &  &  1840 &  \\
$SC$ & $C_{3\times3}$  & $C_{1\times1}$ & $SC$ & $None$ & $SC$ &  1704 &  &  2619 &  & \textbf{3} & 0.57 &  1192 &  &  2454 &  &  380 &  &  502 &  \\
$C_{1\times1}$ & $SC$ & $C_{3\times3}$  & $C_{3\times3}$  & $C_{1\times1}$ & $SC$ &  390 &  &  1208 &  & \textbf{4} & 0.57 &  607 &  &  2598 &  &  11 &  &  1119 &  \\
$C_{3\times3}$  & $SC$ & $SC$ & $SC$ & $None$ & $SC$ &  1042 &  &  2754 &  & \textbf{5} & 0.57 &  2432 &  &  2535 &  &  365 &  &  2312 &  \\
$C_{1\times1}$ & $SC$ & $C_{3\times3}$  & $SC$ & $C_{3\times3}$  & $None$ &  2314 &  &  2159 &  & \textbf{6} & 0.57 &  2205 &  &  1612 &  &  1161 &  &  836 &  \\
$SC$ & $SC$ & $C_{3\times3}$  & $SC$ & $None$ & $SC$ &  1445 &  &  2821 &  & \textbf{7} & 0.57 &  2549 &  &  2789 &  &  234 &  &  953 &  \\
$C_{3\times3}$  & $C_{1\times1}$ & $C_{3\times3}$  & $SC$ & $C_{1\times1}$ & $None$ &  313 &  &  348 &  & \textbf{8} & 0.57 &  906 &  &  2042 &  &  1241 &  &  30 &  \\
$SC$ & $C_{1\times1}$ & $C_{3\times3}$  & $SC$ & $SC$ & $C_{3\times3}$  &  145 &  &  790 &  & \textbf{9} & 0.57 &  1894 &  &  2647 &  &  21 &  &  431 &  \\
$SC$ & $C_{3\times3}$  & $C_{3\times3}$  & $None$ & $SC$ & $C_{1\times1}$ &  468 &  &  404 &  & \textbf{10} & 0.57 &  1431 &  &  1781 &  &  1406 &  &  827 &  \\
$C_{3\times3}$  & $C_{3\times3}$  & $SC$ & $C_{3\times3}$  & $C_{3\times3}$  & $C_{1\times1}$ &  1011 &  &  203 &  &  2933 &  & \textbf{1} & 0.60 &  97 &  &  2643 &  &  1089 &  \\
$C_{3\times3}$  & $C_{3\times3}$  & $None$ & $SC$ & $SC$ & $C_{1\times1}$ &  173 &  &  206 &  &  384 &  & \textbf{2} & 0.59 &  685 &  &  242 &  &  503 &  \\
$SC$ & $C_{3\times3}$  & $None$ & $SC$ & $C_{1\times1}$ & $C_{3\times3}$  &  96 &  &  96 &  &  1102 &  & \textbf{3} & 0.59 &  1847 &  &  38 &  &  1099 &  \\
$C_{3\times3}$  & $SC$ & $C_{3\times3}$  & $C_{1\times1}$ & $SC$ & $SC$ &  20 &  &  76 &  &  246 &  & \textbf{4} & 0.58 &  1539 &  &  364 &  &  871 &  \\
$C_{3\times3}$  & $C_{1\times1}$ & $SC$ & $SC$ & $C_{1\times1}$ & $C_{1\times1}$ &  598 &  &  766 &  &  141 &  & \textbf{5} & 0.58 &  1937 &  &  1069 &  &  625 &  \\
$C_{3\times3}$  & $SC$ & $C_{3\times3}$  & $SC$ & $None$ & $C_{1\times1}$ &  672 &  &  1203 &  &  1352 &  & \textbf{6} & 0.58 &  1385 &  &  1221 &  &  275 &  \\
$C_{3\times3}$  & $C_{1\times1}$ & $C_{3\times3}$  & $C_{3\times3}$  & $C_{1\times1}$ & $C_{1\times1}$ &  2488 &  &  51 &  &  2304 &  & \textbf{7} & 0.58 &  12 &  &  2368 &  &  1471 &  \\
$SC$ & $C_{3\times3}$  & $None$ & $SC$ & $None$ & $C_{3\times3}$  &  663 &  &  15 &  &  364 &  & \textbf{8} & 0.58 &  864 &  &  240 &  & \textbf{9} &  94.90 \\
$SC$ & $C_{3\times3}$  & $C_{1\times1}$ & $C_{3\times3}$  & $SC$ & $SC$ &  614 &  &  2518 &  &  825 &  & \textbf{9} & 0.57 &  629 &  &  610 &  &  1129 &  \\
$C_{3\times3}$  & $None$ & $C_{3\times3}$  & $C_{1\times1}$ & $C_{3\times3}$  & $SC$ &  953 &  &  785 &  &  2451 &  & \textbf{10} & 0.57 &  414 &  &  2456 &  &  2023 &  \\
$C_{3\times3}$  & $C_{3\times3}$  & $C_{3\times3}$  & $C_{1\times1}$ & $SC$ & $C_{3\times3}$  &  845 &  &  24 &  &  1782 &  &  101 &  & \textbf{1} & 26.27 &  2250 &  &  781 &  \\
$C_{1\times1}$ & $None$ & $C_{3\times3}$  & $C_{3\times3}$  & $C_{3\times3}$  & $C_{3\times3}$  &  1448 &  &  1324 &  &  2708 &  &  1187 &  & \textbf{2} & 26.10 &  2622 &  &  1910 &  \\
$C_{3\times3}$  & $None$ & $C_{3\times3}$  & $SC$ & $C_{1\times1}$ & $C_{3\times3}$  &  30 &  &  224 &  &  912 &  &  342 &  & \textbf{3} & 25.95 &  2184 &  &  175 &  \\
$C_{3\times3}$  & $SC$ & $C_{3\times3}$  & $C_{3\times3}$  & $C_{3\times3}$  & $C_{1\times1}$ &  658 &  &  210 &  &  3139 &  &  44 &  & \textbf{4} & 25.91 & 1464 &  &  136 &  \\
$C_{3\times3}$  & $C_{1\times1}$ & $C_{3\times3}$  & $C_{3\times3}$  & $None$ & $SC$ &  362 &  &  540 &  &  1928 &  &  206 &  & \textbf{5} & 25.83 &  1693 &  &  1957 &  \\
$C_{3\times3}$  & $None$ & $C_{3\times3}$  & $C_{3\times3}$  & $C_{1\times1}$ & $C_{3\times3}$  &  2081 &  &  788 &  &  2778 &  &  197 &  & \textbf{6} & 25.82 &  1923 &  &  2194 &  \\
$C_{3\times3}$  & $C_{3\times3}$  & $C_{3\times3}$  & $C_{1\times1}$ & $C_{3\times3}$  & $C_{1\times1}$ &  1802 &  &  620 &  &  2358 &  &  36 &  & \textbf{7} & 25.80 &  2713 &  &  2172 &  \\
$C_{3\times3}$  & $C_{3\times3}$  & $C_{3\times3}$  & $C_{1\times1}$ & $C_{3\times3}$  & $SC$ &  703 &  &  507 &  &  2552 &  &  367 &  & \textbf{8} &  25.79 &  2617 &  &  1598 &  \\
$C_{3\times3}$  & $None$ & $C_{3\times3}$  & $None$ & $C_{1\times1}$ & $C_{1\times1}$ &  2848 &  &  2103 &  &  3500 &  &  1208 &  & \textbf{9} & 25.78 &  2697 &  &  2822 &  \\
$C_{3\times3}$  & $C_{3\times3}$  & $C_{3\times3}$  & $None$ & $C_{1\times1}$ & $C_{1\times1}$ &  2707 &  &  1878 &  &  3260 &  &  971 &  & \textbf{10} & 25.76 &  2573 &  &  2821 &  \\
$C_{3\times3}$  & $C_{1\times1}$ & $None$ & $SC$ & $C_{3\times3}$  & $None$ &  344 &  &  59 &  &  510 &  &  252 &  &  1391 &  & \textbf{2} & 0.60 &  1263 &  \\
$C_{3\times3}$  & $C_{1\times1}$ & $SC$ & $SC$ & $C_{3\times3}$  & $None$ &  345 &  &  60 &  &  511 &  &  253 &  &  1392 &  & \textbf{3} & 0.60 &  1264 &  \\
$SC$ & $C_{3\times3}$  & $C_{3\times3}$  & $SC$ & $SC$ & $C_{1\times1}$ &  241 &  &  1246 &  &  103 &  &  1927 &  &  2559 &  & \textbf{4} & 0.60 &  938 &  \\
$SC$ & $C_{3\times3}$  & $C_{3\times3}$  & $SC$ & $C_{1\times1}$ & $SC$ &  2058 &  &  2406 &  &  552 &  &  291 &  &  1383 &  & \textbf{5} & 0.60 &  292 &  \\
$C_{3\times3}$  & $SC$ & $SC$ & $SC$ & $C_{3\times3}$  & $C_{3\times3}$  &  461 &  &  70 &  &  1118 &  &  15 &  &  2081 &  & \textbf{6} & 0.60 &  684 &  \\
$C_{3\times3}$  & $C_{1\times1}$ & $None$ & $C_{1\times1}$ & $C_{1\times1}$ & $C_{1\times1}$ &  862 &  &  1389 &  &  2117 &  &  1659 &  &  370 &  & \textbf{7} & 0.60 &  1148 &  \\
$SC$ & $C_{3\times3}$  & $C_{3\times3}$  & $SC$ & $SC$ & $C_{3\times3}$  &  15 &  &  529 &  &  1306 &  &  512 &  &  1805 &  & \textbf{8} & 0.60 &  599 &  \\
$C_{3\times3}$  & $SC$ & $None$ & $SC$ & $C_{3\times3}$  & $SC$ &  125 &  &  384 &  &  25 &  &  53 &  &  2323 &  & \textbf{9} & 0.60 &  346 &  \\
$SC$ & $C_{3\times3}$  & $SC$ & $SC$ & $C_{3\times3}$  & $C_{3\times3}$  &  307 &  &  807 &  &  1072 &  &  54 &  &  2595 &  & \textbf{10} & 0.60 &  1709 &  \\
$C_{3\times3}$  & $C_{3\times3}$  & $None$ & $SC$ & $C_{3\times3}$  & $C_{1\times1}$ &  428 &  &  95 &  &  1241 &  &  186 &  &  613 &  &  887 &  & \textbf{1} & 95.37 \\
$C_{3\times3}$  & $C_{1\times1}$ & $SC$ & $SC$ & $None$ & $C_{1\times1}$ &  1431 &  &  1059 &  &  402 &  &  1357 &  &  2361 &  &  144 &  & \textbf{2} & 95.22 \\
$C_{3\times3}$  & $C_{3\times3}$  & $None$ & $SC$ & $SC$ & $None$ &  1405 &  &  2652 &  &  1309 &  &  1761 &  &  2619 &  &  213 &  & \textbf{3} & 95.00 \\
$C_{3\times3}$  & $C_{3\times3}$  & $SC$ & $SC$ & $SC$ & $None$ &  1407 &  &  2653 &  &  1310 &  &  1762 &  &  2620 &  &  214 &  & \textbf{4} & 95.00 \\
$None$ & $C_{3\times3}$  & $C_{1\times1}$ & $C_{1\times1}$ & $C_{1\times1}$ & $C_{3\times3}$  &  2671 &  &  321 &  &  2855 &  &  352 &  &  1529 &  &  2302 &  & \textbf{5} & 95.00 \\
$C_{3\times3}$  & $C_{1\times1}$ & $C_{1\times1}$ & $C_{3\times3}$  & $SC$ & $SC$ &  1275 &  &  361 &  &  1404 &  &  311 &  &  958 &  &  1245 &  & \textbf{6} & 94.99 \\
$C_{3\times3}$  & $None$ & $None$ & $SC$ & $C_{1\times1}$ & $C_{1\times1}$ &  328 &  &  1158 &  &  1172 &  &  1681 &  &  2442 &  &  649 &  & \textbf{7} & 94.96 \\
$C_{3\times3}$  & $C_{3\times3}$  & $C_{1\times1}$ & $C_{1\times1}$ & $SC$ & $C_{1\times1}$ &  2146 &  &  493 &  &  1975 &  &  169 &  &  209 &  &  2041 &  & \textbf{8} & 94.91 \\
$C_{1\times1}$ & $SC$ & $C_{1\times1}$ & $C_{3\times3}$  & $SC$ & $SC$ &  922 &  &  2121 &  &  196 &  &  1249 &  &  2145 &  &  421 &  & \textbf{10} & 94.88 \\ \bottomrule
\end{tabular}
}
\vspace{-1em}
\end{table*}

% topcell %
%%%%%%%%%%%

%%%%%%%%%%
% rank %

% rank %
%%%%%%%%%%%

\section{Suggestions for Future NAS Work}
The presented results shed insights on the importance of the different operations in a NAS search space, how they are influenced by their positioning, their combinations and how this influences the performance of an architecture. Based on those, we consider that they have a two-fold application: 1) on how NAS methods should be evaluated; and 2) what researchers should take into consideration when designing and creating new NAS benchmarks.

\subsection{Evaluating NAS Methods}
Our findings suggest that specific operations, such as $C_{3\times3}$, have a considerable impact on the final performance of an architecture. To promote a thorough evaluation of NAS methods, we recommend that a NAS evaluation method should consider:
\begin{itemize}
    \item Evaluating NAS methods over time, reporting performance as an average and standard deviation to allow analysis of precision and trainability of the method.
    \item Analyzing the cell's design to determine if a NAS method focuses only on a small subset of the operation pool, which can hinder optimal results. If this is done over time, it can provide information about how well a NAS method generalizes and learns.
    \item Evaluating NAS methods on multiple benchmarks, not just a single one, especially if the analyzed task allows most architectures to attain high performance, such as the CIFAR-10 data set. This corroborates past suggestions and findings \cite{lindauer2020best,Yang2020NAS,wan2022on,mehta2022nasbenchsuite}. This would allow a proper evaluation of the method's generability, as it could be easily overfitting a setting from a small search space on a given benchmark.
    \item Evaluating NAS methods by both searching directly on all data sets of a benchmark and also by transferring searched architectures to all other data sets. This will allow an evaluation of the method's generability and the capability of generating architectures in different scenarios and constraints.
\end{itemize}
%1) NAS methods are evaluated over time, and the performance of a NAS method is given as average and standard deviation, allowing the analysis of the precision and trainability of the NAS method; 
%2) authors should promote analysis of the cell's design, to further provide information if a NAS method focuses only on a small subset of the operation pool, which can be detrimental to achieve optimal results; 
%3) evaluating on a single benchmark is not sufficient, specially if the analyzed data set allows most architectures to attain high performance, which is the case of CIFAR-10; 
%4) NAS methods should be evaluated on searching directly on all data sets of a benchmark, thus allowing a proper evaluation of the method's generability. More, the searched architectures on each data set should also be transferred and evaluated on all other data sets. This is usually done by searching on CIFAR-10 and transferring to other data sets, but this setting doesn't allow a proper evaluation of a method capability of generating architectures that are capable of properly learning new problems;
%and 5) corroborating past suggestions and findings \cite{lindauer2020best,Yang2020NAS,wan2022on,mehta2022nasbenchsuite}, researchers should focus on evaluating NAS methods on multiple benchmarks, allowing a proper evaluation of the method generability, as it could be easily overfitting a setting from a small search space on a given benchmark.

\subsection{Designing NAS benchmarks}
Based on the findings presented in this paper, we propose several suggestions for researchers interested in designing and creating new NAS benchmarks. Firstly, it is important to note that within the search space of all three benchmarks evaluated, there are operations that have the most importance -- convolutional layers and $SC$. This suggests that a subset of the operations pool is sufficient to generate the best scoring architectures. Additionally, our results show a high correlation between data sets within a benchmark, hindering the evaluation of NAS methods in terms of generability and transferability. To address this, researchers should consider the following when designing NAS benchmarks:

\begin{itemize}
    \item The cell structure has a high impact on the final architecture performance. This is seen in NAS-Bench-201 and TransNAS-Bench-101 when $SC$ layers are often placed in the 4th position. Therefore, researchers should take this into account when designing their benchmarks by progressively analyzing the architectures as the benchmark is developed. This can be done by looking at a small set of architectures and partially training them. In this, looking the learning curve and zero-proxy estimators can help in obtaining quicker estimates of how the operations and the cell structure impacts the final performance.
    \item Small search spaces could lead to operation overfitting, suggesting that if possible, authors should create an operation space that contains more operations than existing ones. This could be manageable by reducing the cell structure, fixing specific operations, and increasing the operation pool to include more operations such as activation layers and a wider range of pooling operations.
    \item New benchmarks should strive to increase the number of architectures. While a small number of architectures is required due to the extreme computational costs of training thousands of architectures, benchmarks could be designed with a larger search space and similar computational costs by training the architectures with a smaller number of epochs or simplifying the architecture's outer-skeleton to reduce the number of parameters. In the studied benchmarks, NAS-Bench-101 is the only one with a large set of possible architectures, with over 423k candidates. In contrast, NAS-Bench-201 only has 15k architectures and TransNAS-Bench-101 cell space has approximately 4k possible architectures.
    \item NAS benchmarks should include different types of constraints and information regarding the architectures, such as latency, power consumption, and model size, among others. This will allow the evaluation of NAS methods that search architectures that meet different practical constraints.
    \item Benchmarks should provide a comprehensive evaluation metric that takes into consideration both accuracy and practical constraints to evaluate the overall performance of the searched architectures. This will ensure that the searched architectures are not only accurate but also practical and feasible for real-world applications.
\end{itemize}

Furthermore, as architectures become less dependent on convolutional layers, their training time will reduce. Convolutional layers are known for their excellent feature extraction capabilities but come with the cost of more parameters. To reduce inference time and the number of parameters, researchers can use convolutional layers that reduce computational costs, such as depth-wise separable convolutions \cite{DBLP:conf/cvpr/Chollet17}. These layers have been used extensively in the DARTS search space \cite{DBLP:conf/iclr/LiuSY19}.

\section{Conclusions}
\label{sec:concs}
In this work, we present an analysis of three popular NAS benchmarks in terms of their operations. We found that convolutional layers are essential to have architectures with high performance, and also found that the distribution of the accuracy ranges are skewed, suggesting that finding architectures with accuracies closer to the upper bound is not hard, as several operation patterns, such as their positioning in a cell and their combinations, tend to yield top-scoring architectures. From the analysis conducted, it is clear that specific patterns tend to yield better architectures, such as having specific operations in specific positions of the cell's structure. These findings were also supported by showing that two consecutive combination of operations can have tremendous impact on the architecture's performance, suggesting that convolutional layers and skip connections are more important to generate top-scoring architectures in the evaluated benchmarks. Moreover, TransNAS-Bench-101 presents more heterogeneity in terms of the top-scoring architectures, suggesting that NAS methods should refrain from searching only on a single data set and then transferring the top-scoring architecture, to directly search on the three data sets independently. This was further supported by finding that inter-data set ranking has a high Kendall tau correlation. Finally, we discuss how NAS methods should take the findings of this paper into consideration to promote fair comparisons and to design new benchmarks.

Notwithstanding that the findings of this work show that some operations can be essential for having good architectures and that there is a high correlation between operations and the final performance of the architectures, we believe that benchmarks are a key part of NAS research. In summary, we believe that the suggestions presented in this paper can guide future research in NAS, both in terms of evaluating NAS methods and designing new NAS benchmarks. By considering the importance of different operations in a search space and their impact on the final performance of the searched architectures, researchers can better design and evaluate NAS methods and benchmarks, leading to more efficient and effective NAS approaches.

{\small
\bibliographystyle{ieee_fullname}
\bibliography{bib}
}

\vfill

\end{document}